\RequirePackage{fix-cm}
\documentclass{article} 
\usepackage{iclr2026_conference,times}
\usepackage{amsmath}
\usepackage{tikz}
\usetikzlibrary{calc, shapes.geometric,arrows.meta}
\usepackage{pgfplots}
\pgfplotsset{compat=1.18}
\usepgfplotslibrary{groupplots}
\usepackage{wrapfig}
\usepackage{tabularx}
\usepackage{titlesec}
\usepackage{pifont}
  \titlespacing*{\paragraph}{0pt}{4pt}{0.7em}

\usepackage{amsmath,amsfonts,bm}

\def\eqref#1{equation~\ref{#1}}

\def\1{\bm{1}}

\def\rx{{\textnormal{x}}}

\DeclareMathAlphabet{\mathsfit}{\encodingdefault}{\sfdefault}{m}{sl}
\SetMathAlphabet{\mathsfit}{bold}{\encodingdefault}{\sfdefault}{bx}{n}

\usepackage{hyperref}
\usepackage{url}
\usepackage{subcaption}
\usepackage{cleveref}
\usepackage[inline]{enumitem}
\usepackage{colortbl}
\usepackage{booktabs}
\usepackage{multirow}
\usepackage[T1]{fontenc}

\definecolor{MatGreen}{HTML}{2E7D32}
\definecolor{MatRed}{HTML}{C62828}
\definecolor{CiteGray}{HTML}{3F51B5}
\definecolor{FigPink}{HTML}{FF4081}
\definecolor{AuthorAccent}{HTML}{35708E}
\providecolor{adPff}{HTML}{FF6F00}

\DeclareRobustCommand{\correspondencemark}{%
  \textcolor{AuthorAccent}{%
    \fontsize{14}{14}\selectfont\textasteriskcentered}}

\DeclareRobustCommand{\advisingmark}{%
  \textcolor{adPff}{%
    \fontsize{10}{10}\selectfont\ding{80}}}

\newcommand{\jointadvising}{%
  {\normalfont\small\advisingmark\,\textbf{Joint advising}}%
}

\hypersetup{
    colorlinks=true,
    citecolor=CiteGray, 
    linkcolor=FigPink,  
    urlcolor=CiteGray   
}

\title{Planetary Feature Fields Are Scalable \\ Earth Representations}

\author{{\normalfont\small
\begin{tabular}[t]{@{}*{3}{>{\raggedright\arraybackslash}p{\dimexpr(\textwidth-2\tabcolsep)/3\relax}@{}}}
\textbf{\normalsize Arjun Rao}%
\rlap{\,\hyperlink{pff-correspondence}{\correspondencemark}} &
\textbf{\normalsize Sebastian Loeschcke} &
\textbf{\normalsize Anthony Fuller} \\
University of British Columbia &
University of Copenhagen &
Carleton University \\
University of Copenhagen && Vector Institute \\
Vector Institute && \\[9pt]
\textbf{\normalsize Isaac Corley} &
\textbf{\normalsize Nico Lang}\rlap{\,\advisingmark} &
\textbf{\normalsize Evan Shelhamer}\rlap{\,\advisingmark} \\
Taylor Geospatial &
University of Copenhagen &
University of British Columbia \\
&& Vector Institute \\[1pt]
\multicolumn{3}{c}{\jointadvising}
\end{tabular}}
}

\iclrfinalcopy

\begin{document}

\maketitle

\lhead{Preprint}

\begingroup
\renewcommand{\thefootnote}{\protect\correspondencemark}
\footnotetext[0]{%
  \hypertarget{pff-correspondence}{}%
  Corresponding author. Please email 
  \href{mailto:arjun.rao@ubc.ca}{%
    \textcolor{AuthorAccent}{arjun.rao@ubc.ca}}.%
}
\endgroup

\vspace{-10pt}
\begin{figure}[h]
\centering
\resizebox{\linewidth}{!}{%
\providecolor{matOrangeD}{HTML}{EF8A2C}%
\definecolor{heroEmbedding}{HTML}{6B5AA8}%
\definecolor{heroRaw}{HTML}{35708E}%
\definecolor{heroMap}{HTML}{4F8A5C}%
\begin{tikzpicture}[x=1cm,y=1cm,font=\normalfont\fontsize{7}{8}\selectfont]
\node[inner sep=0,anchor=base,font=\fontsize{7.5}{9}\selectfont] at (1.0250,5.0663) {Original Product};
\node[inner sep=0,anchor=base,font=\fontsize{7.5}{9}\selectfont] at (3.3550,5.0663) {\textcolor{matOrangeD}{\textbf{PFF}}-VM};
\node[inner sep=0,anchor=base,font=\fontsize{7.5}{9}\selectfont] at (5.5250,5.0663) {\textcolor{matOrangeD}{\textbf{PFF}}-K-Planes};
\node[inner sep=0,anchor=base,font=\fontsize{7.5}{9}\selectfont] at (7.8550,5.0663) {TensoRF (VM)};
\node[inner sep=0,anchor=base,font=\fontsize{7.5}{9}\selectfont] at (10.0250,5.0663) {NGP};
\node[inner sep=0,anchor=base,font=\fontsize{7.5}{9}\selectfont] at (12.1950,5.0663) {SIREN};
\node[inner sep=0,anchor=south west] at (0.0000,3.3350) {\includegraphics[width=2.05cm]{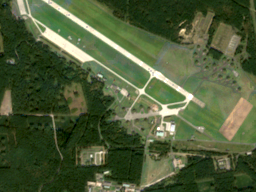}};
\draw[heroRaw!65,line width=.25pt] (0.0000,3.3350) rectangle ++(2.05,1.5375);
\node[inner sep=0,anchor=south west] at (2.3300,3.3350) {\includegraphics[width=2.05cm]{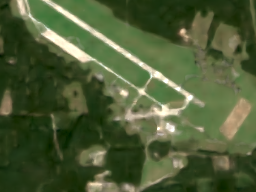}};
\draw[heroRaw!65,line width=.25pt] (2.3300,3.3350) rectangle ++(2.05,1.5375);
\node[inner sep=0,anchor=south west] at (4.5000,3.3350) {\includegraphics[width=2.05cm]{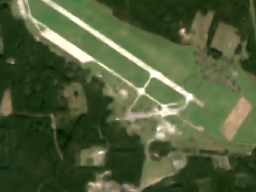}};
\draw[heroRaw!65,line width=.25pt] (4.5000,3.3350) rectangle ++(2.05,1.5375);
\node[inner sep=0,anchor=south west] at (6.8300,3.3350) {\includegraphics[width=2.05cm]{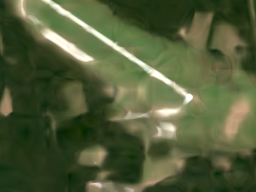}};
\draw[heroRaw!65,line width=.25pt] (6.8300,3.3350) rectangle ++(2.05,1.5375);
\node[inner sep=0,anchor=south west] at (9.0000,3.3350) {\includegraphics[width=2.05cm]{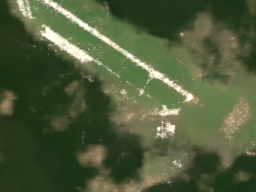}};
\draw[heroRaw!65,line width=.25pt] (9.0000,3.3350) rectangle ++(2.05,1.5375);
\node[inner sep=0,anchor=south west] at (11.1700,3.3350) {\includegraphics[width=2.05cm]{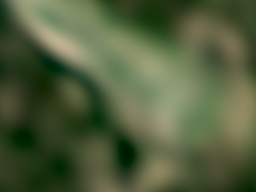}};
\draw[heroRaw!65,line width=.25pt] (11.1700,3.3350) rectangle ++(2.05,1.5375);
\node[inner sep=0,anchor=base east,text=heroRaw,font=\fontsize{7.6}{9}\selectfont] at (-.22,4.3440) {\textbf{Observations}};
\node[inner sep=0,anchor=base east,font=\fontsize{6.9}{8}\selectfont] at (-.22,4.0840) {Sentinel-2\,\textcolor{black!35}{\textperiodcentered}\,\textcolor{heroRaw}{\textbf{2017}}};
\node[inner sep=0,anchor=base east,text=black!65,font=\fontsize{5.8}{6.6}\selectfont] at (-.22,3.8837) {53.389$^\circ$N};
\node[inner sep=0,anchor=base east,text=black!65,font=\fontsize{5.8}{6.6}\selectfont] at (-.22,3.6637) {16.089$^\circ$E};
\node[inner sep=0,anchor=south west] at (0.0000,1.6675) {\includegraphics[width=2.05cm]{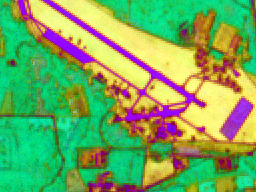}};
\draw[heroEmbedding!65,line width=.25pt] (0.0000,1.6675) rectangle ++(2.05,1.5375);
\node[inner sep=0,anchor=south west] at (2.3300,1.6675) {\includegraphics[width=2.05cm]{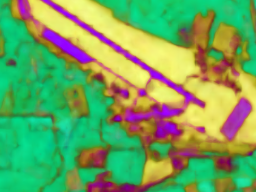}};
\draw[heroEmbedding!65,line width=.25pt] (2.3300,1.6675) rectangle ++(2.05,1.5375);
\node[inner sep=0,anchor=south west] at (4.5000,1.6675) {\includegraphics[width=2.05cm]{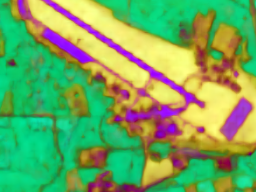}};
\draw[heroEmbedding!65,line width=.25pt] (4.5000,1.6675) rectangle ++(2.05,1.5375);
\node[inner sep=0,anchor=south west] at (6.8300,1.6675) {\includegraphics[width=2.05cm]{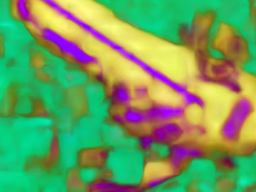}};
\draw[heroEmbedding!65,line width=.25pt] (6.8300,1.6675) rectangle ++(2.05,1.5375);
\node[inner sep=0,anchor=south west] at (9.0000,1.6675) {\includegraphics[width=2.05cm]{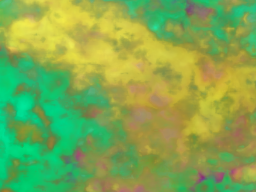}};
\draw[heroEmbedding!65,line width=.25pt] (9.0000,1.6675) rectangle ++(2.05,1.5375);
\node[inner sep=0,anchor=south west] at (11.1700,1.6675) {\includegraphics[width=2.05cm]{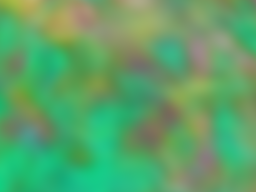}};
\draw[heroEmbedding!65,line width=.25pt] (11.1700,1.6675) rectangle ++(2.05,1.5375);
\node[inner sep=0,anchor=base east,text=heroEmbedding,font=\fontsize{7.6}{9}\selectfont] at (-.22,2.6765) {\textbf{Embeddings}};
\node[inner sep=0,anchor=base east,font=\fontsize{6.9}{8}\selectfont] at (-.22,2.4165) {TESSERA\,\textcolor{black!35}{\textperiodcentered}\,\textcolor{heroEmbedding}{\textbf{2021}}};
\node[inner sep=0,anchor=base east,text=black!65,font=\fontsize{5.8}{6.6}\selectfont] at (-.22,2.2162) {53.389$^\circ$N};
\node[inner sep=0,anchor=base east,text=black!65,font=\fontsize{5.8}{6.6}\selectfont] at (-.22,1.9962) {16.089$^\circ$E};
\node[inner sep=0,anchor=south west] at (0.0000,0.0000) {\includegraphics[width=2.05cm]{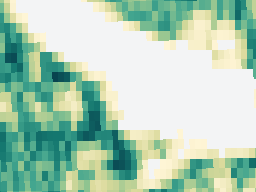}};
\draw[heroMap!65,line width=.25pt] (0.0000,0.0000) rectangle ++(2.05,1.5375);
\node[inner sep=0,anchor=south west] at (2.3300,0.0000) {\includegraphics[width=2.05cm]{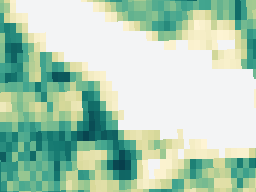}};
\draw[heroMap!65,line width=.25pt] (2.3300,0.0000) rectangle ++(2.05,1.5375);
\node[inner sep=0,anchor=south west] at (4.5000,0.0000) {\includegraphics[width=2.05cm]{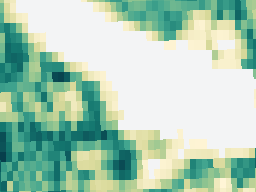}};
\draw[heroMap!65,line width=.25pt] (4.5000,0.0000) rectangle ++(2.05,1.5375);
\node[inner sep=0,anchor=south west] at (6.8300,0.0000) {\includegraphics[width=2.05cm]{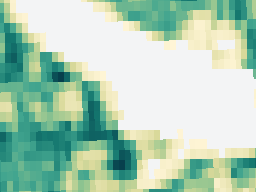}};
\draw[heroMap!65,line width=.25pt] (6.8300,0.0000) rectangle ++(2.05,1.5375);
\node[inner sep=0,anchor=south west] at (9.0000,0.0000) {\includegraphics[width=2.05cm]{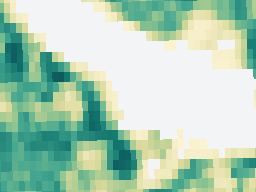}};
\draw[heroMap!65,line width=.25pt] (9.0000,0.0000) rectangle ++(2.05,1.5375);
\node[inner sep=0,anchor=south west] at (11.1700,0.0000) {\includegraphics[width=2.05cm]{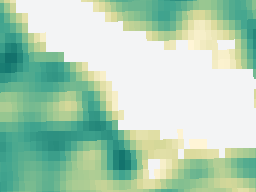}};
\draw[heroMap!65,line width=.25pt] (11.1700,0.0000) rectangle ++(2.05,1.5375);
\node[inner sep=0,anchor=base east,text=heroMap,font=\fontsize{7.6}{9}\selectfont] at (-.22,1.0090) {\textbf{Map Products}};
\node[inner sep=0,anchor=base east,font=\fontsize{6.9}{8}\selectfont] at (-.22,0.7490) {Biomass\,\textcolor{black!35}{\textperiodcentered}\,\textcolor{heroMap}{\textbf{2022}}};
\node[inner sep=0,anchor=base east,text=black!65,font=\fontsize{5.8}{6.6}\selectfont] at (-.22,0.5487) {53.389$^\circ$N};
\node[inner sep=0,anchor=base east,text=black!65,font=\fontsize{5.8}{6.6}\selectfont] at (-.22,0.3287) {16.089$^\circ$E};
\node[inner sep=0,anchor=south west] at (-1.27,0.1350) {\includegraphics[width=1.05cm,height=.065cm]{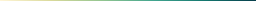}};
\node[inner sep=0,anchor=north west,text=black!65,font=\fontsize{5}{6}\selectfont] at (-1.27,0.1050) {0};
\node[inner sep=0,anchor=north east,text=black!65,font=\fontsize{5}{6}\selectfont] at (-.22,0.1050) {300 Mg/ha};
\end{tikzpicture}
}
\caption{\textbf{Planetary Feature Fields (PFFs) jointly represent heterogeneous Earth observation (EO) products with high fidelity at high compression ratios ($\approx$1800$\times$).} 
We present PFFs as spatially local, explicit--implicit (hybrid) neural fields that share a factored feature volume across raw observations, precomputed embeddings, and map products.
Queried by space--time coordinates, this shared representation exploits redundancy across products, preserves downstream utility, and supports adding new timesteps and products while leaving the existing representation unchanged. }
\label{fig:pff-teaser}
\end{figure}
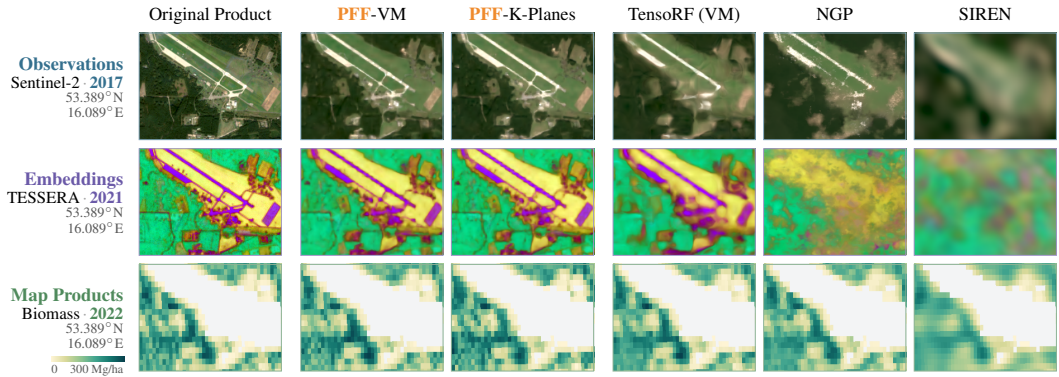

\begin{abstract}
Satellite observations, precomputed embeddings, and map products describe the same evolving Earth, yet are stored as independent, petabyte-scale data products.
Their continued growth calls for compact representations of multiple  products while preserving spatial and temporal detail.
We introduce Planetary Feature Fields (PFFs), which exploit redundancy across data products by modeling them jointly as continuous functions of space and time at planetary scale.
PFFs are spatially local explicit--implicit (\emph{hybrid}) neural fields.
Each field shares a \emph{factored feature volume}---a decomposition of an explicit 3D grid with smaller factors---across products, while lightweight implicit decoders reconstruct individual products across multiple timesteps.
PFFs reconstruct EO products over space and time more accurately than single-product fields at matched compression rates.
At $1800\times$ compression relative to the uncompressed source data, reconstructed features retain approximately $90\%$ or more of the performance achieved with the original features on pixel-level segmentation, change detection, and patch-level classification tasks.
PFFs can add new timesteps by extending their factored feature volumes and add new products by attaching new decoders, while leaving existing outputs unchanged.
PFFs reduce end-to-end feature access latency by an order of magnitude relative to evaluated API and cloud-storage pipelines.
\end{abstract}

\section{Introduction}

Learning compact representations of petabyte-scale Earth observation (EO) data is a central goal in machine learning for remote sensing, supporting socioeconomic \citep{povertymaps,poverty2,distastermapping}, climatic \citep{ccai,climplicit}, and ecological \citep{ecological1,cropharvest,miam,presto} applications. 
Raw observations, precomputed embeddings, and map products encode different properties of the same evolving Earth and can provide complementary information for mapping \citep{gordon2026mmearthbench,van2026better,raomm}.
Although these products differ in purpose, resolution, and semantics, they cover overlapping locations and times. 
They are nevertheless produced, stored, and updated independently, leaving an opportunity to share their representation while preserving the information needed to reconstruct each product.
Such a shared representation must (i) preserve detail across space and time at planetary scale, (ii) be compact to store and support fast retrieval, and (iii) extend efficiently to new timesteps and products.
These requirements are particularly difficult to reconcile for learned representations in EO. Pretrained models extract features from raw observations, but applying them across large geographic areas still requires substantial computation and access to large data archives \citep{ai2_olmoearth_platform_2026}. 
Two alternatives reduce this dependence on raw inputs. 
Geographic location encoders provide compact \emph{implicit} representations that retrieve features directly from geographic coordinates \citep{mai2022review}.
At global scale, these implicit representations capture broad geographic structure but do not preserve spatial detail needed to reconstruct the source data \citep{dhakal2025range,rao2026localized}. 
They are therefore useful as spatial/geographic priors for downstream models \citep{wrap}, but are poor substitutes for the data from which they were learned.
Precomputed embeddings instead store model outputs as \emph{explicit} representations, preserving features at sampled locations and times without rerunning the source model \citep{czerkawski2024global,earthembeddings,alphaearth,tessera}. 
This moves computation offline but results in substantial data storage and transfer costs.
As temporal coverage grows, each timestep must be materialized and maintained as another global array, causing storage, transfer, and versioning costs to grow roughly linearly with time.\footnote{A single year of global $10\,\mathrm{m}$ Earth embeddings is estimated to require $96~\mathrm{TB}$ for a 64-dimensional int8 field and $6.1~\mathrm{PB}$ for a 1024-dimensional float32 field \citep{compressingearthembeddings1}.}

\ifdefined\pffMethodsFigureLoaded\else
\gdef\pffMethodsFigureLoaded{}%

\usetikzlibrary{arrows.meta, calc}

\definecolor{pffLearned}{HTML}{6B5AA8}
\definecolor{pffRaw}{HTML}{35708E}
\definecolor{pffMaps}{HTML}{4F8A5C}
\definecolor{pffInk}{HTML}{3E434A}
\definecolor{pffFrame}{HTML}{BFC3CA}

\definecolor{aeV}{HTML}{8574C9}\definecolor{aeT}{HTML}{5FA3A8}
\definecolor{aeM}{HTML}{B06FB2}\definecolor{aeI}{HTML}{5661B3}
\definecolor{aeD}{HTML}{46397E}\definecolor{aeBg}{HTML}{DDD8F0}
\definecolor{teA}{HTML}{C7BDE9}\definecolor{teB}{HTML}{A393D8}
\definecolor{teC}{HTML}{7E6DBE}\definecolor{teD}{HTML}{A8C2D8}
\definecolor{teE}{HTML}{6D5CA8}\definecolor{teO}{HTML}{4E4180}
\definecolor{teBg}{HTML}{D5CEEC}
\definecolor{sTwoA}{HTML}{B5D18B}\definecolor{sTwoB}{HTML}{8CB868}
\definecolor{sTwoC}{HTML}{659A4C}\definecolor{sTwoY}{HTML}{E0CD74}
\definecolor{sTwoT}{HTML}{D9C9A3}\definecolor{sTwoBg}{HTML}{EDF3E3}
\definecolor{lsA}{HTML}{DCC492}\definecolor{lsB}{HTML}{C6A76F}
\definecolor{lsC}{HTML}{97A05E}\definecolor{lsSw}{HTML}{A98F5F}
\definecolor{lsBg}{HTML}{F5F0E3}
\definecolor{lstA}{HTML}{F5C868}\definecolor{lstB}{HTML}{EEA050}
\definecolor{lstC}{HTML}{DF6E3D}\definecolor{lstD}{HTML}{C74B30}
\definecolor{lstBg}{HTML}{FBF1E5}
\definecolor{sOneA}{HTML}{5B87A6}\definecolor{sOneB}{HTML}{35708E}
\definecolor{sOneBg}{HTML}{E9F0F5}
\definecolor{palA}{HTML}{5E6FA8}\definecolor{palB}{HTML}{46548C}
\definecolor{palBg}{HTML}{EBEDF6}
\definecolor{gpwM}{HTML}{CFE2AC}\definecolor{gpwD}{HTML}{6F9A4D}
\definecolor{gpwT}{HTML}{55813D}\definecolor{gpwBg}{HTML}{F1F5E6}
\definecolor{agbA}{HTML}{85B270}\definecolor{agbB}{HTML}{5E9455}
\definecolor{agbO}{HTML}{40703C}\definecolor{agbBg}{HTML}{EFF4EB}
\definecolor{watF}{HTML}{85BADB}\definecolor{watO}{HTML}{3E7CA6}
\definecolor{watR}{HTML}{D6EAF5}\definecolor{watBg}{HTML}{EAF3F8}
\definecolor{demL}{HTML}{B48C60}\definecolor{demI}{HTML}{8A6A44}
\definecolor{demBg}{HTML}{F7F2E9}
\definecolor{canA}{HTML}{74A66E}\definecolor{canB}{HTML}{548549}
\definecolor{canO}{HTML}{3E6B3B}\definecolor{canT}{HTML}{7A5C40}
\definecolor{canH}{HTML}{5A5F66}\definecolor{canBg}{HTML}{EFF5EE}
\definecolor{wetF}{HTML}{ACD5C9}\definecolor{wetO}{HTML}{5FA08F}
\definecolor{wetR}{HTML}{4E8A5F}\definecolor{wetBg}{HTML}{EDF4F1}
\definecolor{soilA}{HTML}{6E4F35}\definecolor{soilB}{HTML}{8A6A44}
\definecolor{soilC}{HTML}{B08A5E}\definecolor{soilD}{HTML}{D2B48C}
\definecolor{soilP}{HTML}{8C7355}\definecolor{soilBg}{HTML}{F5EFE6}

\tikzset{
  pff/corners/.style={rounded corners=3.2pt},
  pff/frame/.style={pff/corners, draw=pffFrame, line width=0.55pt},
  every pff stroke/.style={line cap=round, line join=round},
  pff/tile label/.style={font=\fontsize{4}{5}\selectfont\sffamily, text=black,
                         inner sep=1pt, anchor=north, align=center},
  pff/group title/.style={font=\small\bfseries\sffamily, text=black,
                          inner sep=1pt, anchor=south, align=center},
  pff/panel title/.style={font=\fontsize{7}{8.4}\selectfont\sffamily,
                          text=black, inner sep=1pt, anchor=south,
                          align=center},
  pff/stackframe/.style={pff/corners, draw=pffFrame, line width=0.45pt},
  pff/group box/.style={rounded corners=5pt, line width=0.8pt},
}

\definecolor{archPlane}{HTML}{E8EEF4}
\definecolor{archLine}{HTML}{F4EDE0}
\definecolor{archDot}{HTML}{B03A2E}
\definecolor{globeDot}{HTML}{46397E}
\definecolor{matOrange}{HTML}{FFAB40}
\definecolor{matOrangeD}{HTML}{EF8A2C}
\definecolor{archFlow}{HTML}{B96F2E}
\definecolor{decB}{HTML}{4A7FA5}
\definecolor{decFillB}{HTML}{DEEBF3}
\definecolor{lossC}{HTML}{A845C6}
\definecolor{addG}{HTML}{4CAF50}
\definecolor{remR}{HTML}{E53935}
\definecolor{pcaP}{HTML}{DC569F}
\definecolor{pcaV}{HTML}{9A3277}
\definecolor{pcaG}{HTML}{3FCE62}
\definecolor{pcaB}{HTML}{41AEDD}
\definecolor{pcaO}{HTML}{F2A64B}
\definecolor{decFill}{HTML}{EFE1F5}
\definecolor{decLine}{HTML}{A569BD}
\tikzset{
  pff/arch/lbl/.style={font=\fontsize{4}{4.8}\selectfont\sffamily, text=black,
                       inner sep=1pt},
  pff/arch/input/.style={font=\fontsize{10}{12}\selectfont\bfseries\sffamily,
                         text=black, inner sep=1pt},
  pff/arch/loss/.style={font=\fontsize{7.5}{9}\selectfont, text=black,
                        inner sep=1pt, anchor=south},
  pff/arch/lossfirst/.style={font=\fontsize{5.85}{7}\selectfont\sffamily,
                             text=black, align=center, anchor=south,
                             inner sep=1pt},
  pff/arch/math/.style={font=\fontsize{10}{12}\selectfont, text=black,
                        inner sep=1pt},
  pff/arch/opmath/.style={font=\fontsize{7}{8.4}\selectfont, text=black,
                          fill=white, inner sep=1.4pt},
  pff/arch/net/.style={draw=black!30, line width=0.25pt},
  pff/arch/neuron/.style={fill=decFillB, draw=decB, line width=0.5pt},
  pff/arch/arrow/.style={-{Stealth[length=2.8pt]}, line width=0.7pt,
                         draw=black, line cap=round},
  pff/arch/wire/.style={line width=0.5pt, draw=black, line cap=round},
  pff/arch/fieldbox/.style={rounded corners=5pt, fill=matOrange,
                            fill opacity=0.16, draw=matOrangeD,
                            line width=0.7pt},
  pff/arch/partbox/.style={rounded corners=3.5pt, fill=matOrange!8,
                           draw=matOrangeD!65, line width=0.6pt},
  pff/arch/decbox/.style={rounded corners=2.5pt, fill=decFill, draw=decLine,
                          line width=0.55pt},
  pff/arch/region/.style={rounded corners=2pt, fill=black!4, draw=black!45,
                          line width=0.5pt},
  pff/arch/plane/.style={draw=black, line width=0.5pt, fill=archPlane},
  pff/arch/grid/.style={draw=black!25, line width=0.25pt},
  pff/arch/bar/.style={draw=black, line width=0.5pt, fill=archLine},
  pff/arch/sample/.style={fill=archDot},
  pff/arch/zcell/.style={draw=black!55, line width=0.3pt, fill=aeV!28},
  pff/arch/slot/.style={rounded corners=2pt, draw=decB!70, line width=0.5pt,
                        dash pattern=on 1.8pt off 1.3pt},
  pff/arch/frozenbox/.style={rounded corners=5pt,
                             fill=matOrange!25!gray, fill opacity=0.38,
                             draw=matOrangeD!45!gray, line width=0.7pt},
  pff/globe/circle/.style={draw=black!70, line width=0.55pt},
  pff/globe/grat/.style={draw=black!25, line width=0.3pt},
  pff/globe/dot/.style={fill=globeDot},
  pff/globe/focus/.style={draw=archDot, line width=0.5pt,
                          dash pattern=on 1.6pt off 1.1pt},
}

\tikzset{
  pff/ae/bg/.style={fill=aeBg},
  pff/ae/l1/.style={every pff stroke, fill=aeV, draw=aeV!75!black,
                    line width=0.35pt},
  pff/ae/l2/.style={every pff stroke, fill=aeT, draw=aeT!75!black,
                    line width=0.35pt},
  pff/ae/l3/.style={every pff stroke, fill=aeM, draw=aeM!75!black,
                    line width=0.35pt},
  pff/ae/l4/.style={every pff stroke, fill=aeI, draw=aeI!75!black,
                    line width=0.35pt},
  pff/ae/l5/.style={every pff stroke, fill=aeD, draw=aeD!75!black,
                    line width=0.35pt},
  pff/ae/flow/.style={every pff stroke, draw=white, line width=0.6pt,
                      opacity=0.4},
  pff/te/bg/.style={fill=teBg},
  pff/te/cell a/.style={every pff stroke, fill=teA, draw=teO, line width=0.35pt},
  pff/te/cell b/.style={every pff stroke, fill=teB, draw=teO, line width=0.35pt},
  pff/te/cell c/.style={every pff stroke, fill=teC, draw=teO, line width=0.35pt},
  pff/te/cell d/.style={every pff stroke, fill=teD, draw=teO, line width=0.35pt},
  pff/te/cell e/.style={every pff stroke, fill=teE, draw=teO, line width=0.35pt},
  pff/stwo/bg/.style={fill=sTwoBg},
  pff/stwo/patch a/.style={every pff stroke, fill=sTwoA, draw=sTwoC,
                           line width=0.5pt},
  pff/stwo/patch b/.style={every pff stroke, fill=sTwoB, draw=sTwoC,
                           line width=0.5pt},
  pff/stwo/patch c/.style={every pff stroke, fill=sTwoC, draw=sTwoC!70!black,
                           line width=0.5pt},
  pff/stwo/patch y/.style={every pff stroke, fill=sTwoY, draw=sTwoY!60!black,
                           line width=0.5pt},
  pff/stwo/patch t/.style={every pff stroke, fill=sTwoT, draw=sTwoT!60!black,
                           line width=0.5pt},
  pff/stwo/cloud/.style={every pff stroke, fill=white, draw=gray!65,
                         line width=0.7pt},
  pff/stwo/cloud shadow/.style={fill=sTwoC!30, fill opacity=0.5},
  pff/landsat/bg/.style={fill=lsBg},
  pff/landsat/patch a/.style={every pff stroke, fill=lsA, draw=lsB!80!black,
                              line width=0.5pt},
  pff/landsat/patch b/.style={every pff stroke, fill=lsB, draw=lsB!60!black,
                              line width=0.5pt},
  pff/landsat/patch c/.style={every pff stroke, fill=lsC, draw=lsC!70!black,
                              line width=0.5pt},
  pff/landsat/swath/.style={every pff stroke, draw=lsSw, line width=0.8pt,
                            dash pattern=on 3.5pt off 2.2pt, opacity=0.75},
  pff/lst/bg/.style={fill=lstBg},
  pff/lst/iso a/.style={every pff stroke, fill=lstA, draw=lstB, line width=0.6pt},
  pff/lst/iso b/.style={every pff stroke, fill=lstB, draw=lstC, line width=0.6pt},
  pff/lst/iso c/.style={every pff stroke, fill=lstC, draw=lstD, line width=0.6pt},
  pff/lst/iso d/.style={every pff stroke, fill=lstD, draw=lstD!75!black,
                        line width=0.6pt},
  pff/sone/bg/.style={fill=sOneBg},
  pff/sone/wave a/.style={every pff stroke, draw=sOneA, line width=1.0pt},
  pff/sone/wave b/.style={every pff stroke, draw=sOneB, line width=0.7pt},
  pff/sone/speckle/.style={every pff stroke, draw=sOneB!55, line width=0.6pt},
  pff/palsar/bg/.style={fill=palBg},
  pff/palsar/wave hh/.style={every pff stroke, draw=palA, line width=1.3pt},
  pff/palsar/wave hv/.style={every pff stroke, draw=palB, line width=0.8pt,
                             opacity=0.85},
  pff/pasture/bg/.style={fill=gpwBg},
  pff/pasture/meadow/.style={every pff stroke, fill=gpwM, draw=gpwD!70,
                             line width=0.5pt},
  pff/pasture/dot/.style={fill=gpwD},
  pff/pasture/tuft/.style={every pff stroke, draw=gpwT, line width=0.8pt},
  pff/biomass/bg/.style={fill=agbBg},
  pff/biomass/crown a/.style={every pff stroke, fill=agbA, draw=agbO,
                              line width=0.7pt},
  pff/biomass/crown b/.style={every pff stroke, fill=agbB, draw=agbO,
                              line width=0.7pt},
  pff/water/bg/.style={fill=watBg},
  pff/water/lake/.style={every pff stroke, fill=watF, draw=watO,
                         line width=0.9pt},
  pff/water/river/.style={every pff stroke, draw=watF!80!watO,
                          line width=1.8pt},
  pff/water/ripple/.style={every pff stroke, draw=white, line width=1.0pt,
                           opacity=0.9},
  pff/dem/bg/.style={fill=demBg},
  pff/dem/contour/.style={every pff stroke, draw=demL, line width=0.6pt},
  pff/dem/index/.style={every pff stroke, draw=demI, line width=1.0pt},
  pff/canopy/bg/.style={fill=canBg},
  pff/canopy/ground/.style={every pff stroke, draw=canT, line width=1.0pt},
  pff/canopy/height/.style={every pff stroke, draw=canH, line width=0.6pt,
                            dash pattern=on 2.6pt off 2.0pt},
  pff/canopy/trunk/.style={every pff stroke, draw=canT, line width=1.2pt},
  pff/canopy/crown a/.style={every pff stroke, fill=canA, draw=canO,
                             line width=0.7pt},
  pff/canopy/crown b/.style={every pff stroke, fill=canB, draw=canO,
                             line width=0.7pt},
  pff/wetland/bg/.style={fill=wetBg},
  pff/wetland/band/.style={every pff stroke, fill=wetF, draw=wetO,
                           line width=0.8pt},
  pff/wetland/hummock/.style={every pff stroke, fill=wetF!60!wetO,
                              draw=wetO!85!black, line width=0.6pt},
  pff/wetland/reed/.style={every pff stroke, draw=wetR, line width=0.9pt},
  pff/wetland/ripple/.style={every pff stroke, draw=wetO!60, line width=0.7pt},
  pff/soil/bg/.style={fill=soilBg},
  pff/soil/h1/.style={fill=soilA},
  pff/soil/h2/.style={fill=soilB},
  pff/soil/h3/.style={fill=soilC},
  pff/soil/h4/.style={fill=soilD},
  pff/soil/line/.style={every pff stroke, draw=soilA!60!black, line width=0.6pt},
  pff/soil/pebble/.style={fill=soilP},
}

\newcommand{\archMLP}[5]{%
  \foreach \a in {1,...,#3}{\foreach \b in {1,...,#4}{
    \draw[pff/arch/net] ({#1-0.34},{#2+(\a-(#3+1)/2)*0.18})
      -- ({#1},{#2+(\b-(#4+1)/2)*0.18});}}
  \foreach \b in {1,...,#4}{\foreach \c in {1,...,#5}{
    \draw[pff/arch/net] ({#1},{#2+(\b-(#4+1)/2)*0.18})
      -- ({#1+0.34},{#2+(\c-(#5+1)/2)*0.18});}}
  \foreach \a in {1,...,#3}{
    \path[pff/arch/neuron] ({#1-0.34},{#2+(\a-(#3+1)/2)*0.18})
      circle[radius=0.05];}
  \foreach \b in {1,...,#4}{
    \path[pff/arch/neuron] ({#1},{#2+(\b-(#4+1)/2)*0.18})
      circle[radius=0.05];}
  \foreach \c in {1,...,#5}{
    \path[pff/arch/neuron] ({#1+0.34},{#2+(\c-(#5+1)/2)*0.18})
      circle[radius=0.05];}}


\newcommand{\tileAE}{%
\path[pff/ae/bg, pff/stackframe] (0.27,0.27) rectangle (1.27,1.27);
\path[pff/ae/bg, pff/stackframe] (0.18,0.18) rectangle (1.18,1.18);
\path[pff/ae/bg, pff/stackframe] (0.09,0.09) rectangle (1.09,1.09);
\begin{scope}
  \clip[pff/corners] (0,0) rectangle (1,1);
  \path[pff/ae/bg] (0,0) rectangle (1,1);
  \path[pff/ae/l1] plot[smooth cycle, tension=0.6] coordinates {(-0.003,0.005) (0.206,0.066) (0.455,0.005) (0.744,-0.001) (1.004,0.029) (0.971,0.252) (0.982,0.490) (1.003,0.651) (1.002,0.940) (0.773,1.001) (0.578,0.845) (0.577,0.601) (0.478,0.844) (0.445,0.999) (0.155,1.005) (-0.010,0.871) (-0.010,0.579) (-0.002,0.289)};
  \path[pff/ae/l1] plot[smooth cycle, tension=0.6] coordinates {(0.658,0.072) (0.632,0.096) (0.618,0.128) (0.648,0.143) (0.684,0.147) (0.700,0.178) (0.710,0.212) (0.736,0.242) (0.761,0.264) (0.788,0.264) (0.814,0.244) (0.836,0.217) (0.845,0.178) (0.819,0.144) (0.779,0.135) (0.741,0.129) (0.711,0.105) (0.687,0.077)};
  \path[pff/ae/l1] plot[smooth cycle, tension=0.6] coordinates {(0.330,0.188) (0.305,0.208) (0.288,0.235) (0.283,0.265) (0.296,0.294) (0.325,0.308) (0.352,0.289) (0.364,0.260) (0.366,0.228) (0.358,0.197)};
  \path[pff/ae/l2] plot[smooth cycle, tension=0.6] coordinates {(0.189,0.393) (0.158,0.404) (0.147,0.436) (0.135,0.465) (0.120,0.497) (0.132,0.530) (0.151,0.555) (0.176,0.576) (0.210,0.571) (0.236,0.553) (0.262,0.532) (0.270,0.499) (0.266,0.465) (0.273,0.433) (0.256,0.402) (0.221,0.396)};
  \path[pff/ae/l3] plot[smooth cycle, tension=0.6] coordinates {(0.489,-0.003) (0.542,0.074) (0.517,0.256) (0.489,0.419) (0.369,0.560) (0.352,0.749) (0.443,0.912) (0.327,1.005) (0.174,0.956) (0.031,0.927) (0.025,0.784) (-0.003,0.607) (0.068,0.499) (0.122,0.677) (0.194,0.657) (0.318,0.534) (0.398,0.367) (0.420,0.177)};
  \path[pff/ae/l3] plot[smooth cycle, tension=0.6] coordinates {(0.869,0.642) (0.946,0.652) (0.997,0.686) (1.004,0.766) (1.006,0.847) (1.005,0.928) (0.992,1.000) (0.930,0.958) (0.849,0.950) (0.771,0.952) (0.708,0.926) (0.650,0.868) (0.604,0.805) (0.604,0.730) (0.658,0.673) (0.737,0.656) (0.790,0.716) (0.839,0.707)};
  \path[pff/ae/l3] plot[smooth cycle, tension=0.6] coordinates {(0.975,0.316) (0.999,0.322) (1.003,0.355) (1.007,0.388) (0.999,0.422) (0.993,0.448) (0.967,0.452) (0.934,0.463) (0.901,0.469) (0.866,0.469) (0.837,0.450) (0.830,0.419) (0.834,0.392) (0.850,0.363) (0.879,0.345) (0.913,0.338) (0.944,0.327)};
  \path[pff/ae/l3] plot[smooth cycle, tension=0.6] coordinates {(0.817,0.001) (0.852,0.003) (0.887,0.002) (0.921,-0.000) (0.949,0.008) (0.951,0.043) (0.962,0.078) (0.948,0.112) (0.912,0.115) (0.881,0.104) (0.851,0.090) (0.820,0.071) (0.808,0.037)};
  \path[pff/ae/l3] plot[smooth cycle, tension=0.6] coordinates {(-0.001,0.057) (0.013,0.060) (0.024,0.082) (0.039,0.103) (0.039,0.128) (0.029,0.148) (0.006,0.147) (-0.002,0.130) (-0.000,0.105) (-0.003,0.080)};
  \path[pff/ae/l4] plot[smooth cycle, tension=0.6] coordinates {(0.732,0.004) (0.795,-0.005) (0.859,-0.004) (0.918,0.004) (0.910,0.064) (0.964,0.087) (0.996,0.121) (0.994,0.183) (1.004,0.247) (0.985,0.297) (0.922,0.316) (0.867,0.346) (0.804,0.351) (0.780,0.293) (0.791,0.231) (0.790,0.168) (0.746,0.124) (0.732,0.062)};
  \path[pff/ae/l4] plot[smooth cycle, tension=0.6] coordinates {(0.175,0.185) (0.248,0.195) (0.310,0.239) (0.333,0.300) (0.278,0.344) (0.205,0.327) (0.128,0.336) (0.102,0.407) (0.091,0.480) (0.070,0.552) (0.032,0.616) (0.000,0.602) (0.004,0.528) (0.003,0.453) (-0.005,0.379) (-0.004,0.301) (0.066,0.289) (0.120,0.238)};
  \path[pff/ae/l4] plot[smooth cycle, tension=0.6] coordinates {(0.306,-0.004) (0.356,-0.001) (0.404,0.007) (0.444,0.011) (0.465,0.055) (0.479,0.102) (0.501,0.147) (0.541,0.179) (0.569,0.221) (0.546,0.261) (0.504,0.274) (0.454,0.275) (0.408,0.256) (0.378,0.222) (0.357,0.180) (0.323,0.144) (0.305,0.098) (0.305,0.049)};
  \path[pff/ae/l4] plot[smooth cycle, tension=0.6] coordinates {(0.938,0.766) (0.962,0.764) (0.988,0.765) (1.000,0.780) (0.998,0.803) (0.983,0.816) (0.958,0.816) (0.934,0.810) (0.914,0.798) (0.914,0.775)};
  \path[pff/ae/l5] plot[smooth cycle, tension=0.6] coordinates {(0.303,0.886) (0.337,0.886) (0.360,0.912) (0.376,0.939) (0.373,0.972) (0.364,0.998) (0.332,1.005) (0.299,0.999) (0.266,0.999) (0.230,1.006) (0.219,0.984) (0.230,0.954) (0.248,0.926) (0.276,0.907)};
  \path[pff/ae/l5] plot[smooth cycle, tension=0.6] coordinates {(0.659,0.718) (0.690,0.705) (0.722,0.717) (0.754,0.735) (0.744,0.774) (0.735,0.799) (0.736,0.834) (0.707,0.852) (0.669,0.856) (0.649,0.823) (0.641,0.793) (0.617,0.767) (0.638,0.739)};
  \path[pff/ae/l5] plot[smooth cycle, tension=0.6] coordinates {(-0.001,0.540) (0.028,0.539) (0.064,0.557) (0.061,0.598) (0.062,0.631) (0.049,0.659) (0.028,0.691) (-0.001,0.677) (-0.004,0.641) (-0.000,0.606) (0.005,0.571)};
  \path[pff/ae/l5] plot[smooth cycle, tension=0.6] coordinates {(0.256,0.674) (0.282,0.662) (0.300,0.680) (0.305,0.705) (0.293,0.728) (0.281,0.755) (0.253,0.750) (0.234,0.736) (0.229,0.712) (0.230,0.685)};
  \path[pff/ae/flow] plot[smooth, tension=0.8] coordinates {(0.404,0.000) (0.432,0.111) (0.508,0.197) (0.605,0.258) (0.698,0.325) (0.733,0.431) (0.682,0.532) (0.599,0.612) (0.514,0.689) (0.444,0.781) (0.399,0.886) (0.383,1.000)};
  \path[pff/ae/flow] plot[smooth, tension=0.8] coordinates {(0.208,0.000) (0.208,0.044) (0.207,0.089) (0.205,0.133) (0.200,0.177) (0.190,0.221) (0.173,0.262) (0.150,0.299) (0.119,0.331) (0.083,0.358) (0.043,0.377) (0.000,0.386)};
  \path[pff/ae/flow] plot[smooth, tension=0.8] coordinates {(0.000,0.486) (0.063,0.495) (0.124,0.515) (0.182,0.540) (0.236,0.573) (0.277,0.622) (0.295,0.682) (0.298,0.746) (0.294,0.809) (0.289,0.873) (0.284,0.936) (0.282,1.000)};
\end{scope}
\path[pff/frame] (0,0) rectangle (1,1);
}

\newcommand{\tileTE}{%
\path[pff/te/bg, pff/stackframe] (0.27,0.27) rectangle (1.27,1.27);
\path[pff/te/bg, pff/stackframe] (0.18,0.18) rectangle (1.18,1.18);
\path[pff/te/bg, pff/stackframe] (0.09,0.09) rectangle (1.09,1.09);
\begin{scope}
  \clip[pff/corners] (0,0) rectangle (1,1);
  \path[pff/te/bg] (0,0) rectangle (1,1);
  \path[pff/te/cell b] plot[smooth cycle, tension=0.3] coordinates {(0.652,0.614) (0.688,0.590) (0.729,0.578) (0.767,0.595) (0.792,0.630) (0.803,0.671) (0.790,0.711) (0.765,0.747) (0.731,0.742) (0.695,0.718) (0.662,0.690) (0.640,0.654)};
  \path[pff/te/cell b] plot[smooth cycle, tension=0.3] coordinates {(0.119,-0.001) (0.147,0.026) (0.166,0.057) (0.139,0.083) (0.107,0.106) (0.075,0.128) (0.045,0.123) (0.018,0.094) (0.010,0.058) (0.024,0.023) (0.049,-0.006) (0.084,-0.017)};
  \path[pff/te/cell d] plot[smooth cycle, tension=0.3] coordinates {(0.649,0.860) (0.681,0.852) (0.711,0.858) (0.733,0.882) (0.742,0.915) (0.745,0.948) (0.730,0.973) (0.697,0.976) (0.666,0.965) (0.638,0.947) (0.623,0.918) (0.625,0.885)};
  \path[pff/te/cell c] plot[smooth cycle, tension=0.3] coordinates {(0.096,0.136) (0.125,0.117) (0.154,0.101) (0.187,0.108) (0.203,0.138) (0.209,0.171) (0.199,0.202) (0.176,0.227) (0.146,0.243) (0.117,0.228) (0.098,0.200) (0.086,0.169)};
  \path[pff/te/cell d] plot[smooth cycle, tension=0.3] coordinates {(0.686,0.410) (0.716,0.389) (0.753,0.401) (0.785,0.425) (0.813,0.452) (0.821,0.488) (0.800,0.521) (0.771,0.547) (0.733,0.551) (0.708,0.524) (0.694,0.487) (0.684,0.449)};
  \path[pff/te/cell c] plot[smooth cycle, tension=0.3] coordinates {(0.337,0.279) (0.353,0.243) (0.374,0.217) (0.406,0.239) (0.438,0.262) (0.456,0.293) (0.453,0.330) (0.433,0.363) (0.405,0.389) (0.370,0.385) (0.347,0.355) (0.335,0.318)};
  \path[pff/te/cell a] plot[smooth cycle, tension=0.3] coordinates {(0.032,0.360) (0.057,0.377) (0.066,0.406) (0.059,0.437) (0.046,0.466) (0.027,0.492) (0.000,0.506) (-0.018,0.488) (-0.023,0.457) (-0.023,0.425) (-0.016,0.394) (0.002,0.370)};
  \path[pff/te/cell c] plot[smooth cycle, tension=0.3] coordinates {(0.251,0.324) (0.284,0.317) (0.314,0.329) (0.328,0.360) (0.320,0.391) (0.299,0.416) (0.270,0.435) (0.239,0.449) (0.211,0.444) (0.204,0.412) (0.213,0.379) (0.226,0.348)};
  \path[pff/te/cell c] plot[smooth cycle, tension=0.3] coordinates {(0.531,0.631) (0.571,0.621) (0.605,0.640) (0.619,0.678) (0.605,0.717) (0.581,0.751) (0.553,0.781) (0.517,0.796) (0.480,0.782) (0.477,0.742) (0.490,0.703) (0.507,0.665)};
  \path[pff/te/cell c] plot[smooth cycle, tension=0.3] coordinates {(0.665,0.722) (0.695,0.742) (0.722,0.764) (0.734,0.795) (0.710,0.820) (0.678,0.834) (0.643,0.842) (0.607,0.838) (0.580,0.816) (0.585,0.781) (0.606,0.752) (0.630,0.728)};
  \path[pff/te/cell c] plot[smooth cycle, tension=0.3] coordinates {(-0.015,0.646) (-0.000,0.658) (0.011,0.672) (0.011,0.691) (0.006,0.709) (-0.003,0.726) (-0.012,0.742) (-0.025,0.737) (-0.027,0.718) (-0.027,0.699) (-0.027,0.680) (-0.024,0.662)};
  \path[pff/te/cell c] plot[smooth cycle, tension=0.3] coordinates {(0.459,0.856) (0.487,0.828) (0.526,0.823) (0.564,0.834) (0.593,0.862) (0.602,0.900) (0.592,0.938) (0.565,0.966) (0.529,0.984) (0.496,0.969) (0.478,0.933) (0.463,0.896)};
  \path[pff/te/cell c] plot[smooth cycle, tension=0.3] coordinates {(1.024,-0.018) (1.023,-0.001) (1.009,0.010) (0.991,0.013) (0.974,0.012) (0.957,0.006) (0.944,-0.007) (0.938,-0.019) (0.953,-0.027) (0.971,-0.027) (0.989,-0.027) (1.007,-0.024)};
  \path[pff/te/cell e] plot[smooth cycle, tension=0.3] coordinates {(0.807,0.150) (0.847,0.149) (0.888,0.149) (0.928,0.154) (0.961,0.174) (0.974,0.209) (0.943,0.236) (0.909,0.257) (0.872,0.273) (0.834,0.267) (0.818,0.230) (0.812,0.190)};
  \path[pff/te/cell b] plot[smooth cycle, tension=0.3] coordinates {(0.839,0.963) (0.866,0.945) (0.894,0.928) (0.926,0.920) (0.950,0.942) (0.959,0.973) (0.958,1.004) (0.930,1.021) (0.898,1.023) (0.865,1.023) (0.834,1.014) (0.819,0.989)};
  \path[pff/te/cell a] plot[smooth cycle, tension=0.3] coordinates {(0.391,0.183) (0.406,0.146) (0.443,0.130) (0.483,0.132) (0.519,0.149) (0.545,0.179) (0.558,0.215) (0.543,0.249) (0.511,0.274) (0.476,0.265) (0.442,0.242) (0.410,0.218)};
  \path[pff/te/cell d] plot[smooth cycle, tension=0.3] coordinates {(0.114,0.418) (0.151,0.430) (0.186,0.448) (0.210,0.480) (0.194,0.515) (0.174,0.549) (0.148,0.574) (0.113,0.570) (0.084,0.544) (0.066,0.511) (0.064,0.473) (0.080,0.438)};
  \path[pff/te/cell d] plot[smooth cycle, tension=0.3] coordinates {(0.641,0.033) (0.671,0.013) (0.709,0.012) (0.747,0.016) (0.776,0.039) (0.785,0.073) (0.785,0.112) (0.767,0.135) (0.729,0.140) (0.695,0.129) (0.668,0.103) (0.649,0.070)};
  \path[pff/te/cell c] plot[smooth cycle, tension=0.3] coordinates {(0.982,0.480) (0.985,0.456) (0.987,0.432) (0.996,0.410) (1.012,0.393) (1.023,0.406) (1.027,0.430) (1.027,0.454) (1.027,0.478) (1.021,0.501) (1.009,0.515) (0.990,0.503)};
  \path[pff/te/cell e] plot[smooth cycle, tension=0.3] coordinates {(0.856,0.683) (0.895,0.683) (0.924,0.709) (0.938,0.745) (0.944,0.783) (0.932,0.819) (0.901,0.831) (0.865,0.819) (0.832,0.799) (0.807,0.771) (0.802,0.737) (0.823,0.704)};
  \path[pff/te/cell b] plot[smooth cycle, tension=0.3] coordinates {(0.750,0.331) (0.779,0.310) (0.810,0.294) (0.844,0.297) (0.867,0.322) (0.880,0.355) (0.873,0.390) (0.848,0.415) (0.814,0.416) (0.783,0.401) (0.755,0.380) (0.728,0.356)};
  \path[pff/te/cell d] plot[smooth cycle, tension=0.3] coordinates {(0.990,0.218) (1.002,0.201) (1.020,0.201) (1.025,0.222) (1.027,0.245) (1.027,0.267) (1.027,0.289) (1.025,0.311) (1.014,0.304) (1.007,0.283) (1.000,0.262) (0.995,0.240)};
  \path[pff/te/cell d] plot[smooth cycle, tension=0.3] coordinates {(0.547,0.281) (0.578,0.254) (0.614,0.239) (0.646,0.261) (0.670,0.293) (0.693,0.327) (0.695,0.361) (0.667,0.390) (0.628,0.401) (0.593,0.386) (0.568,0.356) (0.547,0.321)};
  \path[pff/te/cell c] plot[smooth cycle, tension=0.3] coordinates {(0.957,0.534) (0.983,0.525) (1.009,0.533) (1.021,0.558) (1.023,0.585) (1.023,0.613) (1.007,0.635) (0.982,0.642) (0.957,0.639) (0.943,0.616) (0.946,0.588) (0.950,0.560)};
  \path[pff/te/cell d] plot[smooth cycle, tension=0.3] coordinates {(0.519,-0.004) (0.549,0.017) (0.563,0.049) (0.545,0.080) (0.515,0.101) (0.479,0.110) (0.442,0.108) (0.409,0.095) (0.414,0.060) (0.432,0.027) (0.453,-0.003) (0.484,-0.019)};
  \path[pff/te/cell e] plot[smooth cycle, tension=0.3] coordinates {(0.054,0.683) (0.098,0.669) (0.133,0.696) (0.143,0.740) (0.134,0.781) (0.104,0.812) (0.061,0.829) (0.018,0.844) (-0.017,0.839) (-0.019,0.794) (-0.003,0.751) (0.019,0.712)};
  \path[pff/te/cell b] plot[smooth cycle, tension=0.3] coordinates {(0.707,1.018) (0.716,1.000) (0.732,0.988) (0.750,0.980) (0.770,0.977) (0.788,0.983) (0.800,0.999) (0.804,1.017) (0.786,1.025) (0.766,1.026) (0.746,1.026) (0.726,1.024)};
  \path[pff/te/cell a] plot[smooth cycle, tension=0.3] coordinates {(0.123,0.988) (0.145,0.991) (0.168,0.994) (0.189,1.001) (0.196,1.019) (0.175,1.025) (0.153,1.027) (0.130,1.027) (0.108,1.027) (0.086,1.020) (0.084,1.006) (0.101,0.993)};
  \path[pff/te/cell d] plot[smooth cycle, tension=0.3] coordinates {(-0.003,0.211) (0.025,0.187) (0.052,0.164) (0.072,0.193) (0.091,0.225) (0.093,0.260) (0.081,0.292) (0.060,0.322) (0.032,0.334) (0.010,0.306) (-0.009,0.275) (-0.021,0.241)};
  \path[pff/te/cell e] plot[smooth cycle, tension=0.3] coordinates {(-0.019,0.875) (0.006,0.894) (0.029,0.918) (0.050,0.942) (0.066,0.970) (0.067,0.999) (0.041,1.016) (0.009,1.019) (-0.013,1.002) (-0.024,0.972) (-0.024,0.940) (-0.023,0.907)};
  \path[pff/te/cell a] plot[smooth cycle, tension=0.3] coordinates {(0.022,-0.001) (0.013,0.016) (0.003,0.032) (-0.008,0.048) (-0.019,0.064) (-0.026,0.053) (-0.027,0.034) (-0.027,0.015) (-0.025,-0.004) (-0.014,-0.020) (0.004,-0.023) (0.020,-0.020)};
  \path[pff/te/cell d] plot[smooth cycle, tension=0.3] coordinates {(0.259,0.518) (0.296,0.539) (0.332,0.560) (0.352,0.588) (0.335,0.627) (0.299,0.643) (0.258,0.642) (0.216,0.634) (0.175,0.624) (0.173,0.595) (0.193,0.558) (0.218,0.526)};
  \path[pff/te/cell d] plot[smooth cycle, tension=0.3] coordinates {(0.799,0.558) (0.824,0.530) (0.860,0.520) (0.896,0.529) (0.922,0.551) (0.931,0.586) (0.927,0.623) (0.915,0.657) (0.878,0.661) (0.843,0.653) (0.817,0.628) (0.798,0.596)};
  \path[pff/te/cell a] plot[smooth cycle, tension=0.3] coordinates {(0.746,0.822) (0.767,0.786) (0.804,0.804) (0.840,0.827) (0.876,0.851) (0.903,0.881) (0.879,0.913) (0.843,0.937) (0.806,0.957) (0.773,0.946) (0.758,0.906) (0.747,0.865)};
  \path[pff/te/cell d] plot[smooth cycle, tension=0.3] coordinates {(0.814,0.006) (0.845,-0.017) (0.885,-0.020) (0.922,-0.006) (0.944,0.026) (0.947,0.066) (0.938,0.104) (0.905,0.126) (0.864,0.126) (0.828,0.117) (0.806,0.086) (0.807,0.046)};
  \path[pff/te/cell b] plot[smooth cycle, tension=0.3] coordinates {(0.195,0.068) (0.227,0.039) (0.269,0.029) (0.311,0.043) (0.349,0.066) (0.382,0.097) (0.376,0.140) (0.359,0.182) (0.319,0.182) (0.275,0.172) (0.238,0.147) (0.214,0.109)};
  \path[pff/te/cell a] plot[smooth cycle, tension=0.3] coordinates {(0.809,-0.019) (0.791,-0.001) (0.766,0.003) (0.740,0.002) (0.715,-0.001) (0.689,-0.005) (0.666,-0.013) (0.681,-0.025) (0.707,-0.028) (0.733,-0.028) (0.758,-0.028) (0.784,-0.026)};
  \path[pff/te/cell b] plot[smooth cycle, tension=0.3] coordinates {(1.017,0.178) (0.988,0.173) (0.968,0.151) (0.961,0.121) (0.962,0.090) (0.966,0.060) (0.982,0.033) (1.011,0.026) (1.022,0.055) (1.025,0.086) (1.025,0.117) (1.023,0.148)};
  \path[pff/te/cell c] plot[smooth cycle, tension=0.3] coordinates {(0.557,0.994) (0.582,0.978) (0.608,0.968) (0.637,0.968) (0.664,0.980) (0.687,0.997) (0.683,1.019) (0.654,1.025) (0.625,1.026) (0.596,1.026) (0.567,1.024) (0.542,1.016)};
  \path[pff/te/cell a] plot[smooth cycle, tension=0.3] coordinates {(0.408,0.597) (0.441,0.609) (0.471,0.623) (0.486,0.655) (0.475,0.687) (0.459,0.716) (0.425,0.727) (0.395,0.709) (0.366,0.690) (0.344,0.666) (0.356,0.633) (0.373,0.605)};
  \path[pff/te/cell b] plot[smooth cycle, tension=0.3] coordinates {(0.151,0.650) (0.190,0.648) (0.233,0.655) (0.275,0.664) (0.317,0.675) (0.306,0.713) (0.285,0.750) (0.256,0.781) (0.215,0.793) (0.182,0.775) (0.169,0.734) (0.159,0.692)};
  \path[pff/te/cell d] plot[smooth cycle, tension=0.3] coordinates {(0.656,0.210) (0.680,0.180) (0.715,0.162) (0.753,0.158) (0.784,0.178) (0.793,0.214) (0.793,0.251) (0.777,0.286) (0.746,0.309) (0.710,0.306) (0.684,0.280) (0.663,0.248)};
  \path[pff/te/cell a] plot[smooth cycle, tension=0.3] coordinates {(0.213,0.996) (0.234,0.991) (0.256,0.988) (0.278,0.989) (0.299,0.994) (0.317,1.008) (0.319,1.020) (0.298,1.027) (0.276,1.027) (0.254,1.027) (0.232,1.024) (0.212,1.018)};
  \path[pff/te/cell c] plot[smooth cycle, tension=0.3] coordinates {(0.975,1.004) (0.974,0.985) (0.974,0.966) (0.983,0.950) (0.999,0.944) (1.016,0.946) (1.026,0.962) (1.026,0.981) (1.026,1.000) (1.017,1.017) (1.002,1.023) (0.985,1.020)};
  \path[pff/te/cell e] plot[smooth cycle, tension=0.3] coordinates {(0.132,0.265) (0.165,0.273) (0.191,0.297) (0.203,0.329) (0.198,0.364) (0.184,0.396) (0.152,0.410) (0.118,0.399) (0.087,0.382) (0.073,0.352) (0.085,0.319) (0.106,0.289)};
  \path[pff/te/cell d] plot[smooth cycle, tension=0.3] coordinates {(0.052,0.857) (0.094,0.840) (0.136,0.824) (0.179,0.817) (0.190,0.859) (0.194,0.904) (0.198,0.949) (0.186,0.981) (0.141,0.976) (0.100,0.960) (0.065,0.932) (0.035,0.898)};
  \path[pff/te/cell e] plot[smooth cycle, tension=0.3] coordinates {(0.303,0.771) (0.322,0.737) (0.351,0.717) (0.386,0.723) (0.418,0.744) (0.447,0.770) (0.458,0.805) (0.442,0.839) (0.409,0.856) (0.372,0.852) (0.339,0.833) (0.310,0.808)};
  \path[pff/te/cell b] plot[smooth cycle, tension=0.3] coordinates {(-0.004,0.534) (0.031,0.534) (0.062,0.552) (0.089,0.576) (0.116,0.601) (0.130,0.630) (0.099,0.646) (0.064,0.657) (0.028,0.656) (-0.001,0.636) (-0.016,0.603) (-0.017,0.567)};
  \path[pff/te/cell e] plot[smooth cycle, tension=0.3] coordinates {(0.981,0.800) (1.009,0.801) (1.022,0.829) (1.022,0.860) (1.022,0.891) (1.011,0.919) (0.985,0.929) (0.956,0.920) (0.934,0.900) (0.926,0.871) (0.939,0.844) (0.956,0.818)};
  \path[pff/te/cell a] plot[smooth cycle, tension=0.3] coordinates {(0.266,0.191) (0.299,0.198) (0.327,0.213) (0.333,0.242) (0.319,0.273) (0.293,0.295) (0.260,0.303) (0.228,0.298) (0.202,0.278) (0.193,0.248) (0.207,0.219) (0.233,0.199)};
  \path[pff/te/cell c] plot[smooth cycle, tension=0.3] coordinates {(0.457,0.473) (0.495,0.480) (0.523,0.506) (0.536,0.541) (0.542,0.579) (0.525,0.608) (0.487,0.605) (0.449,0.593) (0.413,0.581) (0.388,0.562) (0.406,0.527) (0.424,0.493)};
  \path[pff/te/cell d] plot[smooth cycle, tension=0.3] coordinates {(0.555,0.467) (0.585,0.439) (0.623,0.427) (0.654,0.445) (0.671,0.482) (0.683,0.521) (0.675,0.560) (0.646,0.586) (0.608,0.601) (0.576,0.583) (0.560,0.546) (0.549,0.507)};
  \path[pff/te/cell e] plot[smooth cycle, tension=0.3] coordinates {(0.929,0.673) (0.955,0.663) (0.984,0.661) (1.008,0.669) (1.022,0.694) (1.022,0.723) (1.020,0.751) (1.007,0.776) (0.979,0.775) (0.959,0.755) (0.947,0.728) (0.936,0.701)};
  \path[pff/te/cell d] plot[smooth cycle, tension=0.3] coordinates {(0.259,-0.019) (0.249,0.001) (0.229,0.015) (0.208,0.026) (0.185,0.025) (0.163,0.017) (0.146,-0.000) (0.143,-0.017) (0.163,-0.026) (0.187,-0.026) (0.212,-0.026) (0.235,-0.024)};
  \path[pff/te/cell d] plot[smooth cycle, tension=0.3] coordinates {(0.362,0.408) (0.396,0.426) (0.411,0.458) (0.398,0.494) (0.381,0.529) (0.344,0.537) (0.309,0.524) (0.276,0.504) (0.243,0.484) (0.259,0.460) (0.291,0.439) (0.324,0.419)};
  \path[pff/te/cell a] plot[smooth cycle, tension=0.3] coordinates {(0.427,-0.002) (0.411,0.025) (0.392,0.046) (0.362,0.051) (0.335,0.039) (0.309,0.023) (0.285,0.004) (0.283,-0.019) (0.313,-0.024) (0.344,-0.025) (0.375,-0.025) (0.404,-0.022)};
  \path[pff/te/cell c] plot[smooth cycle, tension=0.3] coordinates {(-0.024,0.081) (-0.006,0.093) (0.010,0.109) (0.026,0.126) (0.029,0.148) (0.022,0.168) (0.004,0.183) (-0.013,0.190) (-0.027,0.173) (-0.027,0.150) (-0.027,0.127) (-0.026,0.104)};
  \path[pff/te/cell c] plot[smooth cycle, tension=0.3] coordinates {(0.390,0.882) (0.425,0.884) (0.447,0.913) (0.461,0.947) (0.459,0.983) (0.433,1.006) (0.399,1.020) (0.363,1.016) (0.331,0.997) (0.322,0.966) (0.338,0.934) (0.362,0.905)};
  \path[pff/te/cell d] plot[smooth cycle, tension=0.3] coordinates {(0.222,0.817) (0.261,0.813) (0.297,0.827) (0.328,0.851) (0.340,0.886) (0.324,0.920) (0.299,0.949) (0.266,0.971) (0.231,0.968) (0.214,0.933) (0.211,0.894) (0.210,0.854)};
  \path[pff/te/cell c] plot[smooth cycle, tension=0.3] coordinates {(-0.015,0.288) (-0.007,0.301) (0.001,0.314) (0.010,0.327) (0.010,0.342) (0.004,0.354) (-0.010,0.361) (-0.020,0.355) (-0.027,0.341) (-0.027,0.326) (-0.027,0.311) (-0.027,0.295)};
  \path[pff/te/cell d] plot[smooth cycle, tension=0.3] coordinates {(0.647,-0.020) (0.642,0.001) (0.627,0.018) (0.609,0.031) (0.589,0.025) (0.569,0.013) (0.551,-0.000) (0.539,-0.014) (0.557,-0.026) (0.579,-0.026) (0.602,-0.026) (0.625,-0.024)};
  \path[pff/te/cell e] plot[smooth cycle, tension=0.3] coordinates {(0.498,0.303) (0.525,0.325) (0.544,0.357) (0.561,0.390) (0.562,0.426) (0.535,0.450) (0.499,0.457) (0.462,0.450) (0.432,0.431) (0.437,0.396) (0.455,0.364) (0.475,0.332)};
  \path[pff/te/cell c] plot[smooth cycle, tension=0.3] coordinates {(0.440,1.018) (0.454,1.010) (0.468,1.001) (0.483,0.996) (0.499,0.995) (0.513,1.003) (0.523,1.015) (0.517,1.024) (0.501,1.028) (0.485,1.028) (0.469,1.028) (0.453,1.028)};
  \path[pff/te/cell c] plot[smooth cycle, tension=0.3] coordinates {(0.551,0.105) (0.574,0.078) (0.599,0.053) (0.625,0.066) (0.644,0.096) (0.661,0.127) (0.663,0.163) (0.644,0.191) (0.616,0.212) (0.586,0.200) (0.565,0.172) (0.549,0.141)};
  \path[pff/te/cell c] plot[smooth cycle, tension=0.3] coordinates {(0.858,0.435) (0.882,0.411) (0.912,0.398) (0.945,0.401) (0.968,0.421) (0.972,0.453) (0.968,0.486) (0.953,0.514) (0.920,0.515) (0.888,0.505) (0.858,0.491) (0.844,0.464)};
  \path[pff/te/cell b] plot[smooth cycle, tension=0.3] coordinates {(0.907,0.285) (0.936,0.267) (0.966,0.257) (0.994,0.274) (1.004,0.306) (1.014,0.339) (1.016,0.370) (0.987,0.386) (0.953,0.384) (0.922,0.372) (0.900,0.347) (0.891,0.315)};
\end{scope}
\path[pff/frame] (0,0) rectangle (1,1);
}

\newcommand{\tileSTwo}{%
\path[pff/stwo/bg, pff/stackframe] (0.18,0.18) rectangle (1.18,1.18);
\path[pff/stwo/bg, pff/stackframe] (0.09,0.09) rectangle (1.09,1.09);
\begin{scope}
  \clip[pff/corners] (0,0) rectangle (1,1);
  \path[pff/stwo/bg] (0,0) rectangle (1,1);
  \path[pff/stwo/patch a] plot[smooth cycle, tension=0.15] coordinates {(-0.011,-0.063) (0.058,-0.057) (0.127,-0.051) (0.196,-0.045) (0.265,-0.039) (0.317,0.000) (0.319,0.069) (0.316,0.138) (0.314,0.208) (0.312,0.277) (0.264,0.320) (0.195,0.314) (0.126,0.309) (0.056,0.303) (-0.013,0.297) (-0.062,0.256) (-0.064,0.187) (-0.062,0.117) (-0.061,0.048) (-0.060,-0.022)};
  \path[pff/stwo/patch b] plot[smooth cycle, tension=0.15] coordinates {(0.369,-0.033) (0.429,-0.034) (0.489,-0.036) (0.549,-0.037) (0.609,-0.038) (0.652,-0.004) (0.655,0.056) (0.657,0.116) (0.658,0.176) (0.658,0.236) (0.615,0.270) (0.555,0.273) (0.495,0.276) (0.435,0.279) (0.375,0.282) (0.332,0.246) (0.331,0.186) (0.330,0.126) (0.329,0.066) (0.329,0.006)};
  \path[pff/stwo/patch y] plot[smooth cycle, tension=0.15] coordinates {(0.750,-0.016) (0.812,-0.018) (0.873,-0.020) (0.935,-0.022) (0.996,-0.024) (1.046,0.005) (1.050,0.065) (1.050,0.127) (1.051,0.189) (1.051,0.250) (1.015,0.294) (0.954,0.297) (0.892,0.299) (0.831,0.301) (0.769,0.303) (0.723,0.270) (0.718,0.209) (0.715,0.147) (0.712,0.086) (0.710,0.024)};
  \path[pff/stwo/patch c] plot[smooth cycle, tension=0.15] coordinates {(-0.025,0.341) (0.040,0.339) (0.105,0.337) (0.170,0.334) (0.235,0.332) (0.290,0.358) (0.299,0.421) (0.303,0.486) (0.306,0.551) (0.309,0.616) (0.263,0.654) (0.198,0.655) (0.133,0.656) (0.068,0.657) (0.003,0.658) (-0.057,0.640) (-0.067,0.577) (-0.068,0.512) (-0.069,0.447) (-0.070,0.382)};
  \path[pff/stwo/patch b] plot[smooth cycle, tension=0.15] coordinates {(0.357,0.333) (0.423,0.333) (0.489,0.333) (0.555,0.333) (0.620,0.333) (0.675,0.361) (0.681,0.426) (0.682,0.492) (0.683,0.558) (0.684,0.623) (0.637,0.663) (0.571,0.663) (0.505,0.663) (0.439,0.663) (0.373,0.663) (0.318,0.636) (0.312,0.572) (0.311,0.506) (0.311,0.440) (0.311,0.374)};
  \path[pff/stwo/patch a] plot[smooth cycle, tension=0.15] coordinates {(0.749,0.330) (0.811,0.331) (0.873,0.332) (0.935,0.333) (0.997,0.334) (1.039,0.372) (1.040,0.434) (1.041,0.496) (1.041,0.558) (1.040,0.620) (0.997,0.657) (0.935,0.656) (0.873,0.655) (0.811,0.655) (0.749,0.654) (0.705,0.617) (0.705,0.555) (0.705,0.493) (0.706,0.431) (0.707,0.370)};
  \path[pff/stwo/patch c] plot[smooth cycle, tension=0.15] coordinates {(-0.011,0.700) (0.055,0.698) (0.120,0.697) (0.186,0.695) (0.251,0.693) (0.306,0.720) (0.310,0.785) (0.309,0.851) (0.308,0.916) (0.307,0.982) (0.266,1.026) (0.201,1.028) (0.135,1.029) (0.070,1.029) (0.004,1.030) (-0.050,1.002) (-0.055,0.937) (-0.056,0.872) (-0.056,0.806) (-0.056,0.741)};
  \path[pff/stwo/patch t] plot[smooth cycle, tension=0.15] coordinates {(0.364,0.698) (0.431,0.701) (0.497,0.705) (0.564,0.709) (0.631,0.712) (0.687,0.742) (0.691,0.808) (0.689,0.875) (0.687,0.942) (0.683,1.008) (0.630,1.040) (0.563,1.038) (0.497,1.036) (0.430,1.034) (0.363,1.032) (0.308,1.003) (0.306,0.937) (0.309,0.870) (0.312,0.803) (0.315,0.736)};
  \path[pff/stwo/patch b] plot[smooth cycle, tension=0.15] coordinates {(0.727,0.707) (0.792,0.704) (0.858,0.701) (0.924,0.698) (0.989,0.696) (1.031,0.740) (1.034,0.805) (1.036,0.871) (1.038,0.937) (1.041,1.003) (1.002,1.049) (0.936,1.053) (0.870,1.056) (0.805,1.059) (0.739,1.060) (0.699,1.015) (0.696,0.949) (0.692,0.883) (0.689,0.818) (0.686,0.752)};
  \path[pff/stwo/cloud shadow] plot[smooth cycle, tension=0.7] coordinates {(0.846,0.710) (0.825,0.750) (0.788,0.775) (0.745,0.791) (0.701,0.800) (0.655,0.802) (0.610,0.796) (0.568,0.780) (0.532,0.753) (0.523,0.711) (0.555,0.679) (0.593,0.654) (0.632,0.631) (0.675,0.617) (0.720,0.614) (0.765,0.620) (0.807,0.636) (0.840,0.666)};
  \path[pff/stwo/cloud] plot[smooth cycle, tension=0.7] coordinates {(0.796,0.800) (0.778,0.836) (0.746,0.861) (0.708,0.877) (0.669,0.887) (0.628,0.892) (0.587,0.891) (0.547,0.882) (0.510,0.866) (0.480,0.839) (0.473,0.801) (0.501,0.771) (0.535,0.749) (0.570,0.727) (0.607,0.711) (0.647,0.704) (0.688,0.705) (0.728,0.713) (0.765,0.730) (0.792,0.760)};
\end{scope}
\path[pff/frame] (0,0) rectangle (1,1);
}

\newcommand{\tileLandsat}{%
\path[pff/landsat/bg, pff/stackframe] (0.18,0.18) rectangle (1.18,1.18);
\path[pff/landsat/bg, pff/stackframe] (0.09,0.09) rectangle (1.09,1.09);
\begin{scope}
  \clip[pff/corners] (0,0) rectangle (1,1);
  \path[pff/landsat/bg] (0,0) rectangle (1,1);
  \path[pff/landsat/patch a] plot[smooth cycle, tension=0.7] coordinates {(-0.001,0.004) (0.259,-0.002) (0.519,-0.002) (0.779,-0.010) (1.003,0.034) (0.991,0.294) (1.008,0.554) (1.013,0.825) (0.775,0.802) (0.546,0.863) (0.343,0.995) (0.089,0.996) (-0.010,0.831) (0.041,0.619) (0.134,0.395) (0.158,0.168)};
  \path[pff/landsat/patch b] plot[smooth cycle, tension=0.7] coordinates {(0.167,0.004) (0.382,-0.002) (0.598,-0.000) (0.814,-0.009) (1.002,0.027) (0.989,0.242) (1.006,0.458) (1.017,0.679) (0.895,0.771) (0.708,0.681) (0.734,0.540) (0.943,0.500) (0.748,0.424) (0.650,0.253) (0.547,0.157) (0.363,0.060)};
  \path[pff/landsat/patch b] plot[smooth cycle, tension=0.7] coordinates {(0.355,0.310) (0.428,0.314) (0.495,0.342) (0.551,0.385) (0.593,0.442) (0.607,0.517) (0.557,0.574) (0.492,0.610) (0.438,0.661) (0.380,0.705) (0.314,0.670) (0.256,0.626) (0.231,0.559) (0.238,0.488) (0.253,0.416) (0.284,0.348)};
  \path[pff/landsat/patch b] plot[smooth cycle, tension=0.7] coordinates {(0.001,0.680) (0.071,0.688) (0.125,0.746) (0.188,0.779) (0.267,0.788) (0.328,0.832) (0.330,0.905) (0.330,0.980) (0.272,0.988) (0.201,0.992) (0.128,1.025) (0.054,1.013) (0.013,0.976) (0.003,0.914) (-0.006,0.841) (0.001,0.768)};
  \path[pff/landsat/patch c] plot[smooth cycle, tension=0.7] coordinates {(0.761,0.004) (0.825,-0.001) (0.888,0.000) (0.953,-0.008) (1.001,0.014) (0.989,0.077) (0.994,0.136) (0.961,0.190) (0.919,0.238) (0.886,0.289) (0.840,0.320) (0.787,0.301) (0.745,0.255) (0.724,0.190) (0.748,0.126) (0.772,0.065)};
  \path[pff/landsat/patch c] plot[smooth cycle, tension=0.7] coordinates {(-0.000,0.741) (0.049,0.752) (0.097,0.784) (0.143,0.822) (0.196,0.850) (0.250,0.881) (0.276,0.937) (0.277,0.996) (0.222,1.002) (0.162,1.005) (0.103,1.001) (0.043,1.000) (0.002,0.984) (0.005,0.926) (-0.004,0.867) (-0.014,0.807)};
  \path[pff/landsat/swath] plot[smooth, tension=0.8] coordinates {(-0.210,-0.100) (0.140,0.500) (0.460,1.100)};
  \path[pff/landsat/swath] plot[smooth, tension=0.8] coordinates {(0.550,-0.100) (0.900,0.500) (1.220,1.100)};
\end{scope}
\path[pff/frame] (0,0) rectangle (1,1);
}

\newcommand{\tileLST}{%
\path[pff/lst/bg, pff/stackframe] (0.18,0.18) rectangle (1.18,1.18);
\path[pff/lst/bg, pff/stackframe] (0.09,0.09) rectangle (1.09,1.09);
\begin{scope}
  \clip[pff/corners] (0,0) rectangle (1,1);
  \path[pff/lst/bg] (0,0) rectangle (1,1);
  \path[pff/lst/iso a] plot[smooth cycle, tension=0.7] coordinates {(0.800,0.003) (0.918,0.086) (0.986,0.222) (1.001,0.376) (1.007,0.533) (0.994,0.689) (0.946,0.790) (0.794,0.803) (0.639,0.822) (0.482,0.834) (0.327,0.839) (0.175,0.802) (0.070,0.683) (0.007,0.539) (0.014,0.388) (0.070,0.275) (0.221,0.230) (0.379,0.227) (0.533,0.194) (0.656,0.093)};
  \path[pff/lst/iso b] plot[smooth cycle, tension=0.7] coordinates {(0.759,0.136) (0.849,0.196) (0.895,0.294) (0.925,0.400) (0.894,0.508) (0.812,0.586) (0.725,0.656) (0.636,0.726) (0.531,0.769) (0.416,0.769) (0.309,0.747) (0.207,0.700) (0.139,0.608) (0.119,0.497) (0.146,0.399) (0.225,0.327) (0.332,0.289) (0.445,0.283) (0.553,0.249) (0.643,0.179)};
  \path[pff/lst/iso c] plot[smooth cycle, tension=0.7] coordinates {(0.729,0.215) (0.798,0.261) (0.822,0.336) (0.818,0.417) (0.768,0.487) (0.696,0.537) (0.636,0.598) (0.583,0.667) (0.505,0.710) (0.417,0.706) (0.337,0.683) (0.259,0.647) (0.209,0.573) (0.216,0.487) (0.260,0.421) (0.322,0.367) (0.403,0.337) (0.489,0.331) (0.571,0.303) (0.639,0.247)};
  \path[pff/lst/iso d] plot[smooth cycle, tension=0.7] coordinates {(0.681,0.317) (0.722,0.348) (0.695,0.391) (0.665,0.440) (0.633,0.488) (0.590,0.528) (0.554,0.572) (0.521,0.623) (0.467,0.652) (0.407,0.640) (0.356,0.616) (0.305,0.587) (0.281,0.529) (0.311,0.476) (0.354,0.445) (0.399,0.409) (0.454,0.389) (0.514,0.392) (0.572,0.383) (0.620,0.345)};
\end{scope}
\path[pff/frame] (0,0) rectangle (1,1);
}

\newcommand{\tileSOne}{%
\path[pff/sone/bg, pff/stackframe] (0.18,0.18) rectangle (1.18,1.18);
\path[pff/sone/bg, pff/stackframe] (0.09,0.09) rectangle (1.09,1.09);
\begin{scope}
  \clip[pff/corners] (0,0) rectangle (1,1);
  \path[pff/sone/bg] (0,0) rectangle (1,1);
  \path[pff/sone/wave b] plot[smooth, tension=0.7] coordinates {(0.588,-0.033) (0.623,-0.002) (0.661,0.025) (0.703,0.045) (0.748,0.058) (0.794,0.069) (0.836,0.089) (0.873,0.117) (0.902,0.153) (0.928,0.193) (0.954,0.231) (0.987,0.265) (1.022,0.296) (1.058,0.326)};
  \path[pff/sone/wave a] plot[smooth, tension=0.7] coordinates {(0.477,-0.010) (0.503,0.043) (0.536,0.091) (0.579,0.131) (0.630,0.160) (0.682,0.188) (0.731,0.221) (0.781,0.252) (0.834,0.277) (0.890,0.297) (0.940,0.328) (0.978,0.371) (1.008,0.422) (1.036,0.474)};
  \path[pff/sone/wave b] plot[smooth, tension=0.7] coordinates {(0.191,-0.005) (0.258,0.042) (0.335,0.069) (0.407,0.105) (0.475,0.151) (0.545,0.194) (0.610,0.241) (0.657,0.307) (0.698,0.378) (0.756,0.432) (0.830,0.465) (0.903,0.502) (0.971,0.547) (1.043,0.585)};
  \path[pff/sone/wave a] plot[smooth, tension=0.7] coordinates {(-0.048,-0.059) (0.036,0.013) (0.125,0.078) (0.220,0.135) (0.325,0.169) (0.408,0.240) (0.466,0.334) (0.547,0.409) (0.633,0.479) (0.726,0.538) (0.832,0.571) (0.919,0.635) (0.980,0.728) (1.060,0.804)};
  \path[pff/sone/wave b] plot[smooth, tension=0.7] coordinates {(-0.049,0.062) (0.046,0.105) (0.110,0.187) (0.168,0.275) (0.252,0.338) (0.336,0.402) (0.428,0.453) (0.528,0.486) (0.607,0.553) (0.662,0.643) (0.737,0.715) (0.819,0.782) (0.905,0.843) (1.005,0.874)};
  \path[pff/sone/wave a] plot[smooth, tension=0.7] coordinates {(-0.055,0.190) (0.027,0.263) (0.095,0.350) (0.176,0.426) (0.250,0.509) (0.345,0.564) (0.452,0.593) (0.538,0.660) (0.610,0.745) (0.690,0.822) (0.760,0.908) (0.852,0.967) (0.959,0.996) (1.048,1.059)};
  \path[pff/sone/wave b] plot[smooth, tension=0.7] coordinates {(-0.031,0.400) (0.031,0.443) (0.098,0.479) (0.162,0.518) (0.209,0.578) (0.242,0.646) (0.289,0.706) (0.355,0.742) (0.425,0.772) (0.489,0.813) (0.553,0.853) (0.623,0.884) (0.686,0.924) (0.729,0.987)};
  \path[pff/sone/wave a] plot[smooth, tension=0.7] coordinates {(-0.056,0.551) (0.006,0.570) (0.060,0.604) (0.107,0.647) (0.153,0.692) (0.201,0.735) (0.247,0.780) (0.284,0.833) (0.320,0.886) (0.370,0.925) (0.430,0.948) (0.492,0.966) (0.550,0.991) (0.601,1.031)};
  \path[pff/sone/wave b] plot[smooth, tension=0.7] coordinates {(-0.008,0.731) (0.023,0.746) (0.055,0.761) (0.086,0.779) (0.116,0.798) (0.139,0.824) (0.159,0.853) (0.177,0.883) (0.193,0.915) (0.210,0.945) (0.234,0.972) (0.258,0.997) (0.290,1.013) (0.321,1.029)};
  \path[pff/sone/speckle] plot[smooth, tension=0.7] coordinates {(0.576,0.696) (0.608,0.720)};
  \path[pff/sone/speckle] plot[smooth, tension=0.7] coordinates {(0.720,0.309) (0.752,0.334)};
  \path[pff/sone/speckle] plot[smooth, tension=0.7] coordinates {(0.647,0.846) (0.678,0.870)};
  \path[pff/sone/speckle] plot[smooth, tension=0.7] coordinates {(0.759,0.085) (0.790,0.110)};
  \path[pff/sone/speckle] plot[smooth, tension=0.7] coordinates {(0.245,0.326) (0.277,0.351)};
  \path[pff/sone/speckle] plot[smooth, tension=0.7] coordinates {(0.596,0.431) (0.627,0.456)};
  \path[pff/sone/speckle] plot[smooth, tension=0.7] coordinates {(0.763,0.603) (0.794,0.628)};
  \path[pff/sone/speckle] plot[smooth, tension=0.7] coordinates {(0.578,0.089) (0.610,0.114)};
  \path[pff/sone/speckle] plot[smooth, tension=0.7] coordinates {(0.158,0.541) (0.190,0.566)};
  \path[pff/sone/speckle] plot[smooth, tension=0.7] coordinates {(0.195,0.826) (0.226,0.851)};
  \path[pff/sone/speckle] plot[smooth, tension=0.7] coordinates {(0.433,0.340) (0.465,0.364)};
  \path[pff/sone/speckle] plot[smooth, tension=0.7] coordinates {(0.339,0.585) (0.371,0.610)};
  \path[pff/sone/speckle] plot[smooth, tension=0.7] coordinates {(0.336,0.157) (0.368,0.181)};
  \path[pff/sone/speckle] plot[smooth, tension=0.7] coordinates {(0.460,0.808) (0.491,0.833)};
  \path[pff/sone/speckle] plot[smooth, tension=0.7] coordinates {(0.831,0.827) (0.863,0.852)};
  \path[pff/sone/speckle] plot[smooth, tension=0.7] coordinates {(0.065,0.143) (0.096,0.167)};
  \path[pff/sone/speckle] plot[smooth, tension=0.7] coordinates {(0.890,0.319) (0.921,0.344)};
  \path[pff/sone/speckle] plot[smooth, tension=0.7] coordinates {(0.903,0.527) (0.935,0.551)};
  \path[pff/sone/speckle] plot[smooth, tension=0.7] coordinates {(0.079,0.346) (0.111,0.370)};
  \path[pff/sone/speckle] plot[smooth, tension=0.7] coordinates {(0.085,0.699) (0.116,0.724)};
\end{scope}
\path[pff/frame] (0,0) rectangle (1,1);
}

\newcommand{\tilePalsar}{%
\path[pff/palsar/bg, pff/stackframe] (0.18,0.18) rectangle (1.18,1.18);
\path[pff/palsar/bg, pff/stackframe] (0.09,0.09) rectangle (1.09,1.09);
\begin{scope}
  \clip[pff/corners] (0,0) rectangle (1,1);
  \path[pff/palsar/bg] (0,0) rectangle (1,1);
  \path[pff/palsar/wave hh] plot[smooth, tension=0.7] coordinates {(0.504,-0.058) (0.536,-0.025) (0.570,0.008) (0.605,0.039) (0.641,0.068) (0.680,0.095) (0.720,0.118) (0.763,0.138) (0.806,0.155) (0.851,0.168) (0.897,0.179) (0.942,0.188) (0.989,0.195) (1.035,0.203)};
  \path[pff/palsar/wave hh] plot[smooth, tension=0.7] coordinates {(-0.028,-0.002) (0.035,0.067) (0.100,0.134) (0.170,0.197) (0.246,0.251) (0.330,0.291) (0.420,0.319) (0.511,0.337) (0.604,0.352) (0.696,0.368) (0.787,0.392) (0.873,0.428) (0.953,0.476) (1.025,0.535)};
  \path[pff/palsar/wave hh] plot[smooth, tension=0.7] coordinates {(-0.031,0.476) (0.060,0.498) (0.152,0.514) (0.245,0.529) (0.336,0.548) (0.425,0.578) (0.509,0.620) (0.585,0.675) (0.654,0.739) (0.718,0.807) (0.783,0.875) (0.851,0.939) (0.926,0.995) (1.008,1.040)};
  \path[pff/palsar/wave hh] plot[smooth, tension=0.7] coordinates {(0.013,0.865) (0.053,0.873) (0.093,0.880) (0.134,0.887) (0.175,0.893) (0.215,0.900) (0.255,0.907) (0.295,0.916) (0.335,0.927) (0.374,0.939) (0.412,0.955) (0.449,0.972) (0.485,0.993) (0.519,1.015)};
  \path[pff/palsar/wave hv] plot[smooth, tension=0.7] coordinates {(1.019,0.567) (0.990,0.595) (0.961,0.624) (0.935,0.655) (0.909,0.687) (0.886,0.720) (0.864,0.755) (0.845,0.791) (0.829,0.828) (0.815,0.866) (0.804,0.906) (0.794,0.945) (0.787,0.985) (0.779,1.025)};
  \path[pff/palsar/wave hv] plot[smooth, tension=0.7] coordinates {(0.973,-0.044) (0.958,0.049) (0.939,0.140) (0.911,0.229) (0.870,0.313) (0.816,0.389) (0.753,0.458) (0.685,0.523) (0.617,0.587) (0.552,0.654) (0.495,0.729) (0.451,0.811) (0.419,0.899) (0.398,0.990)};
  \path[pff/palsar/wave hv] plot[smooth, tension=0.7] coordinates {(0.535,-0.026) (0.476,0.047) (0.428,0.127) (0.393,0.214) (0.369,0.304) (0.353,0.397) (0.338,0.489) (0.319,0.581) (0.292,0.670) (0.250,0.754) (0.196,0.830) (0.133,0.899) (0.065,0.964) (-0.003,1.028)};
  \path[pff/palsar/wave hv] plot[smooth, tension=0.7] coordinates {(0.233,-0.018) (0.212,0.002) (0.191,0.022) (0.169,0.042) (0.147,0.062) (0.126,0.083) (0.104,0.103) (0.083,0.124) (0.063,0.145) (0.043,0.167) (0.024,0.189) (0.005,0.212) (-0.012,0.237) (-0.028,0.261)};
\end{scope}
\path[pff/frame] (0,0) rectangle (1,1);
}

\newcommand{\tilePasture}{%
\path[pff/pasture/bg, pff/stackframe] (0.18,0.18) rectangle (1.18,1.18);
\path[pff/pasture/bg, pff/stackframe] (0.09,0.09) rectangle (1.09,1.09);
\begin{scope}
  \clip[pff/corners] (0,0) rectangle (1,1);
  \path[pff/pasture/bg] (0,0) rectangle (1,1);
  \path[pff/pasture/meadow] plot[smooth cycle, tension=0.7] coordinates {(0.490,0.002) (0.674,0.033) (0.858,0.002) (1.001,0.055) (0.998,0.255) (1.002,0.454) (1.010,0.654) (1.005,0.856) (0.855,0.894) (0.725,0.764) (0.563,0.666) (0.369,0.700) (0.190,0.774) (0.093,0.639) (0.282,0.583) (0.466,0.513) (0.577,0.352) (0.524,0.178)};
  \path[pff/pasture/meadow] plot[smooth cycle, tension=0.7] coordinates {(0.119,0.002) (0.188,-0.007) (0.262,-0.011) (0.289,0.049) (0.310,0.114) (0.319,0.170) (0.289,0.222) (0.236,0.268) (0.177,0.301) (0.118,0.333) (0.059,0.368) (0.009,0.393) (0.009,0.340) (-0.005,0.272) (-0.011,0.203) (-0.003,0.134) (0.030,0.094) (0.089,0.063)};
  \path[pff/pasture/dot] (0.358,0.273) circle[radius=0.012];
  \path[pff/pasture/dot] (0.385,0.586) circle[radius=0.012];
  \path[pff/pasture/dot] (0.675,0.409) circle[radius=0.012];
  \path[pff/pasture/dot] (0.527,0.433) circle[radius=0.012];
  \path[pff/pasture/dot] (0.541,0.913) circle[radius=0.012];
  \path[pff/pasture/dot] (0.748,0.191) circle[radius=0.012];
  \path[pff/pasture/dot] (0.436,0.077) circle[radius=0.012];
  \path[pff/pasture/dot] (0.341,0.685) circle[radius=0.012];
  \path[pff/pasture/dot] (0.534,0.265) circle[radius=0.012];
  \path[pff/pasture/dot] (0.096,0.649) circle[radius=0.012];
  \path[pff/pasture/dot] (0.063,0.221) circle[radius=0.012];
  \path[pff/pasture/dot] (0.796,0.897) circle[radius=0.012];
  \path[pff/pasture/dot] (0.266,0.240) circle[radius=0.012];
  \path[pff/pasture/dot] (0.372,0.519) circle[radius=0.012];
  \path[pff/pasture/dot] (0.793,0.373) circle[radius=0.012];
  \path[pff/pasture/dot] (0.906,0.735) circle[radius=0.012];
  \path[pff/pasture/dot] (0.195,0.635) circle[radius=0.012];
  \path[pff/pasture/dot] (0.894,0.896) circle[radius=0.012];
  \path[pff/pasture/dot] (0.356,0.174) circle[radius=0.012];
  \path[pff/pasture/dot] (0.608,0.133) circle[radius=0.012];
  \path[pff/pasture/dot] (0.885,0.257) circle[radius=0.012];
  \path[pff/pasture/dot] (0.839,0.527) circle[radius=0.012];
  \path[pff/pasture/dot] (0.147,0.264) circle[radius=0.012];
  \path[pff/pasture/dot] (0.838,0.646) circle[radius=0.012];
  \path[pff/pasture/dot] (0.492,0.368) circle[radius=0.012];
  \path[pff/pasture/dot] (0.723,0.480) circle[radius=0.012];
  \path[pff/pasture/dot] (0.770,0.283) circle[radius=0.012];
  \path[pff/pasture/dot] (0.654,0.497) circle[radius=0.012];
  \path[pff/pasture/dot] (0.935,0.127) circle[radius=0.012];
  \path[pff/pasture/dot] (0.696,0.071) circle[radius=0.012];
  \path[pff/pasture/dot] (0.203,0.200) circle[radius=0.012];
  \path[pff/pasture/dot] (0.900,0.066) circle[radius=0.012];
  \path[pff/pasture/dot] (0.836,0.204) circle[radius=0.012];
  \path[pff/pasture/dot] (0.730,0.560) circle[radius=0.012];
  \path[pff/pasture/dot] (0.796,0.762) circle[radius=0.012];
  \path[pff/pasture/dot] (0.178,0.084) circle[radius=0.012];
  \path[pff/pasture/dot] (0.227,0.839) circle[radius=0.012];
  \path[pff/pasture/dot] (0.565,0.542) circle[radius=0.012];
  \path[pff/pasture/dot] (0.484,0.490) circle[radius=0.012];
  \path[pff/pasture/dot] (0.718,0.740) circle[radius=0.012];
  \path[pff/pasture/dot] (0.152,0.418) circle[radius=0.012];
  \path[pff/pasture/dot] (0.603,0.400) circle[radius=0.012];
  \path[pff/pasture/dot] (0.915,0.824) circle[radius=0.012];
  \path[pff/pasture/dot] (0.676,0.198) circle[radius=0.012];
  \path[pff/pasture/dot] (0.148,0.813) circle[radius=0.012];
  \path[pff/pasture/dot] (0.901,0.560) circle[radius=0.012];
  \path[pff/pasture/dot] (0.093,0.158) circle[radius=0.012];
  \path[pff/pasture/dot] (0.051,0.497) circle[radius=0.012];
  \path[pff/pasture/dot] (0.263,0.157) circle[radius=0.012];
  \path[pff/pasture/dot] (0.109,0.321) circle[radius=0.012];
  \path[pff/pasture/dot] (0.919,0.318) circle[radius=0.012];
  \path[pff/pasture/dot] (0.788,0.455) circle[radius=0.012];
  \path[pff/pasture/dot] (0.106,0.066) circle[radius=0.012];
  \path[pff/pasture/tuft] plot[smooth, tension=0.9] coordinates {(0.200,0.240) (0.180,0.310) (0.150,0.370)};
  \path[pff/pasture/tuft] plot[smooth, tension=0.9] coordinates {(0.200,0.240) (0.200,0.310) (0.200,0.370)};
  \path[pff/pasture/tuft] plot[smooth, tension=0.9] coordinates {(0.200,0.240) (0.220,0.310) (0.250,0.370)};
  \path[pff/pasture/tuft] plot[smooth, tension=0.9] coordinates {(0.480,0.620) (0.460,0.683) (0.430,0.737)};
  \path[pff/pasture/tuft] plot[smooth, tension=0.9] coordinates {(0.480,0.620) (0.480,0.683) (0.480,0.737)};
  \path[pff/pasture/tuft] plot[smooth, tension=0.9] coordinates {(0.480,0.620) (0.500,0.683) (0.530,0.737)};
  \path[pff/pasture/tuft] plot[smooth, tension=0.9] coordinates {(0.780,0.360) (0.760,0.437) (0.730,0.503)};
  \path[pff/pasture/tuft] plot[smooth, tension=0.9] coordinates {(0.780,0.360) (0.780,0.437) (0.780,0.503)};
  \path[pff/pasture/tuft] plot[smooth, tension=0.9] coordinates {(0.780,0.360) (0.800,0.437) (0.830,0.503)};
\end{scope}
\path[pff/frame] (0,0) rectangle (1,1);
}

\newcommand{\tileBiomass}{%
\path[pff/biomass/bg, pff/stackframe] (0.18,0.18) rectangle (1.18,1.18);
\path[pff/biomass/bg, pff/stackframe] (0.09,0.09) rectangle (1.09,1.09);
\begin{scope}
  \clip[pff/corners] (0,0) rectangle (1,1);
  \path[pff/biomass/bg] (0,0) rectangle (1,1);
  \path[pff/biomass/crown a] plot[smooth cycle, tension=0.7] coordinates {(0.855,0.874) (0.847,0.914) (0.835,0.953) (0.816,0.989) (0.786,1.014) (0.749,1.012) (0.719,0.985) (0.694,0.955) (0.655,0.942) (0.628,0.915) (0.627,0.875) (0.632,0.834) (0.645,0.796) (0.666,0.761) (0.697,0.737) (0.734,0.738) (0.764,0.764) (0.791,0.793) (0.829,0.806) (0.856,0.834)};
  \path[pff/biomass/crown b] plot[smooth cycle, tension=0.7] coordinates {(0.581,0.873) (0.568,0.931) (0.558,0.989) (0.532,1.038) (0.476,1.051) (0.420,1.030) (0.363,1.018) (0.305,1.008) (0.254,0.978) (0.220,0.931) (0.229,0.873) (0.250,0.819) (0.247,0.760) (0.263,0.707) (0.321,0.694) (0.376,0.716) (0.435,0.715) (0.489,0.726) (0.525,0.772) (0.561,0.819)};
  \path[pff/biomass/crown a] plot[smooth cycle, tension=0.7] coordinates {(0.294,0.772) (0.275,0.823) (0.247,0.869) (0.205,0.902) (0.154,0.921) (0.101,0.934) (0.047,0.937) (-0.003,0.917) (-0.035,0.873) (-0.048,0.820) (-0.045,0.766) (-0.018,0.720) (0.025,0.687) (0.063,0.649) (0.102,0.610) (0.153,0.596) (0.207,0.604) (0.257,0.625) (0.294,0.665) (0.305,0.718)};
  \path[pff/biomass/crown b] plot[smooth cycle, tension=0.7] coordinates {(0.576,0.639) (0.574,0.679) (0.567,0.717) (0.538,0.744) (0.501,0.760) (0.462,0.759) (0.431,0.735) (0.397,0.714) (0.358,0.706) (0.329,0.678) (0.329,0.638) (0.344,0.600) (0.370,0.571) (0.409,0.560) (0.442,0.540) (0.473,0.515) (0.510,0.512) (0.543,0.532) (0.571,0.562) (0.583,0.600)};
  \path[pff/biomass/crown a] plot[smooth cycle, tension=0.7] coordinates {(0.902,0.538) (0.883,0.588) (0.855,0.635) (0.826,0.680) (0.783,0.710) (0.730,0.717) (0.678,0.706) (0.633,0.677) (0.596,0.637) (0.571,0.590) (0.569,0.536) (0.583,0.484) (0.596,0.431) (0.625,0.388) (0.672,0.364) (0.725,0.358) (0.777,0.369) (0.817,0.402) (0.849,0.446) (0.884,0.487)};
  \path[pff/biomass/crown b] plot[smooth cycle, tension=0.7] coordinates {(0.305,0.465) (0.298,0.524) (0.284,0.580) (0.251,0.629) (0.206,0.667) (0.153,0.674) (0.108,0.642) (0.075,0.593) (0.024,0.566) (-0.016,0.530) (-0.006,0.476) (0.005,0.419) (0.001,0.361) (0.030,0.311) (0.081,0.280) (0.136,0.263) (0.189,0.279) (0.228,0.324) (0.269,0.365) (0.307,0.408)};
  \path[pff/biomass/crown a] plot[smooth cycle, tension=0.7] coordinates {(0.499,0.374) (0.482,0.408) (0.468,0.444) (0.444,0.472) (0.408,0.483) (0.370,0.486) (0.331,0.488) (0.295,0.476) (0.268,0.449) (0.257,0.412) (0.260,0.374) (0.271,0.337) (0.292,0.306) (0.323,0.284) (0.358,0.266) (0.393,0.252) (0.431,0.253) (0.464,0.272) (0.490,0.300) (0.505,0.336)};
  \path[pff/biomass/crown b] plot[smooth cycle, tension=0.7] coordinates {(0.755,0.178) (0.752,0.233) (0.719,0.277) (0.677,0.314) (0.635,0.351) (0.584,0.362) (0.533,0.343) (0.483,0.317) (0.441,0.281) (0.425,0.229) (0.414,0.174) (0.408,0.120) (0.433,0.071) (0.476,0.035) (0.523,0.007) (0.576,0.003) (0.628,0.024) (0.681,0.042) (0.727,0.072) (0.747,0.122)};
  \path[pff/biomass/crown a] plot[smooth cycle, tension=0.7] coordinates {(0.984,0.143) (0.987,0.185) (0.971,0.224) (0.944,0.256) (0.907,0.275) (0.867,0.269) (0.830,0.250) (0.789,0.241) (0.750,0.225) (0.731,0.189) (0.734,0.147) (0.743,0.105) (0.759,0.066) (0.789,0.037) (0.829,0.026) (0.871,0.025) (0.914,0.021) (0.955,0.028) (0.980,0.059) (0.983,0.101)};
  \path[pff/biomass/crown b] plot[smooth cycle, tension=0.7] coordinates {(0.263,0.122) (0.257,0.172) (0.248,0.221) (0.233,0.267) (0.198,0.300) (0.149,0.303) (0.103,0.284) (0.068,0.248) (0.054,0.200) (0.032,0.159) (-0.006,0.127) (-0.023,0.080) (-0.014,0.031) (0.014,-0.010) (0.059,-0.033) (0.108,-0.038) (0.158,-0.035) (0.196,-0.008) (0.212,0.040) (0.233,0.082)};
\end{scope}
\path[pff/frame] (0,0) rectangle (1,1);
}

\newcommand{\tileWater}{%
\path[pff/water/bg, pff/stackframe] (0.18,0.18) rectangle (1.18,1.18);
\path[pff/water/bg, pff/stackframe] (0.09,0.09) rectangle (1.09,1.09);
\begin{scope}
  \clip[pff/corners] (0,0) rectangle (1,1);
  \path[pff/water/bg] (0,0) rectangle (1,1);
  \path[pff/water/river] plot[smooth, tension=0.8] coordinates {(0.860,1.060) (0.760,0.860) (0.830,0.700) (0.700,0.560) (0.580,0.500)};
  \path[pff/water/lake] plot[smooth cycle, tension=0.7] coordinates {(0.698,0.420) (0.694,0.494) (0.700,0.566) (0.659,0.627) (0.584,0.628) (0.512,0.616) (0.440,0.607) (0.366,0.611) (0.298,0.588) (0.243,0.540) (0.185,0.494) (0.141,0.435) (0.145,0.365) (0.191,0.310) (0.251,0.266) (0.314,0.228) (0.384,0.202) (0.457,0.204) (0.524,0.233) (0.590,0.266) (0.659,0.294) (0.707,0.349)};
  \path[pff/water/ripple] plot[smooth, tension=0.9] coordinates {(0.255,0.460) (0.340,0.482) (0.425,0.460)};
  \path[pff/water/ripple] plot[smooth, tension=0.9] coordinates {(0.395,0.320) (0.480,0.342) (0.565,0.320)};
\end{scope}
\path[pff/frame] (0,0) rectangle (1,1);
}

\newcommand{\tileDEM}{%
\begin{scope}
  \clip[pff/corners] (0,0) rectangle (1,1);
  \path[pff/dem/bg] (0,0) rectangle (1,1);
  \path[pff/dem/contour] plot[smooth, tension=0.7] coordinates {(1.000,0.334) (0.946,0.438) (0.881,0.535) (0.826,0.639) (0.778,0.745) (0.703,0.835) (0.608,0.904) (0.504,0.959) (0.392,0.992) (0.279,0.971) (0.185,0.904) (0.104,0.820) (0.039,0.724) (0.009,0.611) (0.037,0.496) (0.106,0.401) (0.200,0.331) (0.308,0.289) (0.419,0.247) (0.498,0.159) (0.571,0.068) (0.663,-0.003)};
  \path[pff/dem/index] plot[smooth cycle, tension=0.7] coordinates {(0.784,0.014) (0.908,0.044) (0.980,0.150) (0.981,0.280) (0.934,0.401) (0.858,0.504) (0.779,0.605) (0.718,0.717) (0.636,0.814) (0.533,0.888) (0.415,0.937) (0.289,0.932) (0.180,0.864) (0.098,0.764) (0.055,0.642) (0.078,0.516) (0.156,0.415) (0.266,0.350) (0.390,0.315) (0.493,0.240) (0.566,0.135) (0.662,0.052)};
  \path[pff/dem/contour] plot[smooth cycle, tension=0.7] coordinates {(0.729,0.059) (0.843,0.073) (0.924,0.153) (0.943,0.266) (0.908,0.376) (0.839,0.469) (0.760,0.556) (0.706,0.660) (0.645,0.762) (0.553,0.837) (0.446,0.884) (0.333,0.896) (0.228,0.853) (0.147,0.770) (0.101,0.663) (0.107,0.547) (0.169,0.447) (0.269,0.384) (0.385,0.358) (0.490,0.307) (0.550,0.204) (0.621,0.111)};
  \path[pff/dem/index] plot[smooth cycle, tension=0.7] coordinates {(0.738,0.091) (0.845,0.107) (0.912,0.195) (0.902,0.302) (0.846,0.389) (0.778,0.466) (0.710,0.551) (0.668,0.650) (0.610,0.743) (0.526,0.815) (0.429,0.865) (0.319,0.869) (0.228,0.810) (0.169,0.725) (0.135,0.627) (0.152,0.520) (0.230,0.443) (0.334,0.412) (0.442,0.392) (0.522,0.319) (0.562,0.217) (0.633,0.134)};
  \path[pff/dem/contour] plot[smooth cycle, tension=0.7] coordinates {(0.730,0.127) (0.829,0.140) (0.889,0.223) (0.885,0.325) (0.829,0.412) (0.743,0.466) (0.652,0.514) (0.610,0.609) (0.582,0.706) (0.520,0.788) (0.426,0.835) (0.325,0.832) (0.240,0.781) (0.184,0.701) (0.172,0.606) (0.209,0.517) (0.283,0.450) (0.379,0.415) (0.477,0.405) (0.554,0.351) (0.597,0.260) (0.650,0.178)};
  \path[pff/dem/index] plot[smooth cycle, tension=0.7] coordinates {(0.723,0.154) (0.765,0.144) (0.810,0.151) (0.843,0.183) (0.851,0.228) (0.846,0.267) (0.844,0.304) (0.837,0.342) (0.815,0.376) (0.785,0.402) (0.753,0.428) (0.717,0.449) (0.676,0.457) (0.634,0.450) (0.602,0.427) (0.587,0.387) (0.588,0.346) (0.605,0.307) (0.626,0.272) (0.641,0.238) (0.658,0.202) (0.686,0.173)};
  \path[pff/dem/index] plot[smooth cycle, tension=0.7] coordinates {(0.363,0.460) (0.416,0.460) (0.468,0.468) (0.519,0.476) (0.564,0.497) (0.579,0.547) (0.576,0.600) (0.563,0.651) (0.540,0.699) (0.507,0.739) (0.464,0.769) (0.417,0.790) (0.367,0.802) (0.314,0.799) (0.266,0.774) (0.232,0.733) (0.210,0.686) (0.199,0.636) (0.205,0.585) (0.230,0.539) (0.267,0.502) (0.311,0.475)};
  \path[pff/dem/contour] plot[smooth cycle, tension=0.7] coordinates {(0.727,0.190) (0.765,0.184) (0.802,0.200) (0.818,0.236) (0.825,0.271) (0.828,0.309) (0.810,0.343) (0.781,0.368) (0.751,0.390) (0.714,0.400) (0.678,0.393) (0.646,0.374) (0.627,0.339) (0.635,0.300) (0.652,0.266) (0.667,0.232) (0.692,0.204)};
  \path[pff/dem/contour] plot[smooth cycle, tension=0.7] coordinates {(0.346,0.498) (0.387,0.493) (0.427,0.497) (0.466,0.511) (0.497,0.536) (0.515,0.570) (0.519,0.608) (0.514,0.646) (0.500,0.682) (0.478,0.716) (0.449,0.744) (0.414,0.764) (0.373,0.773) (0.331,0.767) (0.295,0.748) (0.267,0.719) (0.249,0.684) (0.241,0.647) (0.242,0.609) (0.253,0.572) (0.276,0.539) (0.307,0.514)};
  \path[pff/dem/index] plot[smooth cycle, tension=0.7] coordinates {(0.735,0.227) (0.773,0.232) (0.790,0.267) (0.790,0.304) (0.762,0.329) (0.733,0.351) (0.694,0.349) (0.672,0.315) (0.689,0.279) (0.703,0.246)};
  \path[pff/dem/index] plot[smooth cycle, tension=0.7] coordinates {(0.370,0.540) (0.408,0.552) (0.444,0.563) (0.471,0.593) (0.470,0.633) (0.459,0.670) (0.431,0.698) (0.396,0.713) (0.359,0.728) (0.319,0.717) (0.295,0.684) (0.281,0.649) (0.283,0.611) (0.303,0.578) (0.329,0.550)};
\end{scope}
\path[pff/frame] (0,0) rectangle (1,1);
}

\newcommand{\tileCanopy}{%
\begin{scope}
  \clip[pff/corners] (0,0) rectangle (1,1);
  \path[pff/canopy/bg] (0,0) rectangle (1,1);
  \path[pff/canopy/ground] plot[smooth, tension=0.8] coordinates {(-0.050,0.137) (0.050,0.133) (0.150,0.143) (0.250,0.127) (0.350,0.158) (0.450,0.147) (0.550,0.152) (0.650,0.137) (0.750,0.168) (0.850,0.136) (0.950,0.133) (1.050,0.149)};
  \path[pff/canopy/height] plot[smooth, tension=0.7] coordinates {(0.040,0.986) (0.960,0.986)};
  \path[pff/canopy/trunk] plot[smooth, tension=0.7] coordinates {(0.150,0.140) (0.150,0.404)};
  \path[pff/canopy/crown a] plot[smooth cycle, tension=0.7] coordinates {(0.268,0.404) (0.266,0.444) (0.242,0.476) (0.206,0.494) (0.169,0.509) (0.129,0.515) (0.091,0.502) (0.059,0.478) (0.035,0.446) (0.027,0.407) (0.039,0.368) (0.067,0.339) (0.099,0.314) (0.132,0.291) (0.172,0.284) (0.208,0.300) (0.235,0.330) (0.254,0.366)};
  \path[pff/canopy/trunk] plot[smooth, tension=0.7] coordinates {(0.400,0.140) (0.400,0.729)};
  \path[pff/canopy/crown b] plot[smooth cycle, tension=0.7] coordinates {(0.548,0.729) (0.523,0.775) (0.498,0.819) (0.457,0.847) (0.409,0.866) (0.361,0.879) (0.316,0.859) (0.281,0.822) (0.250,0.781) (0.248,0.730) (0.276,0.687) (0.303,0.644) (0.340,0.610) (0.388,0.592) (0.438,0.583) (0.484,0.601) (0.523,0.636) (0.550,0.679)};
  \path[pff/canopy/trunk] plot[smooth, tension=0.7] coordinates {(0.630,0.140) (0.630,0.557)};
  \path[pff/canopy/crown a] plot[smooth cycle, tension=0.7] coordinates {(0.753,0.557) (0.735,0.599) (0.712,0.638) (0.686,0.675) (0.647,0.697) (0.602,0.694) (0.562,0.674) (0.532,0.641) (0.516,0.599) (0.511,0.554) (0.518,0.509) (0.538,0.469) (0.571,0.439) (0.615,0.430) (0.660,0.434) (0.704,0.443) (0.739,0.470) (0.754,0.512)};
  \path[pff/canopy/trunk] plot[smooth, tension=0.7] coordinates {(0.860,0.140) (0.860,0.825)};
  \path[pff/canopy/crown b] plot[smooth cycle, tension=0.7] coordinates {(1.022,0.825) (1.015,0.879) (0.977,0.916) (0.925,0.937) (0.881,0.969) (0.829,0.982) (0.783,0.954) (0.738,0.922) (0.699,0.884) (0.692,0.831) (0.716,0.782) (0.745,0.735) (0.788,0.703) (0.842,0.695) (0.895,0.679) (0.947,0.684) (0.983,0.722) (1.005,0.773)};
\end{scope}
\path[pff/frame] (0,0) rectangle (1,1);
}

\newcommand{\tileWetland}{%
\begin{scope}
  \clip[pff/corners] (0,0) rectangle (1,1);
  \path[pff/wetland/bg] (0,0) rectangle (1,1);
  \path[pff/wetland/band] plot[smooth cycle, tension=0.6] coordinates {(-0.050,0.646) (0.035,0.693) (0.119,0.654) (0.204,0.634) (0.288,0.649) (0.373,0.600) (0.458,0.564) (0.542,0.598) (0.627,0.610) (0.712,0.647) (0.796,0.675) (0.881,0.641) (0.965,0.618) (1.050,0.611) (1.050,-0.050) (-0.050,-0.050)};
  \path[pff/wetland/hummock] plot[smooth cycle, tension=0.7] coordinates {(0.385,0.300) (0.371,0.324) (0.343,0.335) (0.315,0.348) (0.285,0.351) (0.256,0.340) (0.230,0.326) (0.215,0.302) (0.231,0.277) (0.255,0.260) (0.285,0.256) (0.316,0.256) (0.346,0.258) (0.372,0.273)};
  \path[pff/wetland/hummock] plot[smooth cycle, tension=0.7] coordinates {(0.815,0.200) (0.802,0.235) (0.772,0.255) (0.735,0.261) (0.697,0.256) (0.662,0.243) (0.629,0.226) (0.607,0.198) (0.631,0.171) (0.666,0.158) (0.701,0.145) (0.738,0.139) (0.775,0.145) (0.806,0.165)};
  \path[pff/wetland/reed] plot[smooth, tension=0.9] coordinates {(0.140,0.629) (0.121,0.719) (0.085,0.779)};
  \path[pff/wetland/reed] plot[smooth, tension=0.9] coordinates {(0.140,0.629) (0.140,0.755) (0.140,0.839)};
  \path[pff/wetland/reed] plot[smooth, tension=0.9] coordinates {(0.140,0.629) (0.159,0.731) (0.195,0.799)};
  \path[pff/wetland/reed] plot[smooth, tension=0.9] coordinates {(0.360,0.587) (0.341,0.664) (0.305,0.715)};
  \path[pff/wetland/reed] plot[smooth, tension=0.9] coordinates {(0.360,0.587) (0.360,0.694) (0.360,0.766)};
  \path[pff/wetland/reed] plot[smooth, tension=0.9] coordinates {(0.360,0.587) (0.379,0.674) (0.415,0.732)};
  \path[pff/wetland/reed] plot[smooth, tension=0.9] coordinates {(0.550,0.579) (0.531,0.678) (0.495,0.744)};
  \path[pff/wetland/reed] plot[smooth, tension=0.9] coordinates {(0.550,0.579) (0.550,0.718) (0.550,0.810)};
  \path[pff/wetland/reed] plot[smooth, tension=0.9] coordinates {(0.550,0.579) (0.569,0.692) (0.605,0.766)};
  \path[pff/wetland/reed] plot[smooth, tension=0.9] coordinates {(0.880,0.622) (0.861,0.703) (0.825,0.757)};
  \path[pff/wetland/reed] plot[smooth, tension=0.9] coordinates {(0.880,0.622) (0.880,0.735) (0.880,0.811)};
  \path[pff/wetland/reed] plot[smooth, tension=0.9] coordinates {(0.880,0.622) (0.899,0.714) (0.935,0.775)};
  \path[pff/wetland/ripple] plot[smooth, tension=0.7] coordinates {(0.100,0.160) (0.160,0.174) (0.220,0.160)};
  \path[pff/wetland/ripple] plot[smooth, tension=0.7] coordinates {(0.460,0.100) (0.520,0.114) (0.580,0.100)};
  \path[pff/wetland/ripple] plot[smooth, tension=0.7] coordinates {(0.840,0.340) (0.900,0.354) (0.960,0.340)};
\end{scope}
\path[pff/frame] (0,0) rectangle (1,1);
}

\newcommand{\tileSoil}{%
\begin{scope}
  \clip[pff/corners] (0,0) rectangle (1,1);
  \path[pff/soil/bg] (0,0) rectangle (1,1);
  \path[pff/soil/h1] plot[smooth cycle, tension=0.5] coordinates {(-0.050,1.050) (1.050,1.050) (1.050,0.752) (0.950,0.777) (0.850,0.742) (0.750,0.727) (0.650,0.704) (0.550,0.711) (0.450,0.751) (0.350,0.762) (0.250,0.762) (0.150,0.756) (0.050,0.736) (-0.050,0.721)};
  \path[pff/soil/h2] plot[smooth cycle, tension=0.5] coordinates {(-0.050,0.721) (0.050,0.736) (0.150,0.756) (0.250,0.762) (0.350,0.762) (0.450,0.751) (0.550,0.711) (0.650,0.704) (0.750,0.727) (0.850,0.742) (0.950,0.777) (1.050,0.752) (1.050,0.533) (0.950,0.533) (0.850,0.529) (0.750,0.495) (0.650,0.514) (0.550,0.475) (0.450,0.503) (0.350,0.518) (0.250,0.529) (0.150,0.548) (0.050,0.534) (-0.050,0.551)};
  \path[pff/soil/h3] plot[smooth cycle, tension=0.5] coordinates {(-0.050,0.551) (0.050,0.534) (0.150,0.548) (0.250,0.529) (0.350,0.518) (0.450,0.503) (0.550,0.475) (0.650,0.514) (0.750,0.495) (0.850,0.529) (0.950,0.533) (1.050,0.533) (1.050,0.344) (0.950,0.335) (0.850,0.293) (0.750,0.274) (0.650,0.275) (0.550,0.255) (0.450,0.289) (0.350,0.312) (0.250,0.334) (0.150,0.330) (0.050,0.301) (-0.050,0.300)};
  \path[pff/soil/h4] plot[smooth cycle, tension=0.5] coordinates {(-0.050,0.300) (0.050,0.301) (0.150,0.330) (0.250,0.334) (0.350,0.312) (0.450,0.289) (0.550,0.255) (0.650,0.275) (0.750,0.274) (0.850,0.293) (0.950,0.335) (1.050,0.344) (1.050,-0.050) (-0.050,-0.050)};
  \path[pff/soil/line] plot[smooth, tension=0.6] coordinates {(-0.050,0.721) (0.050,0.736) (0.150,0.756) (0.250,0.762) (0.350,0.762) (0.450,0.751) (0.550,0.711) (0.650,0.704) (0.750,0.727) (0.850,0.742) (0.950,0.777) (1.050,0.752)};
  \path[pff/soil/line] plot[smooth, tension=0.6] coordinates {(-0.050,0.551) (0.050,0.534) (0.150,0.548) (0.250,0.529) (0.350,0.518) (0.450,0.503) (0.550,0.475) (0.650,0.514) (0.750,0.495) (0.850,0.529) (0.950,0.533) (1.050,0.533)};
  \path[pff/soil/line] plot[smooth, tension=0.6] coordinates {(-0.050,0.300) (0.050,0.301) (0.150,0.330) (0.250,0.334) (0.350,0.312) (0.450,0.289) (0.550,0.255) (0.650,0.275) (0.750,0.274) (0.850,0.293) (0.950,0.335) (1.050,0.344)};
  \path[pff/soil/pebble] (0.099,0.057) circle[radius=0.016];
  \path[pff/soil/pebble] (0.624,0.194) circle[radius=0.016];
  \path[pff/soil/pebble] (0.271,0.081) circle[radius=0.016];
  \path[pff/soil/pebble] (0.703,0.084) circle[radius=0.016];
  \path[pff/soil/pebble] (0.122,0.176) circle[radius=0.016];
  \path[pff/soil/pebble] (0.871,0.172) circle[radius=0.016];
  \path[pff/soil/pebble] (0.906,0.065) circle[radius=0.016];
  \path[pff/soil/pebble] (0.499,0.158) circle[radius=0.016];
  \path[pff/soil/pebble] (0.436,0.056) circle[radius=0.016];
  \path[pff/soil/pebble] (0.590,0.046) circle[radius=0.016];
  \path[pff/soil/pebble] (0.238,0.198) circle[radius=0.016];
  \path[pff/soil/pebble] (0.372,0.176) circle[radius=0.016];
  \path[pff/soil/pebble] (0.759,0.192) circle[radius=0.016];
\end{scope}
\path[pff/frame] (0,0) rectangle (1,1);
}

\fi
\begin{figure}[t]
\definecolor{explicit}{HTML}{EF8A2C}
\definecolor{implicit}{HTML}{4A7FA5}

\resizebox{\textwidth}{!}{%
\input{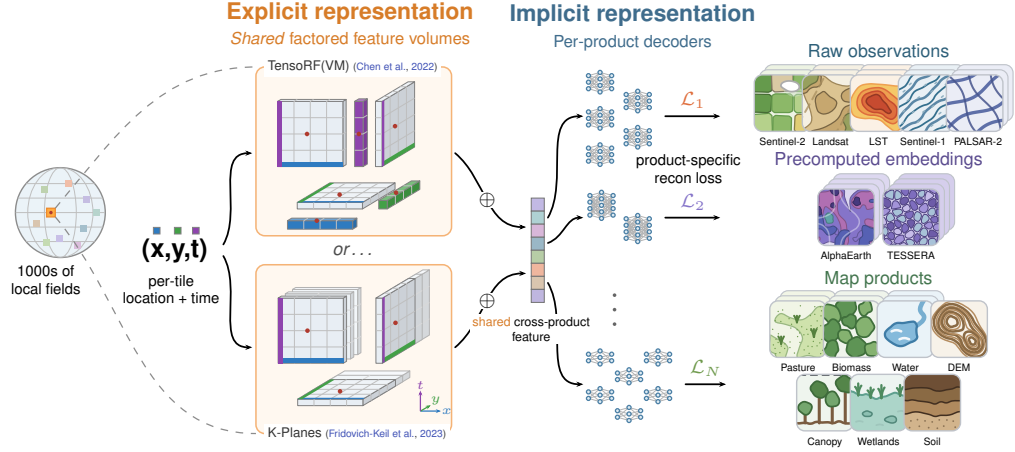}%
\hspace{-1.5mm}%
\begin{tikzpicture}[x=0.78cm, y=0.78cm]

\node[pff/panel title, text=pffRaw!85!black] at (2.8,1.25) {Raw observations};
\node[pff/panel title, text=pffLearned!85!black] at (2.8,-0.72)
  {Precomputed embeddings};
\node[pff/panel title, text=pffMaps!85!black] at (2.8,-2.82) {Map products};

\foreach \x/\y/\tilecmd/\lbl/\lx in {%
    0.60/0/\tileSTwo/Sentinel-2/0.5,
    1.45/0/\tileLandsat/Landsat/0.5,
    2.30/0/\tileLST/LST/0.5,
    3.15/0/\tileSOne/Sentinel-1/0.44,
    4.00/0/\tilePalsar/PALSAR-2/0.56,
    1.725/-2.02/\tileAE/AlphaEarth/0.5,
    2.875/-2.02/\tileTE/TESSERA/0.5,
    0.875/-4.00/\tilePasture/Pasture/0.5,
    1.825/-4.00/\tileBiomass/Biomass/0.5,
    2.775/-4.00/\tileWater/Water/0.5,
    3.725/-4.00/\tileDEM/DEM/0.5,
    1.35/-5.30/\tileCanopy/Canopy/0.5,
    2.30/-5.30/\tileWetland/Wetlands/0.5,
    3.25/-5.30/\tileSoil/Soil/0.5}{
  \begin{scope}[shift={(\x,\y)}]
    \tilecmd
    \node[pff/tile label] at (\lx,-0.06) {\lbl};
  \end{scope}}

\end{tikzpicture}%
}
\caption{\textbf{PFF construction.} 
Each regional PFF combines a shared \textcolor{explicit}{\textbf{explicit}} factored feature volume with product-specific \textcolor{implicit}{\textbf{implicit}} decoders.
The volume represents a dense space--time feature grid using learned lines or planes.
At a queried location and time, we combine factor values into a shared feature vector, which each decoder maps to its product's native feature space.
Thousands of regional PFFs extend this construction to planetary coverage.
}
\label{fig:arch}

\end{figure}

We hypothesize that an ideal Earth representation is a \emph{hybrid}, incorporating characteristics from both explicit and implicit representations and is optimized \emph{per scene}.
Our proposed solution to these usability issues is \textbf{Planetary Feature Fields (PFFs)}, a global representation composed of spatially local explicit--implicit (hybrid) neural fields.
Each field consists of an explicit \emph{factored feature volume} \citep{tilted}, built from learned lines or planes \emph{shared} across products, and lightweight product-specific MLP decoders (\Cref{fig:arch}). 
Our explicit--implicit approach makes it possible to represent products with different dimensions, resolutions, and statistics, achieving reconstruction with high fidelity at high compression ratios.
We instantiate PFFs on 14 EO products (see \Cref{tab:modalities}) with observations spanning nine years (2017--2025), including five raw observations \citep{sentinel2,sentinel1,landsatc2,palsar}, 
two precomputed embeddings \citep{tessera,alphaearth}, 
and seven map products \citep{gpw,biomasscci7,gladwater,copdem,canopy,glwd2,soilgrids}.

As new observations and products become available, especially over time, the challenge with learned representations (including implicit representations) is the need to repeatedly retrain on the new and historical data. 
We seek a learned representation that can be efficiently extended without forgetting historical data. 
PFFs support this growth by extending the temporal factors in their factored feature volumes for new timesteps and fitting additional decoders for new products. 
Because only the extended factors are trained, previous outputs remain unchanged (\emph{bit-exactly}), preserving the features used by existing analyses. 
Adding a year of observations across multiple EO products takes only $2\%$ of the time required for joint pretraining from scratch, while reaching 92\% of the jointly trained model's new-year reconstruction score (\Cref{fig:temporal-append-result}).
Furthermore, each regional field can be extended independently, so products with limited geographic coverage can be incorporated where observations exist without global retraining.

In summary, we make three contributions:
\begin{enumerate*}[label=(\roman*)]
\item We introduce \textbf{Planetary Feature Fields}: spatially local explicit--implicit Earth representations that achieve high-fidelity reconstruction under large compression ratios ($\approx$1800$\times$) by sharing an explicit factored feature volume across data products.
\item We demonstrate that PFF-reconstructed features at their canonical compression ratios retain $\approx$90\% of downstream task performance on patch-level classification, high-resolution semantic segmentation, and change-detection tasks while retrieving features an order of magnitude faster than current APIs or cloud-based systems.
\item We show that PFFs enable efficient continual learning of new timesteps and data products while leaving the existing representation unchanged.
\end{enumerate*}

\begin{figure}[t!]
\captionsetup[subfigure]{justification=raggedright,singlelinecheck=false}
\begin{subfigure}[t]{0.655\textwidth}
\centering
  \makebox[\linewidth][c]{%
    \hspace*{-60pt}%
    \tikz\node[font=\small, inner sep=0pt]
      {\textbf{(a)} New Timesteps};%
    \hspace*{10pt}%
  }
  \par\vspace{3pt}
\input{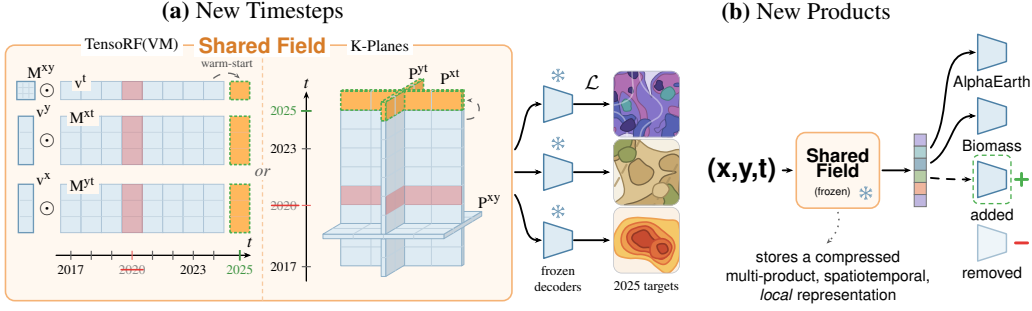}%
\label{fig:arch-temporal}
\end{subfigure}\hfill
\begin{subfigure}[t]{0.335\textwidth}
\centering
  \makebox[\linewidth][c]{%
    \hspace*{-40pt}%
    \tikz\node[font=\small, inner sep=0pt]
      {\textbf{(b)} New Products};%
    \hspace*{10pt}%
  }
  \par\vspace{3pt}

\begin{tikzpicture}[x=1cm, y=1cm]

\path (0,-1.85) (0,2.15);

\providecommand{\adecoder}{}
\renewcommand{\adecoder}[2]{%
  \path[fill=decFillB, draw=decB, line width=0.55pt, line join=round]
    (#1,{#2-0.24}) -- ({#1+0.38},{#2-0.12}) -- ({#1+0.38},{#2+0.12})
    -- (#1,{#2+0.24}) -- cycle;}

\node[pff/arch/input] at (0.48,0.36) {(x,y,t)};
\draw[pff/arch/arrow] (1.02,0.36) -- (1.16,0.36);

\path[pff/arch/partbox] (1.20,-0.14) rectangle (2.30,0.86);
\foreach \fx/\fy in {2.12/0.06}{
  \foreach \a in {30,90,150,210,270,330}{
    \draw[decB!75, line width=0.4pt, line cap=round]
      (\fx,\fy) -- ++(\a:0.085);
    \draw[decB!75, line width=0.35pt, line cap=round]
      (\fx,\fy) ++(\a:0.055) -- ++({\a+35}:0.032);
    \draw[decB!75, line width=0.35pt, line cap=round]
      (\fx,\fy) ++(\a:0.055) -- ++({\a-35}:0.032);}}
\node[pff/group title, anchor=center, text=black,
      font=\fontsize{7}{8.4}\selectfont\bfseries\sffamily] at (1.75,0.46)
  {Shared\\[-2.5pt]Field};
\node[font=\fontsize{4}{4.8}\selectfont\sffamily, text=black,
      anchor=north, inner sep=0.5pt] at (1.68,0.16) {(frozen)};
\draw[pff/arch/arrow] (2.34,0.36) -- (2.72,0.36);
\foreach \i/\zc in {0/aeV, 1/aeT, 2/sOneB, 3/gpwD, 4/lstC, 5/teC}{
  \node[pff/arch/zcell, fill=\zc!50, minimum width=0.16cm,
        minimum height=0.16cm, inner sep=0] at (2.84,{0.76-0.16*\i}) {};}

\draw[dotted, black!60, line width=0.6pt, -{Stealth[length=2.2pt]}]
  (1.75,-0.19) to[out=-75,in=45] (1.60,-0.61);
\node[pff/arch/lossfirst, anchor=north] at (1.62,-0.67)
  {stores a compressed\\multi-product, spatiotemporal,\\\emph{local} representation};

\draw[pff/arch/arrow] (2.98,0.66) to[out=55,in=180] (3.58,1.92);
\draw[pff/arch/arrow] (2.98,0.46) to[out=30,in=180] (3.58,1.08);
\adecoder{3.60}{1.92}
\node[pff/arch/lossfirst, anchor=north] at (3.79,1.65) {AlphaEarth};
\adecoder{3.60}{1.08}
\node[pff/arch/lossfirst, anchor=north] at (3.79,0.81) {Biomass};

\draw[pff/arch/arrow, dashed] (2.98,0.26) to[out=-5,in=180] (3.52,0.24);
\path[rounded corners=2pt, draw=addG, line width=0.5pt,
      dash pattern=on 1.8pt off 1.3pt] (3.54,-0.08) rectangle (4.04,0.56);
\adecoder{3.60}{0.24}
\draw[addG, line width=1.1pt, line cap=round] (4.10,0.24) -- (4.26,0.24);
\draw[addG, line width=1.1pt, line cap=round] (4.18,0.16) -- (4.18,0.32);
\node[pff/arch/lossfirst, anchor=north] at (3.79,-0.11) {added};

\begin{scope}[opacity=0.5]
  \adecoder{3.60}{-0.60}
\end{scope}
\draw[remR, line width=1.1pt, line cap=round] (4.10,-0.60) -- (4.26,-0.60);
\node[pff/arch/lossfirst, anchor=north] at (3.79,-0.87) {removed};

\end{tikzpicture}%
\label{fig:arch-adapt}
\end{subfigure}
\vspace{-10pt}
\caption{\textbf{PFFs are capable of continually learning (a) new timesteps and (b) new products.} \textbf{(a)} New timesteps for one or more products requires a small addition to the shared field (0.2\% params) and are trained with the rest of the PFF frozen. \textbf{(b)} New products are added by training only a decoder against the frozen shared field, and deprecated products can simply be discarded. }
\label{fig:adaptation}
\vspace{-15pt}
\end{figure}

\section{Planetary Feature Fields}
\label{sec:method}

We use the term \emph{feature} to refer to any space-time-indexed array of data about the Earth. 
These features include raw observations, precomputed embeddings, and map products, such as observations from the Sentinel-2 (optical) and Sentinel-1 (radar) satellites, \emph{embeddings} (AlphaEarth (AE) \citep{alphaearth}, TESSERA \citep{tessera}), and \emph{map products} such as elevation data, canopy height, etc. 
We consider products indexed by
$m \in \{1,\ldots,M\}$. For a geographic coordinate $\mathbf{x}$ and time coordinate
$t$, let $\mathbf{y}_m(\mathbf{x},t) \in \mathbb{R}^{d_m}$ denote the observed feature from product $m$. 
Products may differ in spatial resolution, temporal coverage, dimensionality, and availability. 
We partition the geographic domain into bounded regions and learn a local function within each region. 
For product $m$, this function predicts $\hat{\mathbf{y}}_m(\mathbf{x},t) = f_m(\mathbf{x},t)$, with parameters chosen so that
$\hat{\mathbf{y}}_m(\mathbf{x},t)$ reconstructs the corresponding feature where and when the data is available (\Cref{fig:heterogeneity}).
The learned local functions replace the feature samples and form the representation used for reconstruction and downstream use. 


\providecolor{rdEmbeddings}{HTML}{6B5AA8}
\providecolor{rdObservations}{HTML}{35708E}
\providecolor{rdMaps}{HTML}{4F8A5C}
\providecolor{matOrangeD}{HTML}{EF8A2C}
\providecolor{rdVM}{HTML}{EF8A2C}
\providecolor{rdKP}{HTML}{C65A25}
\providecolor{rdSoloKP}{HTML}{168B89}
\providecolor{rdNGP}{HTML}{C0392B}
\providecolor{rdSiren}{HTML}{59616D}
\providecolor{rdRelu}{HTML}{8B3FC6}
\providecolor{rdSoloVM}{HTML}{2E9E38}
\providecolor{rdInk}{HTML}{30333A}
\providecolor{rdGrid}{HTML}{E6E7E9}
\providecolor{rdTint}{HTML}{FCF4E9}
\providecolor{rdCR}{HTML}{D32F2F}
\providecolor{rdCRBg}{HTML}{FDEEEE}
\providecolor{rdLegendBg}{HTML}{F5F5F5}

\providecolor{rdMeanGray}{HTML}{EEEEEE}
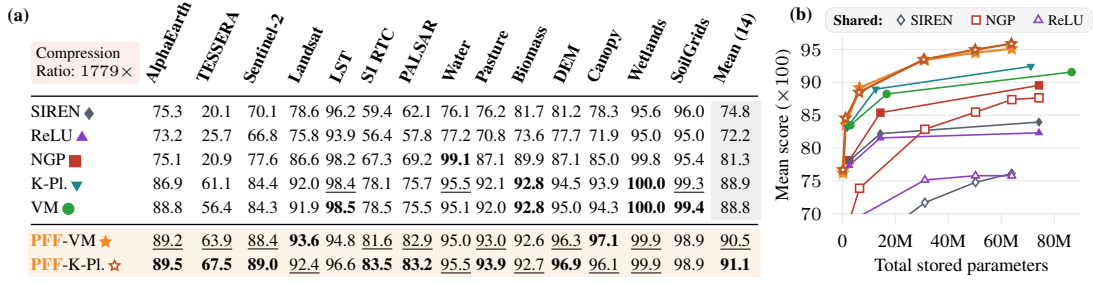
\begin{figure}[t]
\centering
\begin{minipage}{\linewidth}
\begingroup
\color{black}
\newcommand{\rdProductHeader}[2]{\rotatebox{65}{\textcolor{black}{%
  \makebox[0pt][l]{\smash{\textbf{#2}}}\phantom{#2}}}}
\pgfdeclareplotmark{rdstar}{%
  \pgfpathmoveto{\pgfpointpolar{90}{\pgfplotmarksize}}%
  \pgfpathlineto{\pgfpointpolar{126}{0.476\pgfplotmarksize}}%
  \pgfpathlineto{\pgfpointpolar{162}{\pgfplotmarksize}}%
  \pgfpathlineto{\pgfpointpolar{198}{0.476\pgfplotmarksize}}%
  \pgfpathlineto{\pgfpointpolar{234}{\pgfplotmarksize}}%
  \pgfpathlineto{\pgfpointpolar{270}{0.476\pgfplotmarksize}}%
  \pgfpathlineto{\pgfpointpolar{306}{\pgfplotmarksize}}%
  \pgfpathlineto{\pgfpointpolar{342}{0.476\pgfplotmarksize}}%
  \pgfpathlineto{\pgfpointpolar{18}{\pgfplotmarksize}}%
  \pgfpathlineto{\pgfpointpolar{54}{0.476\pgfplotmarksize}}%
  \pgfpathclose\pgfusepathqfillstroke}
\newcommand{\rdmark}[4][0]{\tikz[baseline=-0.5ex]{%
  \draw[#2,mark=#3,mark size=2pt,
    mark options={fill=#4,solid,rotate=#1,line width=0.6pt}]
    plot coordinates {(0,0)};}}
\newsavebox{\rdTableBox}
\sbox{\rdTableBox}{
\fontsize{6.5}{8}\selectfont
\setlength{\tabcolsep}{0.8pt}
\renewcommand{\arraystretch}{1.12}
\setlength{\aboverulesep}{1.5pt}
\setlength{\belowrulesep}{1.5pt}
\begin{tabular}{@{}l *{15}{c}@{}}
\smash{\tikz[baseline=(rdcr.south)]{\node[fill=rdCRBg,text=black,
  rounded corners=1.5pt,inner xsep=2pt,inner ysep=2pt,align=left,
  font=\fontsize{6}{7}\selectfont] (rdcr) {Compression\\Ratio: $1779\times$};}}
 & \rdProductHeader{rdEmbeddings}{AlphaEarth} & \rdProductHeader{rdEmbeddings}{TESSERA}
 & \rdProductHeader{rdObservations}{Sentinel-2} & \rdProductHeader{rdObservations}{Landsat}
 & \rdProductHeader{rdObservations}{LST} & \rdProductHeader{rdObservations}{S1 RTC}
 & \rdProductHeader{rdObservations}{PALSAR} & \rdProductHeader{rdMaps}{Water}
 & \rdProductHeader{rdMaps}{Pasture} & \rdProductHeader{rdMaps}{Biomass}
 & \rdProductHeader{rdMaps}{DEM} & \rdProductHeader{rdMaps}{Canopy}
 & \rdProductHeader{rdMaps}{Wetlands} & \rdProductHeader{rdMaps}{SoilGrids}
 & \rotatebox{65}{\textbf{Mean (14)}} \\
\midrule
SIREN\ \rdmark{rdSiren}{diamond*}{rdSiren} & 75.3 & 20.1 & 70.1 & 78.6 & 96.2 & 59.4 & 62.1 & 76.1 & 76.2 & 81.7 & 81.2 & 78.3 & 95.6 & 96.0 & \cellcolor{rdMeanGray}74.8 \\
ReLU\ \rdmark{rdRelu}{triangle*}{rdRelu} & 73.2 & 25.7 & 66.8 & 75.8 & 93.9 & 56.4 & 57.8 & 77.2 & 70.8 & 73.6 & 77.7 & 71.9 & 95.0 & 95.0 & \cellcolor{rdMeanGray}72.2 \\
NGP\ \rdmark{rdNGP}{square*}{rdNGP} & 75.1 & 20.9 & 77.6 & 86.6 & 98.2 & 67.3 & 69.2 & \textbf{99.1} & 87.1 & 89.9 & 87.1 & 85.0 & 99.8 & 95.4 & \cellcolor{rdMeanGray}81.3 \\
K-Pl.\ \rdmark[180]{rdSoloKP}{triangle*}{rdSoloKP} & 86.9 & 61.1 & 84.4 & 92.0 & \underline{98.4} & 78.1 & 75.7 & \underline{95.5} & 92.1 & \textbf{92.8} & 94.5 & 93.9 & \textbf{100.0} & \underline{99.3} & \cellcolor{rdMeanGray}88.9 \\
VM\ \rdmark{rdSoloVM}{*}{rdSoloVM} & 88.8 & 56.4 & 84.3 & 91.9 & \textbf{98.5} & 78.5 & 75.5 & 95.1 & 92.0 & \textbf{92.8} & 95.0 & 94.3 & \textbf{100.0} & \textbf{99.4} & \cellcolor{rdMeanGray}88.8 \\
\midrule
\rowcolor{rdTint}
\textcolor{matOrangeD}{\textbf{PFF}}-VM\ \rdmark{rdVM}{rdstar}{rdVM} & \underline{89.2} & \underline{63.9} & \underline{88.4} & \textbf{93.6} & 94.8 & \underline{81.6} & \underline{82.9} & 95.0 & \underline{93.0} & 92.6 & \underline{96.3} & \textbf{97.1} & \underline{99.9} & 98.9 & \underline{90.5} \\
\rowcolor{rdTint}
\textcolor{matOrangeD}{\textbf{PFF}}-K-Pl.\ \rdmark{rdKP}{rdstar}{white} & \textbf{89.5} & \textbf{67.5} & \textbf{89.0} & \underline{92.4} & 96.6 & \textbf{83.5} & \textbf{83.2} & \underline{95.5} & \textbf{93.9} & \underline{92.7} & \textbf{96.9} & \underline{96.1} & \underline{99.9} & 98.9 & \textbf{91.1} \\
\end{tabular}}
\pgfmathsetlengthmacro{\rdPlotWidth}{\linewidth-\wd\rdTableBox-8pt}
\pgfmathsetlengthmacro{\rdUpperHeight}{\ht\rdTableBox+\dp\rdTableBox-23pt-11.65pt}
\pgfmathsetlengthmacro{\rdLabelHeight}{(\ht\rdTableBox+\dp\rdTableBox-23pt-11.65pt)/2}
\begin{minipage}[t]{\wd\rdTableBox}
\vspace{0pt}
\begin{tikzpicture}
\path[use as bounding box] (0,0) rectangle (\wd\rdTableBox,\ht\rdTableBox+\dp\rdTableBox);

\node[anchor=south west,inner sep=0pt,outer sep=0pt] at (0,0) {\usebox{\rdTableBox}};
\node[anchor=north west,inner sep=0pt,font=\scriptsize\bfseries,overlay] at (0,\ht\rdTableBox+\dp\rdTableBox) {(a)};
\end{tikzpicture}\par
\end{minipage}\hfill
\begin{minipage}[t]{\rdPlotWidth}
\vspace{0pt}
\pgfmathsetlengthmacro{\rdAxisWidth}{\linewidth-17pt}
\pgfplotsset{
  rd shared/.style={solid,line width=0.85pt,mark size=1.8pt},
  rd solo/.style={solid,line width=0.6pt,mark size=1.3pt},
  rd axis/.style={scale only axis,width=\rdAxisWidth,
    scaled ticks=false,
    tick label style={font=\scriptsize},
    label style={font=\small},
    tick align=outside,tick style={rdGrid},
    axis line style={rdGrid,line width=0.4pt},
    ymajorgrids=true,grid style={rdGrid,line width=0.3pt},
    minor tick num=0,clip=true},
}
\makebox[\linewidth][l]{\begin{tikzpicture}
\begin{axis}[
  rd axis,at={(0,0)},anchor=south west,height=\rdUpperHeight,
  xmin=0,xmax=90000000,ymin=70,ymax=97,ytick={70,75,80,85,90,95},
  axis lines*=left,
  xtick={0,20000000,40000000,60000000,80000000},
  xticklabels={0,20M,40M,60M,80M},
  xlabel={\scriptsize{Total stored parameters}},
  x label style={at={(axis description cs:0.5,-0.198)},anchor=north},
]
\addplot[rd shared,rdVM,mark=rdstar,mark options={fill=rdVM,solid}]
coordinates {(259714,76.187173) (1299933,83.556194) (6475948,89.173774)
             (31143229,93.414209) (50273522,94.508697) (63964774,95.112598)};
\addplot[rd shared,rdKP,mark=rdstar,mark options={fill=white,solid}]
coordinates {(215784,76.701852) (1191144,84.548906) (6295368,88.490249)
             (30725864,93.482485) (50134344,94.974569) (63802824,95.880059)};
\addplot[rd solo,rdSoloKP,mark=triangle*,mark options={fill=rdSoloKP,solid,rotate=180}]
coordinates {(1858370,83.151157) (12585378,88.969643) (71134946,92.431792)};

\addplot[rd solo,rdSoloVM,mark=*,mark options={fill=rdSoloVM,solid}]
coordinates {(3043924,83.484) (16725070,88.229) (86546848,91.564)};
\addplot[rd solo,rdNGP,mark=square*,mark options={fill=rdNGP,solid}]
coordinates {(2612882,78.152280) (14345634,85.387731) (74211218,89.541328)};
\addplot[rd solo,rdSiren,mark=diamond*,mark options={fill=rdSiren,solid}]
coordinates {(2612842,78.093703) (14340316,82.194864) (74213945,83.954200)};
\addplot[rd solo,rdRelu,mark=triangle*,mark options={fill=rdRelu,solid}]
coordinates {(2610700,77.363875) (14334149,81.544215) (74223091,82.321791)};

\addplot[rd solo,rdNGP,mark=square*,mark size=1.5pt,mark options={fill=white,solid}]
coordinates {(259716,51.224841) (1299944,67.256096) (6475948,73.881552) (31143224,82.860800) (50273520,85.471700) (63964778,87.374100) (74223091,87.653795)};
\addplot[rd solo,rdSiren,mark=diamond*,mark size=1.6pt,mark options={fill=white,solid}]
coordinates {(258728,40.965078) (1296868,49.891000) (6466788,64.288202) (31138020,71.694695) (50268660,74.791045) (63956724,76.118949)};
\addplot[rd solo,rdRelu,mark=triangle*,mark size=1.6pt,mark options={fill=white,solid}]
coordinates {(258041,48.712770) (1296308,56.217714) (6472082,69.599188) (31130408,75.139813) (50264998,75.793653) (63955402,75.777850)};
\end{axis}
\newcommand{\rdLegendEntries}[1]{%
\node[anchor=east,font=\fontsize{5.5}{6}\selectfont,text=black,inner sep=0pt]
  (lgrelu#1) at (0.5*\rdAxisWidth+46.2pt,\rdUpperHeight+6.15pt) {ReLU};
\draw[rdRelu,line width=0.6pt] plot[mark=triangle*,mark size=1.8pt,mark options={fill=white,solid}] coordinates {([xshift=-3.5pt]lgrelu#1.west)};
\node[anchor=east,font=\fontsize{5.5}{6}\selectfont,text=black,inner sep=0pt]
  (lgngp#1) at ([xshift=-12pt]lgrelu#1.west) {NGP};
\node[draw=rdNGP,fill=white,line width=0.6pt,inner sep=0pt,minimum size=3pt]
  at ([xshift=-3.5pt]lgngp#1.west) {};
\node[anchor=east,font=\fontsize{5.5}{6}\selectfont,text=black,inner sep=0pt]
  (lgsiren#1) at ([xshift=-12pt]lgngp#1.west) {SIREN};
\draw[rdSiren,line width=0.6pt] plot[mark=diamond*,mark size=1.8pt,mark options={fill=white,solid}] coordinates {([xshift=-3.5pt]lgsiren#1.west)};
\node[anchor=east,font=\fontsize{5.5}{6}\selectfont,text=black,inner sep=0pt]
  (lgshared#1) at ([xshift=-10pt]lgsiren#1.west) {\textbf{Shared:}};}
\rdLegendEntries{}
\fill[rdLegendBg,rounded corners=3pt]
  ([shift={(-3pt,-2.6pt)}]lgshared.west |- lgrelu.south) rectangle ([shift={(3pt,2.6pt)}]lgrelu.east |- lgrelu.north);
\draw[rdGrid,line width=0.35pt,rounded corners=3pt]
  ([shift={(-3pt,-2.6pt)}]lgshared.west |- lgrelu.south) rectangle ([shift={(3pt,2.6pt)}]lgrelu.east |- lgrelu.north);
\rdLegendEntries{b}
\node[rotate=90,font=\scriptsize,anchor=south,inner sep=1pt]
  at (-18pt,\rdLabelHeight) {Mean score ($\times100$)};
\node[anchor=north west,inner sep=0pt,font=\scriptsize\bfseries,overlay] at (-20pt,\rdUpperHeight+11pt) {(b)};
\end{tikzpicture}}
\end{minipage}
\endgroup
\par\vspace{2pt}
\vspace{-2pt}
\caption{\textbf{Reconstruction results on PFFBench.}
  \textbf{(a)} One shared PFF beats compression rate-matched single-product neural fields on EO product reconstructions tested across 100 global \textsc{PFFBench} tiles (see \Cref{fig:pffbench-selection}).
  \textbf{(b)} Rate-distortion curves on the mean reconstruction metric (10-tile subset). Sharing the explicit representation is uniquely beneficial to factored feature volumes (NGP is an explicit--implicit representation, but not a factored feature volume). Full table shown in \Cref{tab:bitweighted}.}

\label{fig:rd-fairshare-ladder}
\end{minipage}
\end{figure}
\subsection{Constructing PFFs }
\label{sec:field_design}
We learn a separate field for each region to concentrate representation capacity on local detail. 
Global coordinate models represent the Earth within one function, using spatial encodings such as spherical bases or tessellations \citep{sh,rao2026localized,cher2026tte}. 
PFFs instead route each query to its tile and transform the location into that tile's Cartesian coordinates. 
Each field therefore models a bounded region rather than the geometry of the entire Earth. 
We rescale each tile's spatial extent to $[0,1]^2$, so every field uses the same relative coordinates within its own region.

Within each region, the field must preserve local detail while representing products with different output spaces. 
Explicit neural fields fit data by optimizing values on coarse or sparse grids \citep{yu2021plenoxels,grids} or local primitives such as Gaussians \citep{kerbl3Dgaussians}. 
This works well when the goal is to reconstruct a single input, such as a single object or scene, but provides no structure to share or separate information across the multiple EO products. 
To separate common structure from product-specific outputs, PFFs use a \textit{hybrid} field. 
The explicit part is a shared, coordinate-indexed set of arrays covering space and time. 
The implicit part consists of small product-specific MLPs that map this representation to each product.

Our explicit field must (i) compactly preserve spatial and temporal structure, (ii) allocate capacity asymmetrically across dense spatial axes and sparse temporal observations, and (iii) share information across products at high compression ratios.
Requirement (ii) is specific and important to EO. 
Unlike most video data or dynamic scenes where time is sampled densely at the frame level, EO products are far more \emph{anisotropic}: a region may contain thousands of samples along each spatial axis but only a handful of annual observations along time. 

We store a region’s features across space, time, and products in a set of small, learned arrays rather than one large grid. 
These arrays are the \emph{factors} in a \emph{factored feature volume} \citep{tilted}. 
To query a location and time, we read and combine values from the factors into a shared feature vector.
The factors let us choose different grid sizes for space and time, matching the dense spatial samples and sparse timesteps of EO data. 
This replaces storing $N_xN_yN_td_h$ values of a dense grid, with spatial resolution $N_x,N_y$, temporal resolution $N_t$, and the shared feature dimension $d_h$.

We now fix one region and omit its index. Let $\mathbf{h}(\mathbf{x},t)\in\mathbb{R}^{d_h}$ denote the feature computed from its factors and $g_m$ the decoder for product $m$:
\begin{equation*}
    f_m(\mathbf{x},t)
    =
    g_m\!\left(\mathbf{h}(\mathbf{x},t)\right).
\end{equation*}
The shared feature $\mathbf{h}$ is an internal representation used by the product decoders. We construct $\mathbf{h}$ using either the vector--matrix (VM) decomposition of TensoRF \citep{chen2022tensorf} or the three-plane factorization of K-Planes \citep{fridovich2023kplanes}.
Both interpolate learned factors at the query coordinates, but differ in how these factors are stored and combined.

Writing $\mathbf{x}=(x,y)$, TensoRF(VM) pairs each scalar-valued plane factor $M_r^{bc}$ with a line factor $v_r^a$ along the remaining axis. 
Each orientation has $R_a$ components with learned basis vectors $\mathbf{b}_r^a\in\mathbb{R}^{d_h}$. 
K-Planes instead stores vector-valued plane factors $\mathbf{P}^{xy}$, $\mathbf{P}^{xt}$, and $\mathbf{P}^{yt}$ and combines them by elementwise multiplication, denoted $\odot$. 
At a single resolution, the TensoRF(VM) and K-Planes constructions are:
\par\smallskip
\begingroup
\small
\setlength{\jot}{1pt}
\setlength{\abovedisplayskip}{-1pt} 
\setlength{\abovedisplayshortskip}{-1pt}
\setlength{\belowdisplayskip}{4pt}
\setlength{\belowdisplayshortskip}{4pt}
\begin{subequations}
\label{eq:pff_factorizations}
\noindent
\begin{minipage}[t]{0.64\linewidth}
\vspace{0pt}
\centering
\textbf{TensoRF(VM)}\\[-1pt]
{\footnotesize Plane--line factors\par}
\vspace{-2pt}
\begin{equation*}
\begin{aligned}
&\textstyle
\mathbf{h}(x,y,t)
=
\sum_{r=1}^{R_x}
v_r^x(x)\,M_r^{yt}(y,t)\,\mathbf{b}_r^x
\\
&\textstyle
\;{}+
\sum_{r=1}^{R_y}
v_r^y(y)\,M_r^{xt}(x,t)\,\mathbf{b}_r^y
+
\sum_{r=1}^{R_t}
v_r^t(t)\,M_r^{xy}(x,y)\,\mathbf{b}_r^t .
\end{aligned}
\label{eq:pff_vm}
\end{equation*}
\end{minipage}\hfill%
\begin{minipage}[t]{0.33\linewidth}
\vspace{0pt}
\centering
\textbf{K-Planes}\\[-1pt]
{\footnotesize Plane factors\par}
\begin{equation*}
\begin{aligned}
&\textstyle
\vphantom{\sum_{r=1}^{R_x}}
\mathbf{h}(x,y,t)
=
\mathbf{P}^{xy}(x,y)
\\
&\textstyle
\vphantom{\sum_{r=1}^{R_t}}
\;{}\odot
\mathbf{P}^{xt}(x,t)
\odot
\mathbf{P}^{yt}(y,t).
\end{aligned}
\label{eq:pff_kplanes}
\end{equation*}
\end{minipage}
\par
\end{subequations}
\endgroup
Line and plane factors are queried with linear and bilinear interpolation. 
Spatial and temporal resolutions are chosen independently, with nine temporal entries for 2017--2025 in both constructions. 
Our canonical PFF uses VM with spatial resolution $S=955$ and 65 vector--matrix components per orientation ($R_x=R_y=R_t=65$). 
The combined factors produce a shared feature of dimension $d_h=64$. 
Our K-Planes PFF uses four spatial scales at resolutions $147$, $294$, $588$, and $1176$. 
Each plane stores 32 channels, and concatenating the features across scales gives $d_h=128$.

\subsection{Optimizing PFFs} 
\usetikzlibrary{arrows.meta}
\providecolor{ink}{HTML}{3E434A}
\providecolor{matOrangeD}{HTML}{EF8A2C}
\providecolor{slotOn}{HTML}{FF9100}
\providecolor{replay}{HTML}{F50057}

\begin{wrapfigure}{r}{0.50\linewidth}
  \vspace{-\intextsep}
  \centering
  \begin{tikzpicture}[x=1cm, y=1cm]
  \path[fill=aeV!30, draw=aeV!70!black, line width=0.35pt, rounded corners=0.8pt] (0.355,0.165) rectangle (0.935,0.745);
  \path[fill=aeV!30, draw=aeV!70!black, line width=0.35pt, rounded corners=0.8pt] (0.270,0.110) rectangle (0.850,0.690);
  \path[fill=aeV!30, draw=aeV!70!black, line width=0.35pt, rounded corners=0.8pt] (0.185,0.055) rectangle (0.765,0.635);
  \begin{scope}[shift={(0.100,0.000)}, scale=0.58]
    \clip[rounded corners=0.04] (0,0) rectangle (1,1);
    \tileAE
  \end{scope}
  \path[draw=aeV!70!black, line width=0.4pt, rounded corners=0.8pt] (0.100,0.000) rectangle (0.680,0.580);
  \path[fill=aeV!85!black] (0.114,-0.170) rectangle (0.231,-0.085);
  \path[fill=aeV!85!black] (0.259,-0.170) rectangle (0.376,-0.085);
  \path[fill=aeV!85!black] (0.404,-0.170) rectangle (0.521,-0.085);
  \path[fill=aeV!85!black] (0.549,-0.170) rectangle (0.666,-0.085);
  \node[anchor=north, align=center, font=\tiny, text=ink, inner sep=1.5pt] at (0.470,-0.230) {all\\timesteps};
  \path[fill=sTwoB!30, draw=sTwoB!70!black, line width=0.35pt, rounded corners=0.8pt] (1.655,0.165) rectangle (2.235,0.745);
  \path[draw=black!28, line width=0.35pt, dash pattern=on 1.1pt off 1.1pt, rounded corners=0.8pt] (1.570,0.110) rectangle (2.150,0.690);
  \path[draw=black!28, line width=0.35pt, dash pattern=on 1.1pt off 1.1pt, rounded corners=0.8pt] (1.485,0.055) rectangle (2.065,0.635);
  \begin{scope}[shift={(1.400,0.000)}, scale=0.58]
    \clip[rounded corners=0.04] (0,0) rectangle (1,1);
    \tileSTwo
  \end{scope}
  \path[draw=sTwoB!70!black, line width=0.4pt, rounded corners=0.8pt] (1.400,0.000) rectangle (1.980,0.580);
  \path[fill=sTwoB!85!black] (1.414,-0.170) rectangle (1.531,-0.085);
  \path[fill=sTwoB!85!black] (1.559,-0.170) rectangle (1.676,-0.085);
  \path[fill=sTwoB!85!black] (1.704,-0.170) rectangle (1.821,-0.085);
  \path[fill=sTwoB!85!black] (1.849,-0.170) rectangle (1.966,-0.085);
  \node[anchor=north, align=center, font=\tiny, text=ink, inner sep=1.5pt] at (1.770,-0.230) {missing\\timesteps};
  \path[fill=sOneB!30, draw=sOneB!70!black, line width=0.35pt, rounded corners=0.8pt] (2.955,0.165) rectangle (3.535,0.745);
  \path[fill=sOneB!30, draw=sOneB!70!black, line width=0.35pt, rounded corners=0.8pt] (2.870,0.110) rectangle (3.450,0.690);
  \path[fill=sOneB!30, draw=sOneB!70!black, line width=0.35pt, rounded corners=0.8pt] (2.785,0.055) rectangle (3.365,0.635);
  \begin{scope}[shift={(2.700,0.000)}, scale=0.58]
    \clip[rounded corners=0.04] (0,0) rectangle (1,1);
    \tileSOne
  \end{scope}
  \path[draw=sOneB!70!black, line width=0.4pt, rounded corners=0.8pt] (2.700,0.000) rectangle (3.280,0.580);
  \path[fill=sOneB!85!black] (2.714,-0.170) rectangle (2.831,-0.085);
  \path[fill=sOneB!85!black] (2.859,-0.170) rectangle (2.976,-0.085);
  \path[draw=black!30, line width=0.3pt] (3.004,-0.170) rectangle (3.121,-0.085);
  \path[draw=black!30, line width=0.3pt] (3.149,-0.170) rectangle (3.266,-0.085);
  \node[anchor=north, align=center, font=\tiny, text=ink, inner sep=1.5pt] at (3.070,-0.230) {missing\\channels};
  \path[fill=lsB!30, draw=lsB!70!black, line width=0.35pt, rounded corners=0.8pt] (4.255,0.165) rectangle (4.835,0.745);
  \path[fill=white, draw=black!35, line width=0.35pt, dash pattern=on 1.1pt off 1.1pt] (4.305,0.465) rectangle (4.575,0.695);
  \path[fill=lsB!30, draw=lsB!70!black, line width=0.35pt, rounded corners=0.8pt] (4.170,0.110) rectangle (4.750,0.690);
  \path[fill=white, draw=black!35, line width=0.35pt, dash pattern=on 1.1pt off 1.1pt] (4.220,0.410) rectangle (4.490,0.640);
  \path[fill=lsB!30, draw=lsB!70!black, line width=0.35pt, rounded corners=0.8pt] (4.085,0.055) rectangle (4.665,0.635);
  \path[fill=white, draw=black!35, line width=0.35pt, dash pattern=on 1.1pt off 1.1pt] (4.135,0.355) rectangle (4.405,0.585);
  \begin{scope}[shift={(4.000,0.000)}, scale=0.58]
    \clip[rounded corners=0.04] (0,0) rectangle (1,1);
    \tileLandsat
  \end{scope}
  \path[fill=white, draw=black!35, line width=0.35pt, dash pattern=on 1.1pt off 1.1pt] (4.050,0.300) rectangle (4.320,0.530);
  \path[draw=lsB!70!black, line width=0.4pt, rounded corners=0.8pt] (4.000,0.000) rectangle (4.580,0.580);
  \path[fill=lsB!85!black] (4.014,-0.170) rectangle (4.131,-0.085);
  \path[fill=lsB!85!black] (4.159,-0.170) rectangle (4.276,-0.085);
  \path[fill=lsB!85!black] (4.304,-0.170) rectangle (4.421,-0.085);
  \path[fill=lsB!85!black] (4.449,-0.170) rectangle (4.566,-0.085);
  \node[anchor=north, align=center, font=\tiny, text=ink, inner sep=1.5pt] at (4.370,-0.230) {missing\\areas};
  \path[draw=black!28, line width=0.35pt, dash pattern=on 1.1pt off 1.1pt, rounded corners=0.8pt] (5.555,0.165) rectangle (6.135,0.745);
  \path[draw=black!28, line width=0.35pt, dash pattern=on 1.1pt off 1.1pt, rounded corners=0.8pt] (5.470,0.110) rectangle (6.050,0.690);
  \path[draw=black!28, line width=0.35pt, dash pattern=on 1.1pt off 1.1pt, rounded corners=0.8pt] (5.385,0.055) rectangle (5.965,0.635);
  \path[draw=black!28, line width=0.35pt, dash pattern=on 1.1pt off 1.1pt, rounded corners=0.8pt] (5.300,0.000) rectangle (5.880,0.580);
  \path[draw=black!30, line width=0.3pt] (5.314,-0.170) rectangle (5.431,-0.085);
  \path[draw=black!30, line width=0.3pt] (5.459,-0.170) rectangle (5.576,-0.085);
  \path[draw=black!30, line width=0.3pt] (5.604,-0.170) rectangle (5.721,-0.085);
  \path[draw=black!30, line width=0.3pt] (5.749,-0.170) rectangle (5.866,-0.085);
  \node[anchor=north, align=center, font=\tiny, text=ink, inner sep=1.5pt] at (5.670,-0.230) {absent\\product};
  \draw[black!45, line width=0.4pt, -{Stealth[length=2pt]}] (0.16,0.86) -- (0.62,1.03);
  \node[anchor=west, font=\tiny, text=black!55, inner sep=1.5pt] at (0.64,1.03) {time};
  \node[anchor=east, font=\tiny, text=black!55, inner sep=1.5pt] at (0.04,-0.13) {channels};
  \end{tikzpicture}
  \caption{\textbf{PFFs are robust to missing data.} We train with products that may be observed at every timestep, miss entire timesteps, carry only some of its channels, miss spatial information in one or more years, or be entirely absent from the region.}
  \label{fig:heterogeneity}
  \vspace{-1.0\baselineskip}
\end{wrapfigure}
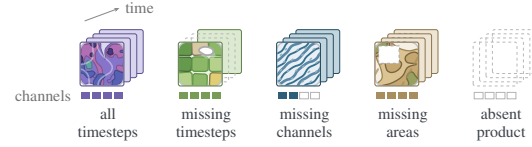 

\label{sec:pff-training}

PFFs are trained jointly on datasets of products that may omit arbitrary timesteps, channels, spatial regions, or entire products (\Cref{fig:heterogeneity}). 
Each product supervises the shared field wherever observations are available. 
We mask missing values, so training does not require all products to be observed at the same location and time.
Coarse products are mapped to a common $10\,\mathrm{m}$ grid by nearest-neighbor resampling, preserving source-cell and nodata values.
Each coarse observation therefore supervises its spatial footprint without introducing interpolated targets, which is essential when reconstructing derived map products.
We use per-channel standardized mean-squared error for Euclidean-valued products and cosine distortion $1-\cos(\hat{\mathbf{y}}_m,\mathbf{y}_m)$ for unit-normalized embeddings. 
We average each product loss over its observed targets and combine the losses with equal weight.

A full regional product stack occupies a median of 227.5 GB in memory, making simple data loading to memory impractical. 
We instead maintain a fixed-capacity GPU buffer of randomly selected spatial block--year pairs, loading the same window from all available products through partial Cloud-Optimized GeoTIFF (COG) reads \citep{cog}. 
Each batch samples valid coordinates uniformly from the buffer and combines observed product losses, masking missing targets. 
Refreshing the resident blocks exposes the field to new observations while keeping memory use bounded (\Cref{fig:blockstream-sampling}).
Training settings are detailed in \Cref{sec:appendix_pff_training}.

\usetikzlibrary{arrows.meta,calc}
\providecolor{pffRowBg}{HTML}{FCF4E9}  
\providecolor{adPff}{HTML}{FF6F00}  
\providecolor{adPffMid}{HTML}{FFA043}  
\providecolor{adPffLite}{HTML}{FFCC8F} 
\providecolor{adPretrained}{HTML}{FFAB91} 
\providecolor{adPffTxt}{HTML}{C25E00}  
\providecolor{adCopy}{HTML}{C2185B}
\providecolor{adLin}{HTML}{5E35B1}
\providecolor{adSiren}{HTML}{59616D}
\providecolor{adMlp}{HTML}{8B3FC6}
\providecolor{adLoss}{HTML}{C0392B}
\providecolor{adAE}{HTML}{168B89}
\providecolor{adTE}{HTML}{8B3FC6}
\begin{figure}[t]
\begin{minipage}[t]{0.50\linewidth}
\vspace{0pt}
\centering
{\fontsize{9.5}{11.5}\selectfont \textbf{(a)} New Timesteps\par}
\vspace{-0.5mm}
\begin{tikzpicture}
\pgfplotsset{
  admark/.style={only marks,mark=*,mark size=2.7pt,
    mark options={draw=black!80,line width=0.45pt}},
  aderr/.style={only marks,mark=none,
    error bars/.cd,x dir=both,x explicit,y dir=both,y explicit,
    error bar style={line width=0.45pt},
    error mark options={rotate=90,mark size=1.25pt,line width=0.45pt},/pgfplots/.cd},
}
\tikzset{
  adlab/.style={font=\scriptsize,inner sep=2pt,align=left},
}
\begin{axis}[
  scale only axis,width=\dimexpr\linewidth-42pt\relax,height=41mm,
  xmin=-82,xmax=107,ymin=44.5,ymax=96.8,
  xtick={-50,0,50,100},xticklabels={\textminus{}50,0,50,100},
  ytick={50,60,70,80,90},yticklabels={50,60,70,80,90},
  tick label style={font=\scriptsize},
  xticklabel style={yshift=2pt},yticklabel style={xshift=2pt},
  tick style={draw=none},
  xmajorgrids,ymajorgrids,
  grid style={black!10,line width=0.3pt},
  axis line style={black!55,line width=0.4pt},
  axis background/.style={fill=white},
  ylabel={\textbf{New} Time Recon.\ R\textsuperscript{2} (\texttimes100) $\to$},
  ylabel style={font=\small},
  xlabel={\textbf{Old} R$^2$ Retained \emph{After} New Time Adaptation (\%) $\rightarrow$},
  xlabel style={font=\small,align=center,xshift=-7pt},
  title={Mean \textpm{} 1 SE across 3 tiles},
  title style={font=\scriptsize,align=center,xshift=-4pt,yshift=-6.5pt},
  clip=true,
]
\addplot[aderr,color=adPretrained!65,error bars/error mark options={rotate=90,mark size=1.1pt,line width=0.4pt,draw=adPretrained!65}] coordinates {(99.726979,88.694949) +- (0.032346,1.417054)};
\addplot[aderr,color=adPffTxt!48,error bars/error mark options={rotate=90,mark size=1.1pt,line width=0.4pt,draw=adPffTxt!48}] coordinates {(39.102749,93.786352) +- (9.088514,1.246086)};
\addplot[aderr,color=adPffTxt!48,error bars/error mark options={rotate=90,mark size=1.1pt,line width=0.4pt,draw=adPffTxt!48}] coordinates {(-47.523818,87.573827) +- (30.076781,0.560143)};
\addplot[aderr,color=adCopy!65,error bars/error mark options={rotate=90,mark size=1.1pt,line width=0.4pt,draw=adCopy!65}] coordinates {(100.000000,53.975526) +- (0.000000,7.583018)};
\addplot[aderr,color=adPff!65,error bars/error mark options={rotate=90,mark size=1.1pt,line width=0.4pt,draw=adPff!65}] coordinates {(100.000000,81.027299) +- (0.000000,1.267788)};
\addplot[aderr,color=adSiren!55,error bars/y dir=none,error bars/error mark options={rotate=90,mark size=1.1pt,line width=0.4pt,draw=adSiren!55}] coordinates {(-33.000000,78.550000) +- (17.320508,0)};
\addplot[aderr,color=adSiren!32,error bars/x dir=none,error bars/error bar style={line width=0.3pt},error bars/error mark options={rotate=90,mark size=0.75pt,line width=0.3pt,draw=adSiren!32}] coordinates {(-33.000000,78.550000) +- (0,2.684679)};
\addplot[aderr,color=adMlp!55,error bars/y dir=none,error bars/error mark options={rotate=90,mark size=1.1pt,line width=0.4pt,draw=adMlp!55}] coordinates {(35.350000,74.100000) +- (6.148780,0)};
\addplot[aderr,color=adMlp!32,error bars/x dir=none,error bars/error bar style={line width=0.3pt},error bars/error mark options={rotate=90,mark size=0.75pt,line width=0.3pt,draw=adMlp!32}] coordinates {(35.350000,74.100000) +- (0,3.752777)};
\addplot[only marks,mark=*,mark size=3.0pt,
  mark options={draw=adPretrained,fill=white,line width=1.1pt}]
  coordinates {(99.726979,88.694949)};
\addplot[admark,mark options={fill=adPffMid}] coordinates {(39.102749,93.786352)};
\addplot[admark,mark options={fill=adPffLite}] coordinates {(-47.523818,87.573827)};
\addplot[admark,mark options={fill=adCopy}] coordinates {(100.000000,53.975526)};
\addplot[admark,mark options={fill=adPff}] coordinates {(100.000000,81.027299)};
\addplot[admark,mark=diamond*,mark options={fill=adSiren}] coordinates {(-33,78.55)};
\addplot[admark,mark=triangle*,mark options={fill=adMlp}] coordinates {(35.35,74.1)};
\node[adlab,text=adPffTxt,font=\fontsize{6.5}{7}\selectfont,anchor=east,align=right,inner sep=1pt]
  at (axis cs:94.8,88.694949) {Pre-trained\\[-1pt]Old + New};
\node[adlab,text=adPffTxt,anchor=east]
  at (axis cs:35,92.7) {PFF-FT-All\\[-1pt]{\fontsize{6}{6.7}\selectfont\textcolor{adLoss}{\textbf{61\% forgetting}}}};
\node[adlab,text=adPffTxt,anchor=south,align=center]
  at (axis cs:-47.523818,88.6) {PFF-FT-Dec\\[-1pt]{\fontsize{6}{6.7}\selectfont\textcolor{adLoss}{\textbf{148\% forgetting}}}};
\node[adlab,text=adCopy,anchor=east]
  at (axis cs:94,53.975526) {Time Copy (previous year)\\[-1pt]{\fontsize{6}{6.7}\selectfont\textcolor{black}{\textbf{0\% forgetting}}}};
\node[adlab,text=adPff,font=\scriptsize\bfseries,anchor=east]
  at (axis cs:94,81.027299) {PFF-Append\\[-1pt]{\fontsize{6}{6.7}\selectfont\textcolor{black}{0\% forgetting}}};
\node[adlab,text=adSiren,anchor=north,align=center]
  at (axis cs:-33,74.7) {SIREN-FT\\[-1pt]{\fontsize{6}{6.7}\selectfont\textcolor{adLoss}{\textbf{133\% forgetting}}}};
\node[adlab,text=adMlp,anchor=north west]
  at (axis cs:40,73) {MLP-FT\\[-1pt]{\fontsize{6}{6.7}\selectfont\textcolor{adLoss}{\textbf{65\% forgetting}}}};
\end{axis}
\end{tikzpicture}
\end{minipage}\hfill
\begin{minipage}[t]{0.50\linewidth}
\vspace{0pt}%
\centering
{\fontsize{9.5}{11.5}\selectfont \textbf{(b)} New Products\par}
\vspace{1.2mm}
{\fontsize{6}{7}\selectfont
\tikz[baseline=-0.5ex]{\draw[adPff,line width=0.9pt](0,0)--(0.36,0);
  \node[circle,fill=adPff,draw=black!80,line width=0.45pt,inner sep=0pt,minimum size=4.2pt] at (0.18,0){};}\,New decoder only (PFF frozen)\qquad
\tikz[baseline=-0.5ex]{\draw[adPffMid,line width=0.9pt](0,0)--(0.36,0);
  \node[rectangle,fill=adPffMid,draw=black!80,line width=0.45pt,inner sep=0pt,minimum size=3.8pt] at (0.18,0){};}\,Fine-tune PFF + new decoder\par}
\vspace{1mm}
\begin{tikzpicture}
\pgfplotsset{
  adax/.style={
    scale only axis,
    tick label style={font=\fontsize{6}{7}\selectfont},
    xticklabel style={yshift=2pt},yticklabel style={xshift=2pt},
    tick style={draw=none},
    ymajorgrids,
    grid style={black!10,line width=0.3pt},
    axis line style={black!55,line width=0.4pt},
    axis background/.style={fill=white},
    ylabel style={font=\fontsize{6}{7}\selectfont,
      at={(axis description cs:-0.11,0.5)},anchor=south,inner sep=0pt},
    xlabel style={font=\fontsize{6}{7}\selectfont},
    title style={font=\fontsize{6.5}{7.5}\selectfont,yshift=-3pt},
    clip=true,
  },
  adsweep/.style={adax,
    width=\dimexpr0.40\linewidth-27pt\relax,height=17mm,
    ylabel style={at={(axis description cs:-0.22,0.5)},anchor=south,inner sep=0pt},
    xmin=0.72,xmax=85,
    xtick={1,5,20,60},xticklabels={1,5,20,100},xmajorgrids,
    clip=false,
  },
  addec/.style={adPff,line width=0.9pt,mark=*,mark size=2.1pt,
    mark options={solid,fill=adPff,draw=black!80,line width=0.45pt}},
  adft/.style={adPffMid,line width=0.9pt,mark=square*,mark size=1.9pt,
    mark options={solid,fill=adPffMid,draw=black!80,line width=0.45pt}},
}
\tikzset{
  adref/.style={font=\fontsize{5.5}{6}\selectfont,text=black!70,inner sep=1pt},
  adforget/.style={font=\fontsize{4.5}{5}\selectfont\bfseries,text=adLoss,inner sep=1pt},
  adftbar/.style={fill=adPffMid!75,
    draw=adPffTxt!65,line width=0.35pt},
}
\begin{axis}[adsweep,xmode=log,name=ga,at={(0,0)},anchor=north west,
  ymin=79.5,ymax=93.5,ytick={80,85,90},yticklabels={80,85,90},
  xticklabel=\empty,
  ylabel={Recon.\ R\textsuperscript{2} (\texttimes100) $\to$},
  title={(i) \textbf{New} product: Galileo (GA)}]
\draw[black!55,densely dashed,line width=0.55pt](axis cs:0.72,90.5)--(axis cs:85,90.5);
\node[adref,anchor=south west] at (axis cs:0.78,90.7){Pre-train w/ GA};
\addplot[adft] coordinates {(1,81.8)(5,86.2)(20,87.3)};
\addplot[addec] coordinates {(1,82.5)(5,84.6)(20,84.7)};
\addplot[adft,dash pattern=on 0.7pt off 0.7pt,mark=none] coordinates {(20,87.3)(60,87.8)};
\addplot[addec,dash pattern=on 0.7pt off 0.7pt,mark=none] coordinates {(20,84.7)(60,84.8)};
\addplot[adft,only marks] coordinates {(60,87.8)};
\addplot[addec,only marks] coordinates {(60,84.8)};
\node[font=\fontsize{5}{5}\selectfont,text=black!60,fill=white,inner sep=0pt]
  at (axis cs:34,79.5) {/\!/};
\node[font=\fontsize{5}{5}\selectfont,text=black!60,fill=white,inner sep=0pt]
  at (axis cs:34,93.5) {/\!/};
\end{axis}
\begin{axis}[adsweep,xmode=log,name=gb,at={(ga.south west)},anchor=north west,yshift=-3mm,
  ymin=33.5,ymax=45.6,ytick={35,40,45},yticklabels={35,40,45},
  ylabel={PASTIS mIoU (\%) $\to$},
  xlabel={Training pixels (\%)},
  xlabel style={at={(axis description cs:0.5,0)},anchor=north,yshift=-6.5pt}]
\draw[black!55,densely dashed,line width=0.55pt](axis cs:0.72,43.6)--(axis cs:85,43.6);
\draw[black!50,densely dotted,line width=0.65pt](axis cs:0.72,42.1)--(axis cs:85,42.1);
\node[adref,anchor=south west] at (axis cs:0.78,43.78){Pre-train w/ GA};
\node[adref,anchor=south west,fill=white,inner ysep=0.3pt] at (axis cs:0.9,42.1){Raw GA};
\addplot[adft] coordinates {(1,35.5)(5,37.0)(20,37.2)};
\addplot[addec] coordinates {(1,40.0)(5,41.4)(20,42.2)};
\addplot[adft,dash pattern=on 0.7pt off 0.7pt,mark=none] coordinates {(20,37.2)(60,37.2)};
\addplot[addec,dash pattern=on 0.7pt off 0.7pt,mark=none] coordinates {(20,42.2)(60,42.2)};
\addplot[adft,only marks] coordinates {(60,37.2)};
\addplot[addec,only marks] coordinates {(60,42.2)};
\node[font=\fontsize{5}{5}\selectfont,text=black!60,fill=white,inner sep=0pt]
  at (axis cs:34,33.5) {/\!/};
\node[font=\fontsize{5}{5}\selectfont,text=black!60,fill=white,inner sep=0pt]
  at (axis cs:34,45.6) {/\!/};
\end{axis}
\begin{axis}[adax,name=gc,at={(ga.north east)},anchor=north west,xshift=26pt,
  width=\dimexpr0.60\linewidth-28pt\relax,height=37mm,
  xmin=0.42,xmax=4.66,ymin=0,ymax=104,
  xtick={1,2,3,4},
  xticklabels={\strut AE\\\strut Skill,\strut AE\\\strut mIoU,\strut TE\\\strut R\textsuperscript{2},\strut TE\\\strut mIoU},
  x tick label style={align=center},
  clip=false,
  ytick={0,25,50,75,100},yticklabels={0,25,50,75,100},
  ylabel={\textbf{Old} product score, 100\% pixels $\to$},
  title={(ii) \textbf{Existing} products: AE, TE}]
\filldraw[fill=adPff,draw=adPffTxt,line width=0.35pt] (axis cs:0.67,0) rectangle (axis cs:0.94,91.4);
\path[adftbar] (axis cs:1.06,0) rectangle (axis cs:1.33,80.7);
\draw[black!75,line width=0.95pt] (axis cs:0.63,91.4)--(axis cs:1.37,91.4);
\node[adforget,anchor=south,inner sep=0.8pt,xshift=2.3pt,yshift=1pt] at (axis cs:1.195,80.7) {\textminus{}10.7\%};
\filldraw[fill=adPff,draw=adPffTxt,line width=0.35pt] (axis cs:1.67,0) rectangle (axis cs:1.94,47.7);
\path[adftbar] (axis cs:2.06,0) rectangle (axis cs:2.33,36.9);
\draw[black!75,line width=0.95pt] (axis cs:1.63,47.7)--(axis cs:2.37,47.7);
\node[adforget,anchor=south,inner sep=0.8pt,xshift=2.3pt,yshift=1pt] at (axis cs:2.195,36.9) {\textminus{}10.8\%};
\filldraw[fill=adPff,draw=adPffTxt,line width=0.35pt] (axis cs:2.67,0) rectangle (axis cs:2.94,61.1);
\path[adftbar] (axis cs:3.06,0) rectangle (axis cs:3.33,41.8);
\draw[black!75,line width=0.95pt] (axis cs:2.63,61.1)--(axis cs:3.37,61.1);
\node[adforget,anchor=south,inner sep=0.8pt,xshift=2.3pt,yshift=1pt] at (axis cs:3.195,41.8) {\textminus{}19.3\%};
\filldraw[fill=adPff,draw=adPffTxt,line width=0.35pt] (axis cs:3.67,0) rectangle (axis cs:3.94,55.3);
\path[adftbar] (axis cs:4.06,0) rectangle (axis cs:4.33,39.4);
\draw[black!75,line width=0.95pt] (axis cs:3.63,55.3)--(axis cs:4.37,55.3);
\node[adforget,anchor=south,inner sep=0.8pt,xshift=2.3pt,yshift=1pt] at (axis cs:4.195,39.4) {\textminus{}15.9\%};
\draw[black!75,line width=0.95pt] (axis cs:1.90,89)--(axis cs:2.35,89);
\node[adref,text=black!85,anchor=west,align=left] at (axis cs:2.43,89)
  {PFF performance\\[-1pt]before GA add};
\end{axis}
\end{tikzpicture}
\end{minipage}
\caption{\textbf{Continual learning results. } \textbf{(a)} Only training the small temporal slice (0.19\% params) achieves $\approx$92\% of joint pretraining's new-year reconstruction while leaving existing years unchanged (\textcolor{adPff}{\textbf{PFF-Append}}). When the product decoders are trained with this slice (\textcolor{adPffTxt}{\textbf{FT-Dec}}), or the entire PFF is
finetuned (\textcolor{adPffTxt}{\textbf{FT-All}}), existing-year retention
collapses.
\textbf{(b)} We add Galileo (GA) \citep{galileo} 10m embeddings to PFFs covering the PASTIS segmentation benchmark. Only training a new decoder on a small subset of pixels significantly outperforms finetuning the PFF on GA's PASTIS mIoU while fully retaining AlphaEarth (AE) and TESSERA's (TE) reconstruction and performance on PASTIS. Expanded in \Cref{fig:temporal-adaptation-per-tile}.}
\label{fig:temporal-append-result}
\vspace{-15pt}
\end{figure}
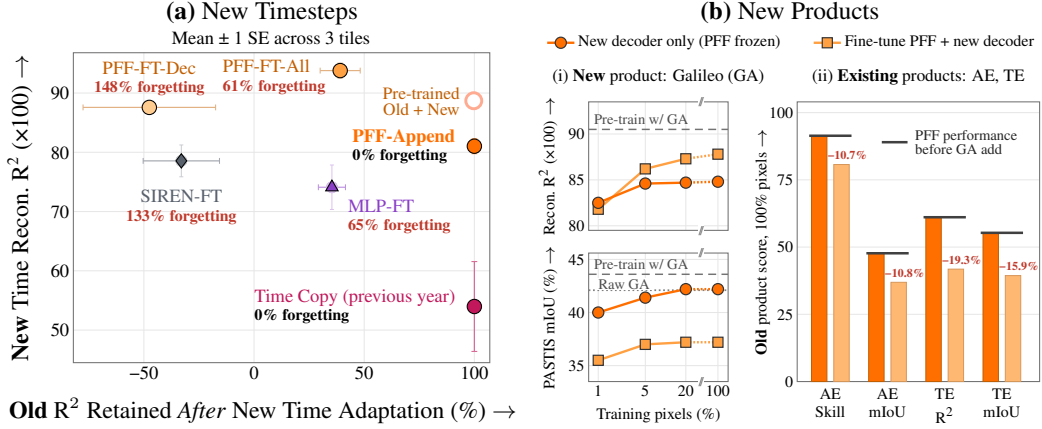

\subsection{Continually Learning PFFs}
\textbf{Learning a new timestep.}
A PFF stores time as an ordered sequence of learned parameter slices, each associated with a fixed calendar year. 
Queries interpolate each time-dependent factor between its neighboring slices before combining and decoding the features.
To learn a new timestep, we append temporal slices without changing the existing slices or their calendar positions (\Cref{fig:adaptation}).
We initialize these slices from the most recent available timestep, freeze all existing parameters, and train \emph{only} the new slices using available observations and products in the new timestep.
This construction applies to any factored feature volume that decomposes a dense grid into temporal factors.
For TensoRF(VM), we extend $v_r^t$, $M_r^{xt}$, and $M_r^{yt}$, adding 124K parameters (0.19\% of a canonical PFF).
For K-Planes, we extend $P^{xt}$ and $P^{yt}$ at every scale, adding 141K parameters (0.22\%).
Because existing parameters and interpolation weights remain unchanged, both variants preserve all outputs over previously represented years with bit-exact precision. 
Deprecating/deleting an existing temporal slice, however, does not maintain this bit-exact precision property since interpolation anchors are changed.

\textbf{Learning a new product.}
After training a PFF, the shared field contains a compact spatiotemporal representation learned from the trained products. 
To incorporate a new product, we attach a new decoder $g_{\mathrm{new}}$ \emph{per PFF} that maps the existing shared field $\mathbf{h}(\mathbf{x},t)$ to this new feature.
We freeze the shared field and every existing decoder, then train only $g_{\mathrm{new}}$ using any reconstruction loss that matches the product's properties.
In our experiments, this decoder contains $1.25$M parameters, or approximately $2\%$ of a canonical PFF, and can be trained with a subset of the new product's valid pixels.
Since no existing parameter is updated, all previously supported products remain bit-exact.


\section{Experiments}
We evaluate PFFs on reconstruction quality and downstream performance retention, using compression rate-matched single-product neural fields as baselines.

\textbf{Baselines.}
We compare PFFs with a range of representation families: single-product ReLU MLPs with Fourier features \citep{tancik2020fourier}, SIREN \citep{sitzmann2020siren}, Instant-NGP (iNGP) \citep{muller2022instant}, K-Planes \citep{fridovich2023kplanes}, and TensoRF(VM) \citep{chen2022tensorf}. 
For each baseline, we match the combined parameter storage of the 14 fields to one PFF within each tile.
We evaluate two simple allocations of the total parameter budget across the 14 products: uniform allocation and allocation proportional to each product's source size in bits. 
Bit-proportional allocation improves embedding reconstruction but reduces fidelity on lower-dimensional sensor and map products, lowering the mean score (\Cref{tab:bitweighted}). 
Source size alone is therefore a poor guide to capacity allocation, and we use uniform allocation for the main experiments.
All methods are fitted to the same regional extent: $8192\times8192$ pixels at $10\,\mathrm{m}$ resolution, following the AlphaEarth tiling manifest \citep{alphaearth}.

\textbf{Reconstruction.}
We evaluate reconstruction on our 100-tile \textsc{PFFBench} dataset, totaling approximately 6.7 billion spatial locations and $16\,\mathrm{TB}$ of source data (\Cref{sec:pffbench}). 
Each product is scored on its native grid and observed years.
Image metrics such as PSNR and SSIM, used in prior EO neural fields \citep{lianet}, do not provide a common interpretation across physical measurements and learned vectors. 
We therefore report per-channel $R^2$, averaged equally across channels, for Euclidean-valued products. 
For unit-normalized AlphaEarth vectors, we report angular \emph{skill}, which measures the reduction in cosine distortion relative to the mean-direction predictor \citep{skill}. Both scores equal one for exact reconstruction and zero for the corresponding mean prediction.

\textbf{Downstream task retention.}
We test whether PFF reconstructions preserve predictive information, emphasizing fine spatial structure and interannual change. 
For each downstream task and product, we train a task-specific linear classifier on either the original features or their reconstructions.
We report \textbf{retention} as the reconstructed-feature score divided by the source-feature score, expressed as a percentage; $100\%$ matches the source and higher values indicate improvement.
Our pixel-based tasks include PASTIS and SwissCrop25 crop-type and boundary segmentation \citep{pastis,swisscrop25}, field-instance recovery in the Brazil split of the Fields of the World dataset (FTW) \citep{ftw}, and field-extent segmentation in the South Africa split of FTW. 
We selectively probe products, excluding products where raw features significantly underperform (for example, derived map products).
We experiment with different combinations of concatenated features in \Cref{tab:pff-feature-merging}.
We evaluate temporal fidelity through annual mining-footprint segmentation and change detection on EuroMineNet \citep{eurominenet}, and crop mapping across years through SwissCrop25's leave-one-year-out protocol.
Dense evaluations use $10\,\mathrm{m}$ labels except for FTW's native label lattice (nominally $6\,\mathrm{m}$).
For patch-level classification (see \Cref{tab:classification_tables}), we pool features using the \texttt{stats} pooling method in \cite{pooling} and evaluate BigEarthNet-v2 (BEN-v2) under the GEO-Bench-2 geographic partition and So2Sat-LCZ42 (So2Sat) under geographic train, validation, and test splits \citep{benv2,so2sat,simumba2025geo}.

\providecolor{pffRowBg}{HTML}{FCF4E9}
\providecommand{\pffz}[1]{\makebox[0pt][c]{#1}}
\providecommand{\pc}{{\scriptsize\%}}
\begin{table}[t]
\caption{\textbf{PFF features retain most downstream performance and can even outperform the original features.}
We report \emph{retention \%}: the percentage recovery of the \textcolor{black!55}{Original features}' performance.
\textbf{(a)} PASTIS crop-type semantic segmentation.
\textbf{(b)} FTW field delineation in Brazil and South Africa.
\textbf{(c)} SwissCrop25 crop-type segmentation from 2019--2025.
\textbf{(d)} EuroMineNet annual change detection.
Original features are AlphaEarth (AE), TESSERA (TE), and Sentinel-2 (S2).
}
\label{tab:downstream-seg}
\vspace{2pt}
\small
\noindent
\begin{minipage}[t]{0.43\linewidth}
\vspace{0pt}\setlength{\tabcolsep}{2pt}\renewcommand{\arraystretch}{1.08}
\begin{tabularx}{\linewidth}{@{}l *{3}{>{\centering\arraybackslash}X}@{}}
\textbf{(a) PASTIS} & \multicolumn{3}{c@{}}{mIoU} \\
\cmidrule(l){2-4}
 & AE & TE & S2 \\
\midrule
\textcolor{black!55}{Original features} & \textcolor{black!55}{53.6} & \textcolor{black!55}{61.3} & \textcolor{black!55}{13.3} \\
\addlinespace[2.5pt]
TensoRF(VM) & 67\pc & 69\pc & 92\pc \\
K-Planes & 69\pc & 59\pc & 91\pc \\
iNGP & 46\pc & 46\pc & 75\pc \\
\rowcolor{pffRowBg}
\textbf{PFF-VM} & \textbf{90\pc} & \textbf{90\pc} & \textbf{136\pc} \\
\rowcolor{pffRowBg}
\textbf{PFF-K-Planes} & \underline{89\pc} & \underline{87\pc} & \underline{121\pc} \\

\end{tabularx}
\end{minipage}\hfill
\begin{minipage}[t]{0.545\linewidth}
\vspace{0pt}\setlength{\tabcolsep}{2pt}\renewcommand{\arraystretch}{1.08}
\begin{tabularx}{\linewidth}{@{}l *{3}{>{\centering\arraybackslash}X} !{\hspace{2pt}\color{black!25}\vrule\hspace{2pt}} *{3}{>{\centering\arraybackslash}X}@{}}
\textbf{(c) SwissCrop25} & \multicolumn{3}{c}{OA} & \multicolumn{3}{c@{}}{Crop-type mIoU} \\
\cmidrule(lr){2-4}\cmidrule(l){5-7}
 & AE & TE & S2 & AE & TE & S2 \\
\midrule
\textcolor{black!55}{Original features} & \textcolor{black!55}{52.4} & \textcolor{black!55}{53.8} & \textcolor{black!55}{32.9} & \textcolor{black!55}{4.9} & \textcolor{black!55}{6.1} & \textcolor{black!55}{0.9} \\
\addlinespace[2.5pt]
TensoRF(VM) & 94\pc & 96\pc & 101\pc & 73\pc & 80\pc & 102\pc \\
K-Planes & 93\pc & 91\pc & \textbf{106\pc} & 71\pc & 65\pc & \pffz{\underline{114\pc}} \\
iNGP & 89\pc & 88\pc & \underline{104\pc} & 57\pc & 53\pc & \pffz{\textbf{115\pc}} \\
\rowcolor{pffRowBg}
\textbf{PFF-VM} & \textbf{100\pc} & \underline{99\pc} & \underline{104\pc} & \textbf{104\pc} & \underline{92\pc} & \pffz{111\pc} \\
\rowcolor{pffRowBg}
\textbf{PFF-K-Planes} & \underline{99\pc} & \textbf{100\pc} & \underline{104\pc} & \underline{98\pc} & \textbf{97\pc} & 109\pc \\

\end{tabularx}
\end{minipage}
\par\vspace{9pt}
\noindent
\begin{minipage}[t]{0.43\linewidth}
\vspace{0pt}\setlength{\tabcolsep}{2pt}\renewcommand{\arraystretch}{1.08}
\begin{tabularx}{\linewidth}{@{}l *{2}{>{\centering\arraybackslash}X} !{\hspace{2pt}\color{black!25}\vrule\hspace{2pt}} *{2}{>{\centering\arraybackslash}X}@{}}
\textbf{(b) FTW} & \multicolumn{2}{c}{Brazil {\scriptsize (OR)}} & \multicolumn{2}{c@{}}{S.\ Africa {\scriptsize (IoU)}} \\
\cmidrule(lr){2-3}\cmidrule(l){4-5}
& AE & S2 & AE & S2 \\
\midrule
\textcolor{black!55}{Original features} & \textcolor{black!55}{50.0} & \textcolor{black!55}{21.2} & \textcolor{black!55}{73.5} & \textcolor{black!55}{57.1} \\
\addlinespace[2.5pt]
TensoRF(VM) & 70\pc & 69\pc & 99\pc & 101\pc \\
K-Planes & 66\pc & 64\pc & 99\pc & 96\pc \\
iNGP & 22\pc & 46\pc & 93\pc & 104\pc \\
\rowcolor{pffRowBg}
\textbf{PFF-VM} & \underline{90\pc} & \underline{90\pc} & \textbf{103\pc} & \underline{105\pc} \\
\rowcolor{pffRowBg}
\textbf{PFF-K-Planes} & \textbf{95\pc} & \textbf{103\pc} & \underline{102\pc} & \textbf{121\pc} \\

\end{tabularx}
\end{minipage}\hfill
\begin{minipage}[t]{0.545\linewidth}
\vspace{0pt}\setlength{\tabcolsep}{2pt}\renewcommand{\arraystretch}{1.08}
\begin{tabularx}{\linewidth}{@{}l *{2}{>{\centering\arraybackslash}X} !{\hspace{2pt}\color{black!25}\vrule\hspace{2pt}} *{2}{>{\centering\arraybackslash}X}@{}}
\textbf{(d) EuroMineNet} 
 & AUPRC & F1@r & AUPRC & F1@r \\
 \cmidrule(lr){2-3}\cmidrule(l){4-5} & 
 \multicolumn{2}{c}{AE} & \multicolumn{2}{c@{}}{TE} \\
\midrule
\textcolor{black!55}{Original features} & \textcolor{black!55}{10.7} & \textcolor{black!55}{17.4} & \textcolor{black!55}{10.8} & \textcolor{black!55}{21.0} \\
\addlinespace[2.5pt]
TensoRF(VM) & 72\pc & 79\pc & \underline{88\pc} & \underline{90\pc} \\
K-Planes & \underline{90\pc} & \textbf{99\pc} & 43\pc & 41\pc \\
iNGP & 37\pc & 49\pc & 32\pc & 37\pc \\
\rowcolor{pffRowBg}
\textbf{PFF-VM} & \textbf{100\pc} & \underline{97\pc} & \textbf{103\pc} & \textbf{97\pc} \\
\rowcolor{pffRowBg}
\textbf{PFF-K-Planes} & 76\pc & 86\pc & 82\pc & 80\pc \\
\end{tabularx}
\end{minipage}
\end{table}

\section{Results}
\textbf{Sharing across products improves reconstruction at high compression ratios.} 
PFFs with both VM and K-Planes-based factors improve reconstruction when shared across products, reducing mean normalized distortion by approximately 20$\%$ at matched storage (\Cref{fig:rd-fairshare-ladder}). 
This advantage persists across multiple parameter budgets, but does not extend to every shared representation. 
For example, in \Cref{fig:rd-fairshare-ladder}, we replace the factored volume with either a multi-resolution hash-grid of \cite{muller2022instant}, or a ReLU/SIREN coordinate network replicating STRAINER's setup for transferable implicit neural representations \citep{strainer}.
Each produces a shared feature vector for product-specific decoders. 
These controls reconstruct less accurately when shared than when fitted independently to each product. 
The hash-grid result is particularly informative because it retains the explicit--implicit architecture in \Cref{fig:arch} but removes the factorization. 
Together, these comparisons show that factored feature volumes support effective joint encoding of heterogeneous products; sharing an encoder and attaching separate decoders is not sufficient to obtain the same benefit. We visualize additional reconstructions in \Cref{fig:appendix-recon-curated-1,fig:appendix-recon-curated-2}.

\begin{wrapfigure}{r}{0.40\linewidth}
\vspace{-8pt}
\centering
\resizebox{\linewidth}{!}{%
\begingroup%
\definecolor{latRead}{HTML}{35708E}%
\definecolor{latLoad}{HTML}{C6A76F}%
\definecolor{latForward}{HTML}{6B5AA8}%
\definecolor{latOther}{HTML}{A7ADB5}%
\definecolor{latFrame}{HTML}{BFC3CA}%
\tikzset{%
 lat/text/.style={font=\fontsize{6.8}{8}\selectfont\rmfamily,text=black,inner sep=0pt},%
 lat/small/.style={font=\fontsize{6.2}{7.2}\selectfont\rmfamily,text=black,inner sep=0pt},%
 lat/bold/.style={font=\fontsize{6.8}{8}\selectfont\rmfamily\bfseries,text=black,inner sep=0pt},%
 lat/product/.style={font=\fontsize{7.5}{8.5}\selectfont\rmfamily\bfseries,text=black,inner sep=0pt},%
 lat/grid/.style={draw=latFrame!55,line width=.25pt},%
}%
\begin{tikzpicture}[x=1pt,y=-1pt,every node/.style={outer sep=0pt}]
\path[use as bounding box] (0,0) rectangle (159,111);
\node[lat/bold,anchor=west] at (0.0000,5.0000) {Seconds};
\node[lat/small,anchor=center] at (45.0000,5.0000) {0};
\draw[lat/grid] (45.0000,12)--(45.0000,58);
\draw[lat/grid] (45.0000,61)--(45.0000,98);
\node[lat/small,anchor=center] at (58.5554,5.0000) {100};
\draw[lat/grid] (58.5554,12)--(58.5554,58);
\draw[lat/grid] (58.5554,61)--(58.5554,98);
\node[lat/small,anchor=center] at (87.8661,5.0000) {1{,}000};
\draw[lat/grid] (87.8661,12)--(87.8661,58);
\draw[lat/grid] (87.8661,61)--(87.8661,98);
\node[lat/small,anchor=center] at (119.2462,5.0000) {3{,}000};
\draw[lat/grid] (119.2462,12)--(119.2462,58);
\draw[lat/grid] (119.2462,61)--(119.2462,98);
\node[lat/small,anchor=center] at (150.0000,5.0000) {6{,}000};
\draw[lat/grid] (150.0000,12)--(150.0000,58);
\draw[lat/grid] (150.0000,61)--(150.0000,98);
\node[lat/product,rotate=90,anchor=center] at (3.3000,36.0000) {AE};
\node[lat/text,anchor=east] at (41.0000,18.0000) {EE API};
\begin{scope}\clip[rounded corners=.5pt] (45,15.1) rectangle (96.131912,20.9);
\fill[latRead] (45.0000000,15.1) rectangle (96.0788981,20.9);
\fill[latLoad] (96.0788981,15.1) rectangle (96.0813710,20.9);
\fill[latOther] (96.0813710,15.1) rectangle (96.0854934,20.9);
\fill[latOther] (96.0854934,15.1) rectangle (96.1319118,20.9);
\end{scope}
\draw[rounded corners=.5pt,draw=latFrame!80,line width=.3pt] (45,15.1) rectangle (96.131912,20.9);
\node[lat/small,text=white,anchor=east] at (94.1319,18.0000) {1{,}423};
\node[lat/text,anchor=east] at (41.0000,27.0000) {COG Range};
\begin{scope}\clip[rounded corners=.5pt] (45,24.1) rectangle (131.770957,29.9);
\fill[latRead] (45.0000000,24.1) rectangle (53.6045741,29.9);
\begin{scope}\clip (53.6045741,24.1) rectangle (131.6990723,29.9);
\fill[latLoad] (53.6045741,24.1) rectangle (131.6990723,29.9);
\draw[latRead,line width=.65pt] (47.6000,29.9)--(53.4000,24.1);
\draw[latRead,line width=.65pt] (50.4000,29.9)--(56.2000,24.1);
\draw[latRead,line width=.65pt] (53.2000,29.9)--(59.0000,24.1);
\draw[latRead,line width=.65pt] (56.0000,29.9)--(61.8000,24.1);
\draw[latRead,line width=.65pt] (58.8000,29.9)--(64.6000,24.1);
\draw[latRead,line width=.65pt] (61.6000,29.9)--(67.4000,24.1);
\draw[latRead,line width=.65pt] (64.4000,29.9)--(70.2000,24.1);
\draw[latRead,line width=.65pt] (67.2000,29.9)--(73.0000,24.1);
\draw[latRead,line width=.65pt] (70.0000,29.9)--(75.8000,24.1);
\draw[latRead,line width=.65pt] (72.8000,29.9)--(78.6000,24.1);
\draw[latRead,line width=.65pt] (75.6000,29.9)--(81.4000,24.1);
\draw[latRead,line width=.65pt] (78.4000,29.9)--(84.2000,24.1);
\draw[latRead,line width=.65pt] (81.2000,29.9)--(87.0000,24.1);
\draw[latRead,line width=.65pt] (84.0000,29.9)--(89.8000,24.1);
\draw[latRead,line width=.65pt] (86.8000,29.9)--(92.6000,24.1);
\draw[latRead,line width=.65pt] (89.6000,29.9)--(95.4000,24.1);
\draw[latRead,line width=.65pt] (92.4000,29.9)--(98.2000,24.1);
\draw[latRead,line width=.65pt] (95.2000,29.9)--(101.0000,24.1);
\draw[latRead,line width=.65pt] (98.0000,29.9)--(103.8000,24.1);
\draw[latRead,line width=.65pt] (100.8000,29.9)--(106.6000,24.1);
\draw[latRead,line width=.65pt] (103.6000,29.9)--(109.4000,24.1);
\draw[latRead,line width=.65pt] (106.4000,29.9)--(112.2000,24.1);
\draw[latRead,line width=.65pt] (109.2000,29.9)--(115.0000,24.1);
\draw[latRead,line width=.65pt] (112.0000,29.9)--(117.8000,24.1);
\draw[latRead,line width=.65pt] (114.8000,29.9)--(120.6000,24.1);
\draw[latRead,line width=.65pt] (117.6000,29.9)--(123.4000,24.1);
\draw[latRead,line width=.65pt] (120.4000,29.9)--(126.2000,24.1);
\draw[latRead,line width=.65pt] (123.2000,29.9)--(129.0000,24.1);
\draw[latRead,line width=.65pt] (126.0000,29.9)--(131.8000,24.1);
\draw[latRead,line width=.65pt] (128.8000,29.9)--(134.6000,24.1);
\draw[latRead,line width=.65pt] (131.6000,29.9)--(137.4000,24.1);
\draw[latRead,line width=.65pt] (134.4000,29.9)--(140.2000,24.1);
\end{scope}
\fill[latLoad] (131.6990723,24.1) rectangle (131.7066334,29.9);
\fill[latOther] (131.7066334,24.1) rectangle (131.7090488,29.9);
\fill[latOther] (131.7090488,24.1) rectangle (131.7709573,29.9);
\end{scope}
\draw[rounded corners=.5pt,draw=latFrame!80,line width=.3pt] (45,24.1) rectangle (131.770957,29.9);
\begin{scope}\clip[rounded corners=.5pt] (45,24.1) rectangle (131.7709573,29.9);
\fill[white,opacity=.085,rounded corners=.65pt] (113.9709573,23.9000000) rectangle (130.6709573,30.1000000);
\fill[white,opacity=.085,rounded corners=.65pt] (114.0436846,23.9727273) rectangle (130.5982300,30.0272727);
\fill[white,opacity=.085,rounded corners=.65pt] (114.1164118,24.0454545) rectangle (130.5255028,29.9545455);
\fill[white,opacity=.085,rounded corners=.65pt] (114.1891391,24.1181818) rectangle (130.4527755,29.8818182);
\fill[white,opacity=.085,rounded corners=.65pt] (114.2618664,24.1909091) rectangle (130.3800482,29.8090909);
\fill[white,opacity=.085,rounded corners=.65pt] (114.3345937,24.2636364) rectangle (130.3073209,29.7363636);
\fill[white,opacity=.085,rounded corners=.65pt] (114.4073209,24.3363636) rectangle (130.2345937,29.6636364);
\fill[white,opacity=.085,rounded corners=.65pt] (114.4800482,24.4090909) rectangle (130.1618664,29.5909091);
\fill[white,opacity=.085,rounded corners=.65pt] (114.5527755,24.4818182) rectangle (130.0891391,29.5181818);
\fill[white,opacity=.085,rounded corners=.65pt] (114.6255028,24.5545455) rectangle (130.0164118,29.4454545);
\fill[white,opacity=.085,rounded corners=.65pt] (114.6982300,24.6272727) rectangle (129.9436846,29.3727273);
\fill[white,opacity=.085,rounded corners=.65pt] (114.7709573,24.7000000) rectangle (129.8709573,29.3000000);
\end{scope}
\node[lat/small,text=black,anchor=east] at (129.7710,27.0000) {4{,}098};
\node[lat/text,anchor=east] at (41.0000,36.0000) {COG Full};
\begin{scope}\clip[rounded corners=.5pt] (45,33.1) rectangle (148.015878,38.9);
\fill[latRead] (45.0000000,33.1) rectangle (79.8213665,38.9);
\begin{scope}\clip (79.8213665,33.1) rectangle (147.7320761,38.9);
\fill[latLoad] (79.8213665,33.1) rectangle (147.7320761,38.9);
\draw[latRead,line width=.65pt] (72.8000,38.9)--(78.6000,33.1);
\draw[latRead,line width=.65pt] (75.6000,38.9)--(81.4000,33.1);
\draw[latRead,line width=.65pt] (78.4000,38.9)--(84.2000,33.1);
\draw[latRead,line width=.65pt] (81.2000,38.9)--(87.0000,33.1);
\draw[latRead,line width=.65pt] (84.0000,38.9)--(89.8000,33.1);
\draw[latRead,line width=.65pt] (86.8000,38.9)--(92.6000,33.1);
\draw[latRead,line width=.65pt] (89.6000,38.9)--(95.4000,33.1);
\draw[latRead,line width=.65pt] (92.4000,38.9)--(98.2000,33.1);
\draw[latRead,line width=.65pt] (95.2000,38.9)--(101.0000,33.1);
\draw[latRead,line width=.65pt] (98.0000,38.9)--(103.8000,33.1);
\draw[latRead,line width=.65pt] (100.8000,38.9)--(106.6000,33.1);
\draw[latRead,line width=.65pt] (103.6000,38.9)--(109.4000,33.1);
\draw[latRead,line width=.65pt] (106.4000,38.9)--(112.2000,33.1);
\draw[latRead,line width=.65pt] (109.2000,38.9)--(115.0000,33.1);
\draw[latRead,line width=.65pt] (112.0000,38.9)--(117.8000,33.1);
\draw[latRead,line width=.65pt] (114.8000,38.9)--(120.6000,33.1);
\draw[latRead,line width=.65pt] (117.6000,38.9)--(123.4000,33.1);
\draw[latRead,line width=.65pt] (120.4000,38.9)--(126.2000,33.1);
\draw[latRead,line width=.65pt] (123.2000,38.9)--(129.0000,33.1);
\draw[latRead,line width=.65pt] (126.0000,38.9)--(131.8000,33.1);
\draw[latRead,line width=.65pt] (128.8000,38.9)--(134.6000,33.1);
\draw[latRead,line width=.65pt] (131.6000,38.9)--(137.4000,33.1);
\draw[latRead,line width=.65pt] (134.4000,38.9)--(140.2000,33.1);
\draw[latRead,line width=.65pt] (137.2000,38.9)--(143.0000,33.1);
\draw[latRead,line width=.65pt] (140.0000,38.9)--(145.8000,33.1);
\draw[latRead,line width=.65pt] (142.8000,38.9)--(148.6000,33.1);
\draw[latRead,line width=.65pt] (145.6000,38.9)--(151.4000,33.1);
\draw[latRead,line width=.65pt] (148.4000,38.9)--(154.2000,33.1);
\end{scope}
\fill[latLoad] (147.7320761,33.1) rectangle (147.9617991,38.9);
\fill[latOther] (147.9617991,33.1) rectangle (147.9638478,38.9);
\fill[latOther] (147.9638478,33.1) rectangle (148.0158778,38.9);
\end{scope}
\draw[rounded corners=.5pt,draw=latFrame!80,line width=.3pt] (45,33.1) rectangle (148.015878,38.9);
\begin{scope}\clip[rounded corners=.5pt] (45,33.1) rectangle (148.0158778,38.9);
\fill[white,opacity=.085,rounded corners=.65pt] (130.2158778,32.9000000) rectangle (146.9158778,39.1000000);
\fill[white,opacity=.085,rounded corners=.65pt] (130.2886051,32.9727273) rectangle (146.8431505,39.0272727);
\fill[white,opacity=.085,rounded corners=.65pt] (130.3613323,33.0454545) rectangle (146.7704233,38.9545455);
\fill[white,opacity=.085,rounded corners=.65pt] (130.4340596,33.1181818) rectangle (146.6976960,38.8818182);
\fill[white,opacity=.085,rounded corners=.65pt] (130.5067869,33.1909091) rectangle (146.6249687,38.8090909);
\fill[white,opacity=.085,rounded corners=.65pt] (130.5795142,33.2636364) rectangle (146.5522414,38.7363636);
\fill[white,opacity=.085,rounded corners=.65pt] (130.6522414,33.3363636) rectangle (146.4795142,38.6636364);
\fill[white,opacity=.085,rounded corners=.65pt] (130.7249687,33.4090909) rectangle (146.4067869,38.5909091);
\fill[white,opacity=.085,rounded corners=.65pt] (130.7976960,33.4818182) rectangle (146.3340596,38.5181818);
\fill[white,opacity=.085,rounded corners=.65pt] (130.8704233,33.5545455) rectangle (146.2613323,38.4454545);
\fill[white,opacity=.085,rounded corners=.65pt] (130.9431505,33.6272727) rectangle (146.1886051,38.3727273);
\fill[white,opacity=.085,rounded corners=.65pt] (131.0158778,33.7000000) rectangle (146.1158778,38.3000000);
\end{scope}
\node[lat/small,text=black,anchor=east] at (146.0159,36.0000) {5{,}775};
\node[lat/text,anchor=east] at (41.0000,45.0000) {PFF\textsubscript{CPU}};
\begin{scope}\clip[rounded corners=.5pt] (45,42.1) rectangle (58.946179,47.9);
\fill[latRead] (45.0000000,42.1) rectangle (51.7534657,47.9);
\fill[latLoad] (51.7534657,42.1) rectangle (56.2483183,47.9);
\fill[latForward] (56.2483183,42.1) rectangle (58.7463081,47.9);
\fill[latOther] (58.7463081,42.1) rectangle (58.7616775,47.9);
\fill[latOther] (58.7616775,42.1) rectangle (58.9461794,47.9);
\end{scope}
\draw[rounded corners=.5pt,draw=latFrame!80,line width=.3pt] (45,42.1) rectangle (58.946179,47.9);
\node[lat/text,anchor=west] at (61.9462,45.0000) {106};
\node[lat/bold,anchor=east] at (41.0000,54.0000) {PFF\textsubscript{GPU}};
\begin{scope}\clip[rounded corners=.5pt] (45,51.1) rectangle (57.088616,56.9);
\fill[latRead] (45.0000000,51.1) rectangle (51.7416944,56.9);
\fill[latLoad] (51.7416944,51.1) rectangle (56.7283428,56.9);
\fill[latForward] (56.7283428,51.1) rectangle (56.8425841,56.9);
\fill[latOther] (56.8425841,51.1) rectangle (56.8434672,56.9);
\fill[latOther] (56.8434672,51.1) rectangle (57.0886155,56.9);
\end{scope}
\draw[rounded corners=.5pt,draw=latFrame!80,line width=.3pt] (45,51.1) rectangle (57.088616,56.9);
\node[lat/bold,anchor=west] at (60.0886,54.0000) {79.5};
\node[lat/product,rotate=90,anchor=center] at (3.3000,80.5000) {TE};
\node[lat/text,anchor=east] at (41.0000,67.0000) {Tiles (Full)};
\begin{scope}\clip[rounded corners=.5pt] (45,64.1) rectangle (95.337473,69.9);
\fill[latRead] (45.0000000,64.1) rectangle (91.6691544,69.9);
\begin{scope}\clip (91.6691544,64.1) rectangle (95.1954796,69.9);
\fill[latLoad] (91.6691544,64.1) rectangle (95.1954796,69.9);
\draw[latRead,line width=.65pt] (84.0000,69.9)--(89.8000,64.1);
\draw[latRead,line width=.65pt] (86.8000,69.9)--(92.6000,64.1);
\draw[latRead,line width=.65pt] (89.6000,69.9)--(95.4000,64.1);
\draw[latRead,line width=.65pt] (92.4000,69.9)--(98.2000,64.1);
\draw[latRead,line width=.65pt] (95.2000,69.9)--(101.0000,64.1);
\end{scope}
\fill[latLoad] (95.1954796,64.1) rectangle (95.1956591,69.9);
\fill[latOther] (95.1956591,64.1) rectangle (95.2039866,69.9);
\fill[latOther] (95.2039866,64.1) rectangle (95.3374726,69.9);
\end{scope}
\draw[rounded corners=.5pt,draw=latFrame!80,line width=.3pt] (45,64.1) rectangle (95.337473,69.9);
\node[lat/small,anchor=west] at (98.3375,67.0000) {1{,}379};
\node[lat/text,anchor=east] at (41.0000,76.0000) {Zarr Read};
\begin{scope}\clip[rounded corners=.5pt] (45,73.1) rectangle (122.527571,78.9);
\fill[latRead] (45.0000000,73.1) rectangle (122.2398156,78.9);
\fill[latOther] (122.2398156,73.1) rectangle (122.2451755,78.9);
\fill[latOther] (122.2451755,73.1) rectangle (122.5275713,78.9);
\end{scope}
\draw[rounded corners=.5pt,draw=latFrame!80,line width=.3pt] (45,73.1) rectangle (122.527571,78.9);
\node[lat/small,text=white,anchor=east] at (120.5276,76.0000) {3{,}271};
\node[lat/text,anchor=east] at (41.0000,85.0000) {PFF\textsubscript{CPU}};
\begin{scope}\clip[rounded corners=.5pt] (45,82.1) rectangle (58.776765,87.9);
\fill[latRead] (45.0000000,82.1) rectangle (51.4839957,87.9);
\fill[latLoad] (51.4839957,82.1) rectangle (56.0285841,87.9);
\fill[latForward] (56.0285841,82.1) rectangle (58.5546014,87.9);
\fill[latOther] (58.5546014,82.1) rectangle (58.5854118,87.9);
\fill[latOther] (58.5854118,82.1) rectangle (58.7767646,87.9);
\end{scope}
\draw[rounded corners=.5pt,draw=latFrame!80,line width=.3pt] (45,82.1) rectangle (58.776765,87.9);
\node[lat/text,anchor=west] at (61.7768,85.0000) {103};
\node[lat/bold,anchor=east] at (41.0000,94.0000) {PFF\textsubscript{GPU}};
\begin{scope}\clip[rounded corners=.5pt] (45,91.1) rectangle (57.774691,96.9);
\fill[latRead] (45.0000000,91.1) rectangle (53.0140732,96.9);
\fill[latLoad] (53.0140732,91.1) rectangle (57.4811096,96.9);
\fill[latForward] (57.4811096,91.1) rectangle (57.5852459,96.9);
\fill[latOther] (57.5852459,91.1) rectangle (57.5860940,96.9);
\fill[latOther] (57.5860940,91.1) rectangle (57.7746909,96.9);
\end{scope}
\draw[rounded corners=.5pt,draw=latFrame!80,line width=.3pt] (45,91.1) rectangle (57.774691,96.9);
\node[lat/bold,anchor=west] at (60.7747,94.0000) {88.8};
\node[lat/small,anchor=west] at (0.0000,107.0000) {\hbox to 159pt{\raisebox{-.6pt}{\tikz[x=1pt,y=1pt]\fill[latRead] (0,0) rectangle (4.4,4.4);}\hspace{2.6pt}Read / Download\hfil \raisebox{-.6pt}{\tikz[x=1pt,y=1pt]\fill[latLoad] (0,0) rectangle (4.4,4.4);}\hspace{2.6pt}Load data / models\hfil \raisebox{-.6pt}{\tikz[x=1pt,y=1pt]\fill[latForward] (0,0) rectangle (4.4,4.4);}\hspace{2.6pt}Forward Prop}};
\end{tikzpicture}%
\endgroup%
}
\vspace{-10pt}
\caption{\raggedright \textbf{PFFs materialize features an order of magnitude faster than API or cloud-based methods.} Time (s) to materialize 1 million AE and TE embeddings visualized.
}
\vspace{-10pt}
\label{fig:latency}
\end{wrapfigure}
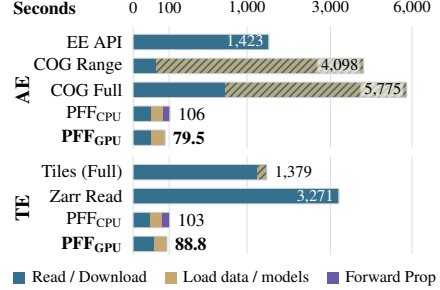 \paragraph{PFFs can continually learn unseen timesteps and products without forgetting.} 
We evaluate \emph{temporal adaptation} on a seasonally varying \textsc{PFFBench} tile and regions affected by the Pacific Palisades fire and Sindh floods (\Cref{fig:temporal-append-result}a). 
We add all products in 2025 to PFF-VMs trained on 2017--2024 for the first two regions, and 2022 after training on 2017--2021 for Sindh. 
Across the three regions, PFF-Append reaches approximately 92\% of the joint-pretraining reconstruction score in 2\% of the training time, while preserving earlier outputs bit-exactly.
In contrast, finetuning the PFF or the product decoders in addition to the added temporal slice achieves higher new-timestep scores but exhibits catastrophic forgetting of existing timesteps \citep{van2024continual}. 
PFF-K-Planes also learns the new year while preserving earlier outputs (\Cref{fig:temporal-adaptation-per-tile}). See \Cref{fig:temporal-adaptation-qualitative} for a qualitative result.

For \emph{new-product adaptation}, we add Galileo embeddings \citep{galileo} covering 17 PASTIS tiles (\Cref{fig:temporal-append-result}b).
Here, training only the new implicit decoders on 20\% of the 2019 target pixels matches source performance ($42.2\%$ versus $42.1\%$ mIoU) while preserving existing outputs. 
Conversely, finetuning the shared explicit field improves Galileo reconstruction but lowers its mIoU to $37.2\%$ and reduces AlphaEarth and TESSERA relative mIoU by 10.8 and 15.9 percentage points.
This divergence suggests that optimizing Galileo reconstruction alone can sacrifice task-relevant structure in the jointly learned field. 
Freezing the field preserves a representation shaped by 14 complementary products, which may benefit the new decoder's downstream predictions.
Galileo-only finetuning weakens this cross-product structure.

\textbf{PFFs retain the original datasets' downstream task performance at high compression ratios. } 
We focus our evaluation on dense prediction tasks, where high spatial and temporal fidelity of the input is crucial.
On PASTIS crop segmentation and FTW Brazil field-instance recall, both PFF variants improve retention over the strongest single-product baseline by 18--44 and 20--34 percentage points, respectively (\Cref{tab:downstream-seg}). 
On EuroMineNet annual change detection, PFF-VM retains 97--103\% of the original AE and TE performance across AUPRC and F1 at the true change rate.
PFF-K-Planes retains 76--86\%, showing that the two factorizations differ in how well they preserve features used to detect temporal change. 
For mining-footprint mapping, both PFF variants outperform the single-product baselines on boundary F1 across all five products, despite similar overall pixel-level F1 (\Cref{tab:eurominenet_footprint_mapping}).
Both variants outperform the original Sentinel-2 features on PASTIS and FTW South Africa. 
These gains may reflect complementary supervision and mild denoising, although our experiments do not isolate their contributions.
For completeness, we also present patch-classification results on common benchmarks (\Cref{tab:transfer}). 
Local fields largely match source performance on BEN-v2, and PFF improves only modestly over a single-product TensoRF on So2Sat. 
This is expected behavior, as overly smooth reconstructions may retain coarse features relevant for patch-level interpretation.

\paragraph{PFFs materialize precomputed embeddings faster than standard data-access pipelines (\Cref{fig:latency}).}  We request one million AE or TE embeddings from 100 globally sampled regions (\Cref{fig:fleet}). All methods run on the same eight-vCPU virtual machine with an NVIDIA L4 GPU, start with empty local caches, and return embeddings to system memory. Including checkpoint download and model loading, PFFs are 17.9$\times$ faster than Earth Engine for AE and 15.5$\times$ faster than full-tile downloads for TE. Most initial latency comes from downloading checkpoints and constructing inference sessions, rather than running the forward pass. We find PFFs to be especially fast as a data loader or a large-scale mapper. PFFs achieve an embedding throughput of 9.7 million AE embeddings per second on an NVIDIA H100 GPU (see \Cref{fig:e3-throughput}).

\section{Related work}

\textbf{Neural fields} encode signals as continuous, coordinate-based functions  \citep{mildenhall2020nerf,chen2019learning}, with applications in view synthesis \citep{barron2021mipnerf}, video and signal compression \citep{chen2021nerv,dupont2021coin,dupont2022coin,inrcompress}, medical image-to-volume reconstruction \citep{xu2022nesvor}, shape modeling \citep{park2019deepsdf}, and multi-image super-resolution \citep{jyhne2026superf}. 
As these representations have scaled, much of the literature has moved toward local or explicit--implicit parameterizations with voxel or feature grids \citep{peng2020convolutional,liu2020neural,yu2021plenoxels,grids}, octrees \citep{yu2021plenoctrees}, hash encodings \citep{muller2022instant}, and tensor or planar factorizations \citep{chen2022tensorf,PuTT,fridovich2023kplanes,tilted,cao2023hexplane}.
PFFs are closest in spirit to \textit{feature-field methods}, which distill pretrained image features into queryable 3D representations, including DFF \citep{kobayashi2022dff}, N3F \citep{tschernezki2022n3f}, LERF \citep{kerr2023lerf}, F3RM \citep{shen2023f3rm}, and FeatureNeRF \citep{ye2023featurenerf}.
Other methods divide scenes into smaller neural fields to reconstruct large areas or speed up rendering, including Block-NeRF \citep{tancik2022blocknerf}, KiloNeRF \citep{reiser2021kilonerf}, Mega-NeRF \citep{turki2022meganerf}, and Switch-NeRF \citep{mi2023switchnerf}.  Functa and spatial functa motivate treating datasets as collections of neural fields \citep{functa22,spatialfuncta23}. Generalizable NeRFs predict fields from new input images \citep{pixelnerf,mvsnerf}. PFFs place observations, embeddings, and map products in the same regional factors, and test whether this joint representation improves reconstruction over separate fields at equal compression rates. Continual methods such as CLNeRF or CD-NGP update existing parameters using replay and parameter isolation, respectively \citep{cai2023clnerf,liu2024cd}. PFFs instead train added temporal slices or product decoders while freezing existing parameters, preserving earlier outputs bit-exactly.

\textbf{Geographic location encoders} learn continuous mappings from location to task values or reusable embeddings \citep{wrap,mai2022review,cole2023spatial}. 
Contrastive and retrieval-based training produce general-purpose geographic features \citep{satclip,geoclip,csp,dhakal2025range}, while recent methods use distillation \citep{climplicit,lane2026sled,mind}. 
These models learn global representations for prediction or transfer. 
Unlike location encoders, PFFs use spatially local fields as compact \textit{substitutes} for dense, space--time-indexed EO products. 
Global retrieval remains simple: each query is routed to the regional field that contains its location.
TerraCodec \citep{terracodec} instead targets data storage, compressing multispectral images and time series with pretrained transforms and entropy models. 
Pretrained neural codecs must generalize beyond the geographic distribution used for training. 
Direct fitting instead optimizes the representation on the data being stored, as in climate grids \citep{huang2023compressing}, hyperspectral imagery \citep{hiner}, and regional Sentinel-2 time series \citep{lianet}.
PFFs store multiple products in a shared explicit representation, so their aggregate compression ratio is not directly comparable to a codec for a single product alone.
Storage costs also differ: PFFs encode regional data in learned parameters, while TerraCodec produces compressed files that require a separate, reusable model to decode. 
PFFs also allow reuse of their learned regional structure as new observations arrive, without refitting historical data and while keeping previous reconstructions unchanged.

\section{Limitations and Conclusion}
\textbf{Limitations.} 
Directly comparing our compression ratios with prior work is non-trivial, as our ratios can easily be inflated by adding redundant embeddings \citep{geoinrid}. 
For example, adding Galileo increases our compression ratio from $1800\times$ to $2800\times$ with the shared field unchanged (\Cref{fig:temporal-append-result}b), while missing products may reduce the ratio similarly.
Our comparisons account for product composition, coverage, numerical precision, and reconstruction quality. 
Each PFF is fitted at a fixed storage budget and does not support variable-rate compression. 
PFF reconstructions are \emph{lossy}, and their effect on downstream performance depends on the task. 
Smoothing may leave patch-level predictions largely unchanged but remove fine structures needed for high-resolution mapping. 
The same smoothing may also reduce noise and contribute to the gains we sometimes observe over the original features, although our experiments do not separate this effect from cross-product supervision.

\textbf{Conclusion.}
We present a spatially local hybrid neural field capable of compressing 14 diverse EO products by sharing a common factored feature volume. 
PFF-materialized features better reconstruct the source features than compression rate-matched single-product neural fields, and significantly outperform other local fields on several downstream tasks. 
By using factored feature volumes, we demonstrate easy continual learning of our method to new timesteps and new products. 
We show that PFFs are capable of materializing features an order of magnitude faster than currently used cloud or API-based methods, and enable fast large-scale mapping efforts at low cost.




\section*{Acknowledgments}

We thank Lucia Gordon and Vivian White for reviewing and providing feedback.
SL is supported by the Danish Data Science Academy, which is funded by the Novo Nordisk Foundation (NNF21SA0069429) and VILLUM FONDEN (40516). 
AF is primarily supported by an NSERC PGS-D scholarship. 
We acknowledge Danish e-Infrastructure Cooperation (DeiC) and the University of Copenhagen (Denmark) for awarding this project access to the LUMI supercomputer, owned by the EuroHPC Joint Undertaking, hosted by CSC (Finland) and the LUMI consortium through the University of Copenhagen’s local allocation of DeiC National HPC resources. 
NL is supported by the
Global Wetland Center (grant number NNF23OC0081089) from Novo Nordisk Foundation.
ES is supported by a Canada CIFAR AI Chair and the Natural Sciences and Engineering Research Council of Canada (NSERC) Discovery Grant (RGPIN-2025-06878). 
Resources used in preparing this research were also provided, in part, by the Province of Ontario, the Government of Canada through CIFAR, and companies sponsoring the Vector Institute.
This research used Killarney, Vulcan, and Fir compute clusters, with support from the Digital Research Alliance of Canada (alliancecan.ca), Compute Ontario (computeontario.ca), the BC DRI Group, Prairies DRI, the Vector Institute, and the Alberta Machine Intelligence Institute (Amii). 
This work was supported in part by the Pioneer Centre for AI, DNRF grant number P1. 
This research used the TGX RAILs advanced compute and data resource, which is supported by the National Science Foundation (award OAC-2232860) and the Taylor Geospatial Institute.

\bibliography{iclr2026_conference}

@inproceedings{sitzmann2020siren,
  author        = {Vincent Sitzmann and Julien N. P. Martel and Alexander W. Bergman and David B. Lindell and Gordon Wetzstein},
  title         = {Implicit Neural Representations with Periodic Activation Functions},
  booktitle     = {Advances in Neural Information Processing Systems (NeurIPS)},
  year          = {2020},
}

@inproceedings{tancik2020fourier,
  author        = {Matthew Tancik and Pratul P. Srinivasan and Ben Mildenhall and Sara Fridovich-Keil and Nithin Raghavan and Utkarsh Singhal and Ravi Ramamoorthi and Jonathan T. Barron and Ren Ng},
  title         = {{Fourier} Features Let Networks Learn High Frequency Functions in Low Dimensional Domains},
  booktitle     = {Advances in Neural Information Processing Systems (NeurIPS)},
  year          = {2020},
}

@inproceedings{PuTT,
  author        = {Loeschcke, Sebastian Bugge and Wang, Dan and Leth-Espensen, Christian Munklinde and Belongie, Serge and Kastoryano, Michael and Benaim, Sagie},
  title         = {Coarse-To-Fine Tensor Trains for Compact Visual Representations},
  booktitle     = {International Conference on Machine Learning (ICML)},
  volume        = {235},
  pages         = {32612--32642},
  series        = {Proceedings of Machine Learning Research},
  year          = {2024},
  url           = {https://proceedings.mlr.press/v235/loeschcke24a.html},
}

@inproceedings{cai2023clnerf,
  author        = {Cai, Zhipeng and M{\"u}ller, Matthias},
  title         = {{CLNeRF}: Continual Learning Meets {NeRF}},
  booktitle     = {IEEE/CVF International Conference on Computer Vision (ICCV)},
  pages         = {23128--23137},
  year          = {2023},
}

@inproceedings{mildenhall2020nerf,
  author        = {Ben Mildenhall and Pratul P. Srinivasan and Matthew Tancik and Jonathan T. Barron and Ravi Ramamoorthi and Ren Ng},
  title         = {{NeRF}: Representing Scenes as Neural Radiance Fields for View Synthesis},
  booktitle     = {European Conference on Computer Vision (ECCV)},
  year          = {2020},
}

@inproceedings{reiser2021kilonerf,
  author        = {Christian Reiser and Songyou Peng and Yiyi Liao and Andreas Geiger},
  title         = {{KiloNeRF}: Speeding up Neural Radiance Fields with Thousands of Tiny {MLPs}},
  booktitle     = {IEEE/CVF International Conference on Computer Vision (ICCV)},
  year          = {2021},
}

@article{muller2022instant,
  author        = {Thomas M\"uller and Alex Evans and Christoph Schied and Alexander Keller},
  title         = {Instant Neural Graphics Primitives with a Multiresolution Hash Encoding},
  journal       = {ACM Transactions on Graphics},
  volume        = {41},
  number        = {4},
  pages         = {102:1--102:15},
  eid           = {102},
  articleno     = {102},
  numpages      = {15},
  year          = {2022},
  url           = {https://doi.org/10.1145/3528223.3530127},
}

@inproceedings{chen2022tensorf,
  author        = {Anpei Chen and Zexiang Xu and Andreas Geiger and Jingyi Yu and Hao Su},
  title         = {{TensoRF}: Tensorial Radiance Fields},
  booktitle     = {European Conference on Computer Vision (ECCV)},
  year          = {2022},
}

@inproceedings{yu2021plenoxels,
  author        = {Sara Fridovich-Keil and Alex Yu and Matthew Tancik and Qinhong Chen and Benjamin Recht and Angjoo Kanazawa},
  title         = {{Plenoxels}: Radiance Fields without Neural Networks},
  booktitle     = {IEEE/CVF Conference on Computer Vision and Pattern Recognition (CVPR)},
  year          = {2022},
}

@inproceedings{fridovich2023kplanes,
  author        = {Sara Fridovich-Keil and Giacomo Meanti and Frederik Rahb{\ae}k Warburg and Benjamin Recht and Angjoo Kanazawa},
  title         = {{K-Planes}: Explicit Radiance Fields in Space, Time, and Appearance},
  booktitle     = {IEEE/CVF Conference on Computer Vision and Pattern Recognition (CVPR)},
  year          = {2023},
}

@inproceedings{povertymaps,
  author        = {Aiken, Emily and Rolf, Esther and Blumenstock, Joshua},
  title         = {Fairness and Representation in Satellite-Based Poverty Maps: Evidence of Urban-Rural Disparities and Their Impacts on Downstream Policy},
  booktitle     = {International Joint Conference on Artificial Intelligence (IJCAI)},
  pages         = {5888--5896},
  year          = {2023},
  url           = {https://doi.org/10.24963/ijcai.2023/653},
}

@article{poverty2,
  author        = {Jean, Neal and Burke, Marshall and Xie, Michael and Alampay Davis, W Matthew and Lobell, David B and Ermon, Stefano},
  title         = {Combining satellite imagery and machine learning to predict poverty},
  journal       = {Science},
  volume        = {353},
  number        = {6301},
  pages         = {790--794},
  year          = {2016},
}

@article{distastermapping,
  author        = {Al Shafian, Sultan and Hu, Da},
  title         = {Integrating Machine Learning and Remote Sensing in Disaster Management: A Decadal Review of Post-Disaster Building Damage Assessment},
  journal       = {Buildings},
  volume        = {14},
  number        = {8},
  eid           = {2344},
  year          = {2024},
  url           = {https://doi.org/10.3390/buildings14082344},
}

@article{ccai,
  author        = {Rolnick, David and Donti, Priya L. and Kaack, Lynn H. and Kochanski, Kelly and Lacoste, Alexandre and Sankaran, Kris and Ross, Andrew Slavin and Milojevic-Dupont, Nikola and Jaques, Natasha and Waldman-Brown, Anna and Luccioni, Alexandra Sasha and Maharaj, Tegan and Sherwin, Evan D. and Mukkavilli, S. Karthik and Kording, Konrad P. and Gomes, Carla P. and Ng, Andrew Y. and Hassabis, Demis and Platt, John C. and Creutzig, Felix and Chayes, Jennifer and Bengio, Yoshua},
  title         = {{Tackling} Climate Change with Machine Learning},
  journal       = {ACM Computing Surveys},
  volume        = {55},
  number        = {2},
  eid           = {42},
  articleno     = {42},
  numpages      = {96},
  year          = {2022},
  url           = {https://doi.org/10.1145/3485128},
}

@inproceedings{ecological1,
  author        = {You, Jiaxuan and Li, Xiaocheng and Low, Melvin and Lobell, David and Ermon, Stefano},
  title         = {Deep {Gaussian} Process for Crop Yield Prediction Based on Remote Sensing Data},
  booktitle     = {Proceedings of the AAAI Conference on Artificial Intelligence},
  volume        = {31},
  year          = {2017},
}

@inproceedings{cropharvest,
  author        = {Tseng, Gabriel and Zvonkov, Ivan and Nakalembe, Catherine Lilian and Kerner, Hannah},
  title         = {{CropHarvest}: A Global Dataset for Crop-Type Classification},
  booktitle     = {NeurIPS Datasets and Benchmarks Track},
  year          = {2021},
}

@article{earthembeddings,
  author        = {Klemmer, Konstantin and Rolf, Esther and Rußwurm, Marc and Camps-Valls, Gustau and Czerkawski, Mikolaj and Ermon, Stefano and Francis, Alistair and Jacobs, Nathan and Kerner, Hannah and Mackey, Lester and Mai, Gengchen and Mac Aodha, Oisin and Reichstein, Markus and Robinson, Caleb and Rolnick, David and Shelhamer, Evan and Sitzmann, Vincent and Tuia, Devis and Zhu, Xiaoxiang},
  title         = {{Earth} {Embeddings}: Toward artificial intelligence-centric representations of our planet},
  journal       = {IEEE Geoscience and Remote Sensing Magazine},
  pages         = {2--15},
  year          = {2026},
  url           = {https://doi.org/10.1109/MGRS.2026.3710416},
}

@misc{compressingearthembeddings1,
  author        = {Robinson, Caleb and Corley, Isaac},
  title         = {Compressing {Earth} Embeddings},
  year          = {2026},
  date          = {2026-03-24},
  url           = {https://geospatialml.com/posts/compressing-earth-embeddings/},
  langid        = {en},
}

@inproceedings{tessera,
  author        = {Feng, Zhengpeng and Atzberger, Clement and Jaffer, Sadiq and Knezevic, Jovana and Sormunen, Silja and Young, Robin and Lisaius, Madeline C. and Immitzer, Markus and Jackson, Toby and Ball, James and Coomes, David A. and Madhavapeddy, Anil and Blake, Andrew and Keshav, Srinivasan},
  title         = {{TESSERA}: Temporal Embeddings of Surface Spectra for {Earth} Representation and Analysis},
  booktitle     = {IEEE/CVF Conference on Computer Vision and Pattern Recognition (CVPR)},
  pages         = {34818--34831},
  year          = {2026},
}

@inproceedings{geoinrid,
  author        = {Arjun Rao and Marc Ru{\ss}wurm and Konstantin Klemmer and Esther Rolf},
  title         = {Measuring the Intrinsic Dimension of {Earth} Representations},
  booktitle     = {International Conference on Learning Representations (ICLR)},
  year          = {2026},
  url           = {https://openreview.net/forum?id=gQPD83DrGp},
}

@inproceedings{kobayashi2022dff,
  author        = {Sosuke Kobayashi and Eiichi Matsumoto and Vincent Sitzmann},
  title         = {Decomposing {NeRF} for Editing via Feature Field Distillation},
  booktitle     = {Advances in Neural Information Processing Systems (NeurIPS)},
  year          = {2022},
}

@inproceedings{tschernezki2022n3f,
  author        = {Vadim Tschernezki and Iro Laina and Diane Larlus and Andrea Vedaldi},
  title         = {Neural Feature Fusion Fields: {3D} Distillation of Self-Supervised {2D} Image Representations},
  booktitle     = {International Conference on 3D Vision (3DV)},
  year          = {2022},
}

@inproceedings{kerr2023lerf,
  author        = {Justin Kerr and Chung Min Kim and Ken Goldberg and Angjoo Kanazawa and Matthew Tancik},
  title         = {{LERF}: Language Embedded Radiance Fields},
  booktitle     = {IEEE/CVF International Conference on Computer Vision (ICCV)},
  year          = {2023},
}

@inproceedings{shen2023f3rm,
  author        = {William Shen and Ge Yang and Alan Yu and Jansen Wong and Leslie Pack Kaelbling and Phillip Isola},
  title         = {Distilled Feature Fields Enable Few-Shot Language-Guided Manipulation},
  booktitle     = {Conference on Robot Learning (CoRL)},
  year          = {2023},
}

@inproceedings{ye2023featurenerf,
  author        = {Jianglong Ye and Naiyan Wang and Xiaolong Wang},
  title         = {{FeatureNeRF}: Learning Generalizable {NeRFs} by Distilling Foundation Models},
  booktitle     = {IEEE/CVF International Conference on Computer Vision (ICCV)},
  year          = {2023},
}

@inproceedings{chen2019learning,
  author        = {Chen, Zhiqin and Zhang, Hao},
  title         = {Learning implicit fields for generative shape modeling},
  booktitle     = {IEEE/CVF Conference on Computer Vision and Pattern Recognition (CVPR)},
  pages         = {5939--5948},
  year          = {2019},
}

@inproceedings{chen2021nerv,
  author        = {Chen, Hao and He, Bo and Wang, Hanyu and Ren, Yixuan and Lim, Ser-Nam and Shrivastava, Abhinav},
  title         = {{NeRV}: Neural Representations for Videos},
  booktitle     = {Advances in Neural Information Processing Systems (NeurIPS)},
  year          = {2021},
}

@inproceedings{dupont2021coin,
  author        = {Emilien Dupont and Adam Golinski and Milad Alizadeh and Yee Whye Teh and Arnaud Doucet},
  title         = {{COIN}: {CO}mpression with Implicit Neural representations},
  booktitle     = {ICLR Workshop on Neural Compression: From Information Theory to Applications},
  year          = {2021},
  url           = {https://openreview.net/forum?id=yekxhcsVi4},
}

@inproceedings{inrcompress,
  author        = {Str{\"u}mpler, Yannick and Postels, Janis and Yang, Ren and Van Gool, Luc and Tombari, Federico},
  title         = {Implicit Neural Representations for Image Compression},
  booktitle     = {European Conference on Computer Vision (ECCV)},
  year          = {2022},
}

@article{dupont2022coin,
  author        = {Emilien Dupont and Hrushikesh Loya and Milad Alizadeh and Adam Golinski and Yee Whye Teh and Arnaud Doucet},
  title         = {{COIN}++: Neural Compression Across Modalities},
  journal       = {Transactions on Machine Learning Research},
  year          = {2022},
  url           = {https://openreview.net/forum?id=NXB0rEM2Tq},
}

@article{xu2022nesvor,
  author        = {Xu, Junshen and Moyer, Daniel and Gagoski, Borjan and Iglesias, Juan Eugenio and Grant, P. Ellen and Golland, Polina and Adalsteinsson, Elfar},
  title         = {{NeSVoR}: Implicit Neural Representation for Slice-to-Volume Reconstruction in {MRI}},
  journal       = {IEEE Transactions on Medical Imaging},
  volume        = {42},
  number        = {6},
  pages         = {1707--1719},
  year          = {2023},
}

@inproceedings{park2019deepsdf,
  author        = {Park, Jeong Joon and Florence, Peter and Straub, Julian and Newcombe, Richard and Lovegrove, Steven},
  title         = {{DeepSDF}: Learning Continuous Signed Distance Functions for Shape Representation},
  booktitle     = {IEEE/CVF Conference on Computer Vision and Pattern Recognition (CVPR)},
  year          = {2019},
}

@inproceedings{barron2021mipnerf,
  author        = {Barron, Jonathan T. and Mildenhall, Ben and Tancik, Matthew and Hedman, Peter and Martin-Brualla, Ricardo and Srinivasan, Pratul P.},
  title         = {{Mip-NeRF}: A Multiscale Representation for Anti-Aliasing Neural Radiance Fields},
  booktitle     = {IEEE/CVF International Conference on Computer Vision (ICCV)},
  year          = {2021},
}

@inproceedings{yu2021plenoctrees,
  author        = {Yu, Alex and Li, Ruilong and Tancik, Matthew and Li, Hao and Ng, Ren and Kanazawa, Angjoo},
  title         = {{PlenOctrees} for Real-Time Rendering of Neural Radiance Fields},
  booktitle     = {IEEE/CVF International Conference on Computer Vision (ICCV)},
  year          = {2021},
}

@inproceedings{peng2020convolutional,
  author        = {Peng, Songyou and Niemeyer, Michael and Mescheder, Lars and Pollefeys, Marc and Geiger, Andreas},
  title         = {Convolutional Occupancy Networks},
  booktitle     = {European Conference on Computer Vision (ECCV)},
  year          = {2020},
}

@inproceedings{liu2020neural,
  author        = {Liu, Lingjie and Gu, Jiatao and Lin, Kyaw Zaw and Chua, Tat-Seng and Theobalt, Christian},
  title         = {Neural Sparse Voxel Fields},
  booktitle     = {Advances in Neural Information Processing Systems (NeurIPS)},
  year          = {2020},
}

@inproceedings{grids,
  author        = {Namhoon Kim and Sara Fridovich-Keil},
  title         = {Grids Often Outperform Implicit Neural Representation at Compressing Dense Signals},
  booktitle     = {Advances in Neural Information Processing Systems (NeurIPS)},
  year          = {2025},
  url           = {https://openreview.net/forum?id=OZljvntsto},
}

@inproceedings{turki2022meganerf,
  author        = {Turki, Haithem and Ramanan, Deva and Satyanarayanan, Mahadev},
  title         = {{Mega-NeRF}: Scalable Construction of Large-Scale {NeRFs} for Virtual Fly-Throughs},
  booktitle     = {IEEE/CVF Conference on Computer Vision and Pattern Recognition (CVPR)},
  pages         = {12922--12931},
  year          = {2022},
}

@inproceedings{tancik2022blocknerf,
  author        = {Tancik, Matthew and Casser, Vincent and Yan, Xinchen and Pradhan, Sabeek and Mildenhall, Ben and Srinivasan, Pratul P. and Barron, Jonathan T. and Kretzschmar, Henrik},
  title         = {{Block-NeRF}: Scalable Large Scene Neural View Synthesis},
  booktitle     = {IEEE/CVF Conference on Computer Vision and Pattern Recognition (CVPR)},
  year          = {2022},
}

@inproceedings{mi2023switchnerf,
  author        = {Zhenxing Mi and Dan Xu},
  title         = {{Switch-NeRF}: Learning Scene Decomposition with Mixture of Experts for Large-Scale Neural Radiance Fields},
  booktitle     = {International Conference on Learning Representations (ICLR)},
  year          = {2023},
  url           = {https://openreview.net/forum?id=PQ2zoIZqvm},
}

@inproceedings{functa22,
  author        = {Dupont, Emilien and Kim, Hyunjik and Eslami, S. M. Ali and Rezende, Danilo Jimenez and Rosenbaum, Dan},
  title         = {From data to functa: Your data point is a function and you can treat it like one},
  booktitle     = {International Conference on Machine Learning (ICML)},
  year          = {2022},
}

@article{spatialfuncta23,
  author        = {Bauer, Matthias and Dupont, Emilien and Brock, Andy and Rosenbaum, Dan and Schwarz, Jonathan Richard and Kim, Hyunjik},
  title         = {Spatial Functa: Scaling Functa to {ImageNet} Classification and Generation},
  journal       = {arXiv preprint arXiv: 2302.03130},
  year          = {2023},
  arxiv         = {https://arxiv.org/abs/2302.03130},
}

@inproceedings{geoclip,
  author        = {Vicente Vivanco Cepeda and Gaurav Kumar Nayak and Mubarak Shah},
  title         = {{GeoCLIP}: {CLIP}-Inspired Alignment between Locations and Images for Effective Worldwide Geo-Localization},
  booktitle     = {Advances in Neural Information Processing Systems (NeurIPS)},
  year          = {2023},
  url           = {https://openreview.net/forum?id=I18BXotQ7j},
}

@inproceedings{csp,
  author        = {Mai, Gengchen and Lao, Ni and He, Yutong and Song, Jiaming and Ermon, Stefano},
  title         = {{CSP}: Self-Supervised Contrastive Spatial Pre-Training for Geospatial-Visual Representations},
  booktitle     = {International Conference on Machine Learning (ICML)},
  year          = {2023},
}

@inproceedings{dhakal2025range,
  author        = {Dhakal, Aayush and Sastry, Srikumar and Khanal, Subash and Ahmad, Adeel and Xing, Eric and Jacobs, Nathan},
  title         = {{RANGE}: Retrieval Augmented Neural Fields for Multi-Resolution Geo-Embeddings},
  booktitle     = {IEEE/CVF Conference on Computer Vision and Pattern Recognition (CVPR)},
  pages         = {24680--24689},
  year          = {2025},
}

@inproceedings{cole2023spatial,
  author        = {Cole, Elijah and Van Horn, Grant and Lange, Christian and Shepard, Alexander and Leary, Patrick and Perona, Pietro and Loarie, Scott and Mac Aodha, Oisin},
  title         = {Spatial Implicit Neural Representations for Global-Scale Species Mapping},
  booktitle     = {International Conference on Machine Learning (ICML)},
  volume        = {202},
  pages         = {6320--6342},
  series        = {Proceedings of Machine Learning Research},
  year          = {2023},
  url           = {https://proceedings.mlr.press/v202/cole23a.html},
}

@inproceedings{satclip,
  author        = {Klemmer, Konstantin and Rolf, Esther and Robinson, Caleb and Mackey, Lester and Ru{\ss}wurm, Marc},
  title         = {{SatCLIP}: Global, general-purpose location embeddings with satellite imagery},
  booktitle     = {Proceedings of the AAAI Conference on Artificial Intelligence},
  volume        = {39},
  pages         = {4347--4355},
  year          = {2025},
}

@article{alphaearth,
  author        = {Christopher F. Brown and Michal R. Kazmierski and Valerie J. Pasquarella and William J. Rucklidge and Masha Samsikova and Chenhui Zhang and Evan Shelhamer and Estefania Lahera and Olivia Wiles and Simon Ilyushchenko and Noel Gorelick and Lihui Lydia Zhang and Sophia Alj and Emily Schechter and Sean Askay and Oliver Guinan and Rebecca Moore and Alexis Boukouvalas and Pushmeet Kohli},
  title         = {{AlphaEarth Foundations}: An Embedding Field Model for Accurate and Efficient Global Mapping from Sparse Label Data},
  journal       = {arXiv preprint arXiv:2507.22291},
  year          = {2025},
}

@inproceedings{wrap,
  author        = {Mac Aodha, Oisin and Cole, Elijah and Perona, Pietro},
  title         = {Presence-only geographical priors for fine-grained image classification},
  booktitle     = {IEEE/CVF International Conference on Computer Vision (ICCV)},
  pages         = {9596--9606},
  year          = {2019},
}

@inproceedings{sh,
  author        = {Marc Ru{\ss}wurm and Konstantin Klemmer and Esther Rolf and Robin Zbinden and Devis Tuia},
  title         = {Geographic Location Encoding with Spherical Harmonics and Sinusoidal Representation Networks},
  booktitle     = {International Conference on Learning Representations (ICLR)},
  year          = {2024},
  url           = {https://iclr.cc/virtual/2024/poster/18690},
}

@article{mai2022review,
  author        = {Mai, Gengchen and Janowicz, Krzysztof and Hu, Yingjie and Gao, Song and Yan, Bo and Zhu, Rui and Cai, Ling and Lao, Ni},
  title         = {A Review of Location Encoding for {GeoAI}: Methods and Applications},
  journal       = {International Journal of Geographical Information Science},
  volume        = {36},
  number        = {4},
  pages         = {639--673},
  year          = {2022},
}

@inproceedings{rao2026localized,
  author        = {Rao, Arjun and Crasto, Ruth and Ooms, Tessa and Rolnick, David and Klemmer, Konstantin and Ru{\ss}wurm, Marc},
  title         = {Localized, High-Resolution Geographic Representations with {Slepian} Functions},
  booktitle     = {International Conference on Machine Learning (ICML)},
  year          = {2026},
  url           = {https://openreview.net/forum?id=eWQQ0tO0kB},
}

@article{terracodec,
  author        = {Costa-Watanabe, Julen and Wittmann, Isabelle and Blumenstiel, Benedikt and Schindler, Konrad},
  title         = {{TerraCodec}: Compressing Optical {Earth} Observation Data},
  journal       = {arXiv preprint arXiv:2510.12670},
  year          = {2025},
}

@inproceedings{cao2023hexplane,
  author        = {Cao, Ang and Johnson, Justin},
  title         = {{HexPlane}: A Fast Representation for Dynamic Scenes},
  booktitle     = {IEEE/CVF Conference on Computer Vision and Pattern Recognition (CVPR)},
  pages         = {130--141},
  year          = {2023},
}

@article{kerbl3Dgaussians,
  author        = {Kerbl, Bernhard and Kopanas, Georgios and Leimk{\"u}hler, Thomas and Drettakis, George},
  title         = {{3D} {Gaussian} Splatting for Real-Time Radiance Field Rendering},
  journal       = {ACM Transactions on Graphics},
  volume        = {42},
  number        = {4},
  pages         = {139:1--139:14},
  eid           = {139},
  numpages      = {14},
  year          = {2023},
  url           = {https://doi.org/10.1145/3592433},
}

@inproceedings{cher2026tte,
  author        = {Cher, Daniel and Iqbal, Hamza and Xing, Eric and Wei, Brian and Jacobs, Nathan},
  title         = {Tessellating the {Earth}: Learnable Spherical {Voronoi} Partitions for Location Encoding},
  booktitle     = {European Conference on Computer Vision (ECCV)},
  year          = {2026},
}

@article{presto,
  author        = {Tseng, Gabriel and Cartuyvels, Ruben and Zvonkov, Ivan and Purohit, Mirali and Rolnick, David and Kerner, Hannah},
  title         = {Lightweight, pre-trained transformers for remote sensing timeseries},
  journal       = {arXiv preprint arXiv:2304.14065},
  year          = {2023},
}

@inproceedings{galileo,
  author        = {Tseng, Gabriel and Fuller, Anthony and Reil, Marlena and Herzog, Henry and Beukema, Patrick and Bastani, Favyen and Green, James R and Shelhamer, Evan and Kerner, Hannah and Rolnick, David},
  title         = {{Galileo}: Learning Global \& Local Features of Many Remote Sensing Modalities},
  booktitle     = {International Conference on Machine Learning (ICML)},
  volume        = {267},
  pages         = {60280--60300},
  series        = {Proceedings of Machine Learning Research},
  year          = {2025},
  url           = {https://proceedings.mlr.press/v267/tseng25a.html},
}

@techreport{cog,
  author        = {Pollack, Nathan},
  title         = {Cloud Optimized {GeoTIFF} ({COG}) File Format},
  number        = {ESDS-RFC-049},
  institution   = {NASA Earth Science Data and Information System Standards Coordination Office},
  year          = {2024},
  url           = {https://doi.org/10.5067/DOC/ESCO/ESDS-RFC-049v1},
}

@inproceedings{benv2,
  author        = {Clasen, Kai Norman and Hackel, Leonard and Burgert, Tom and Sumbul, Gencer and Demir, Beg{\"u}m and Markl, Volker},
  title         = {{reBEN}: Refined {BigEarthNet} Dataset for Remote Sensing Image Analysis},
  booktitle     = {IEEE International Geoscience and Remote Sensing Symposium (IGARSS)},
  pages         = {1264--1268},
  year          = {2025},
  url           = {https://doi.org/10.1109/IGARSS55030.2025.11242834},
}

@article{so2sat,
  author        = {Zhu, Xiao Xiang and Hu, Jingliang and Qiu, Chunping and Shi, Yilei and Kang, Jian and Mou, Lichao and Bagheri, Hossein and Haberle, Matthias and Hua, Yuansheng and Huang, Rong and Hughes, Lloyd and Li, Hao and Sun, Yao and Zhang, Guichen and Han, Shiyao and Schmitt, Michael and Wang, Yuanyuan},
  title         = {{So2Sat LCZ42}: A Benchmark Data Set for the Classification of Global Local Climate Zones [Software and Data Sets]},
  journal       = {IEEE Geoscience and Remote Sensing Magazine},
  volume        = {8},
  number        = {3},
  pages         = {76--89},
  year          = {2020},
  url           = {https://doi.org/10.1109/MGRS.2020.2964708},
}

@inproceedings{pixelnerf,
  author        = {Yu, Alex and Ye, Vickie and Tancik, Matthew and Kanazawa, Angjoo},
  title         = {{pixelNeRF}: Neural Radiance Fields from One or Few Images},
  booktitle     = {IEEE/CVF Conference on Computer Vision and Pattern Recognition (CVPR)},
  pages         = {4576--4585},
  year          = {2021},
}

@inproceedings{mvsnerf,
  author        = {Chen, Anpei and Xu, Zexiang and Zhao, Fuqiang and Zhang, Xiaoshuai and Xiang, Fanbo and Yu, Jingyi and Su, Hao},
  title         = {{MVSNeRF}: Fast Generalizable Radiance Field Reconstruction from Multi-View Stereo},
  booktitle     = {IEEE/CVF International Conference on Computer Vision (ICCV)},
  pages         = {14124--14133},
  year          = {2021},
}

@article{skill,
  author        = {Murphy, Allan H},
  title         = {{Skill} scores based on the mean square error and their relationships to the correlation coefficient},
  journal       = {Monthly Weather Review},
  volume        = {116},
  number        = {12},
  pages         = {2417--2424},
  year          = {1988},
}

@inproceedings{pastis,
  author        = {Sainte Fare Garnot, Vivien and Landrieu, Lo{\"i}c},
  title         = {Panoptic Segmentation of Satellite Image Time Series with Convolutional Temporal Attention Networks},
  booktitle     = {IEEE/CVF International Conference on Computer Vision (ICCV)},
  pages         = {4852--4861},
  year          = {2021},
  url           = {https://doi.org/10.1109/ICCV48922.2021.00483},
}

@inproceedings{swisscrop25,
  author        = {Thomas Lauber and Mehmet Ozgur Turkoglu and S{\'e}l{\`e}ne Ledain and Helge Aasen},
  title         = {{SwissCrop25}: A National Multi-Year Benchmark for Operational Crop Mapping},
  booktitle     = {ECCV Workshop TerraBytes II},
  year          = {2026},
  url           = {https://openreview.net/forum?id=17VbDRJojb},
}

@article{ftw,
  author        = {Kerner, Hannah and Chaudhari, Snehal and Ghosh, Aninda and Robinson, Caleb and Ahmad, Adeel and Choi, Eddie and Jacobs, Nathan and Holmes, Chris and Mohr, Matthias and Dodhia, Rahul and Lavista Ferres, Juan M and Marcus, Jennifer},
  title         = {{Fields of The World}: A Machine Learning Benchmark Dataset for Global Agricultural Field Boundary Segmentation},
  journal       = {Proceedings of the AAAI Conference on Artificial Intelligence},
  volume        = {39},
  number        = {27},
  pages         = {28151--28159},
  year          = {2025},
  url           = {https://doi.org/10.1609/aaai.v39i27.35034},
}

@article{eurominenet,
  author        = {Yu, Weikang and Nwazelibe, Vincent and Ma, Xianping and Zhang, Xiaokang and Gloaguen, Richard and Zhu, Xiao Xiang and Ghamisi, Pedram},
  title         = {{EuroMineNet}: A Multitemporal {Sentinel-2} Benchmark for Spatiotemporal Mining Footprint Analysis in the {European Union} (2015--2024)},
  journal       = {ISPRS Journal of Photogrammetry and Remote Sensing},
  volume        = {237},
  pages         = {409--425},
  year          = {2026},
}

@inproceedings{pooling,
  author        = {Corley, Isaac and Robinson, Caleb and Lavista Ferres, Juan M. and Becker-Reshef, Inbal},
  title         = {From Pixels to Patches: Pooling Strategies for {Earth} Embeddings},
  booktitle     = {ICLR Workshop on Machine Learning for Remote Sensing},
  year          = {2026},
  url           = {https://openreview.net/forum?id=oPfOVLFIrU},
}

@inproceedings{lianet,
  author        = {Madadikhaljan, Mojgan and Prexl, Jonathan and Wittmann, Isabelle and Albrecht, Conrad M and Schmitt, Michael},
  title         = {Location Is All You Need: Continuous Spatiotemporal Neural Representations of {Earth} Observation Data},
  booktitle     = {IEEE/CVF Conference on Computer Vision and Pattern Recognition (CVPR) Workshops},
  pages         = {8000--8010},
  year          = {2026},
}

@article{eeapi,
  author        = {Gorelick, Noel and Hancher, Matt and Dixon, Mike and Ilyushchenko, Simon and Thau, David and Moore, Rebecca},
  title         = {{Google Earth Engine}: Planetary-Scale Geospatial Analysis for Everyone},
  journal       = {Remote Sensing of Environment},
  volume        = {202},
  pages         = {18--27},
  year          = {2017},
  url           = {https://doi.org/10.1016/j.rse.2017.06.031},
}

@incollection{van2024continual,
  author        = {Gido M. {van de Ven} and Nicholas Soures and Dhireesha Kudithipudi},
  title         = {1.09 - Continual learning and catastrophic forgetting},
  editor        = {John Wixted},
  booktitle     = {Learning and Memory: A Comprehensive Reference},
  edition       = {Third},
  volume        = {1},
  pages         = {153--168},
  publisher     = {Academic Press},
  year          = {2025},
  url           = {https://doi.org/10.1016/B978-0-443-15754-7.00073-0},
}

@article{canopy,
  author        = {Lang, Nico and Jetz, Walter and Schindler, Konrad and Wegner, Jan Dirk},
  title         = {A High-Resolution Canopy Height Model of the {Earth}},
  journal       = {Nature Ecology \& Evolution},
  volume        = {7},
  number        = {11},
  pages         = {1778--1789},
  year          = {2023},
}

@inproceedings{tilted,
  author        = {Yi, Brent and Zeng, Weijia and Buchanan, Sam and Ma, Yi},
  title         = {Canonical factors for hybrid neural fields},
  booktitle     = {IEEE/CVF International Conference on Computer Vision (ICCV)},
  pages         = {3391--3403},
  year          = {2023},
}

@article{sentinel2,
  author        = {Drusch, M. and Del Bello, U. and Carlier, S. and Colin, O. and Fernandez, V. and Gascon, F. and Hoersch, B. and Isola, C. and Laberinti, P. and Martimort, P. and Meygret, A. and Spoto, F. and Sy, O. and Marchese, F. and Bargellini, P.},
  title         = {{Sentinel-2}: {ESA}'s Optical High-Resolution Mission for {GMES} Operational Services},
  journal       = {Remote Sensing of Environment},
  volume        = {120},
  pages         = {25--36},
  year          = {2012},
  url           = {https://doi.org/10.1016/j.rse.2011.11.026},
}

@article{sentinel1,
  author        = {Torres, R. and Snoeij, P. and Geudtner, D. and Bibby, D. and Davidson, M. and Attema, E. and Potin, P. and Rommen, B. and Floury, N. and Brown, M. and Navas Traver, I. and Deghaye, P. and Duesmann, B. and Rosich, B. and Miranda, N. and Bruno, C. and L'Abbate, M. and Croci, R. and Pietropaolo, A. and Huchler, M. and Rostan, F.},
  title         = {{GMES Sentinel-1} mission},
  journal       = {Remote Sensing of Environment},
  volume        = {120},
  pages         = {9--24},
  year          = {2012},
  url           = {https://doi.org/10.1016/j.rse.2011.05.028},
}

@misc{landsatc2,
  author        = {{EROS Center}},
  title         = {{Landsat 8-9 Operational Land Imager / Thermal Infrared Sensor Level-2, Collection 2}},
  year          = {2020},
  howpublished  = {U.S. Geological Survey},
  note          = {Dataset},
  url           = {https://doi.org/10.5066/P9OGBGM6},
}

@article{palsar,
  author        = {Shimada, Masanobu and Itoh, Takuya and Motooka, Takeshi and Watanabe, Manabu and Shiraishi, Tomohiro and Thapa, Rajesh and Lucas, Richard},
  title         = {New global forest/non-forest maps from {ALOS PALSAR} data (2007--2010)},
  journal       = {Remote Sensing of Environment},
  volume        = {155},
  pages         = {13--31},
  year          = {2014},
  url           = {https://doi.org/10.1016/j.rse.2014.04.014},
}

@article{gpw,
  author        = {Parente, Leandro and Sloat, Lindsey and Mesquita, Vinicius and Consoli, Davide and Stanimirova, Radost and Hengl, Tomislav and Bonannella, Carmelo and Teles, Nath{\'a}lia and Wheeler, Ichsani and Hunter, Maria and Ehrmann, Steffen and Ferreira, Laerte and Mattos, Ana Paula and Oliveira, Bernard and Meyer, Carsten and {\c{S}}ahin, Murat and Witjes, Martijn and Fritz, Steffen and Malek, Ziga and Stolle, Fred},
  title         = {Annual 30-m maps of global grassland class and extent (2000--2022) based on spatiotemporal machine learning},
  journal       = {Scientific Data},
  volume        = {11},
  pages         = {1303},
  eid           = {1303},
  year          = {2024},
  url           = {https://doi.org/10.1038/s41597-024-04139-6},
}

@misc{biomasscci7,
  author        = {Santoro, M. and Cartus, O.},
  title         = {{ESA Biomass Climate Change Initiative (Biomass\_cci)}: Global datasets of forest above-ground biomass for the years 2005--2012 and 2015--2024, v7.0},
  year          = {2026},
  month         = may,
  howpublished  = {NERC EDS Centre for Environmental Data Analysis},
  version       = {7.0},
  note          = {Dataset, published 21 May 2026},
  url           = {https://doi.org/10.5285/6429d1aafe1e43b9b414e4a5a7f8b903},
}

@article{gladwater,
  author        = {Pickens, A. H. and Hansen, M. C. and Hancher, M. and Stehman, S. V. and Tyukavina, A. and Potapov, P. and Marroquin, B. and Sherani, Z.},
  title         = {Mapping and sampling to characterize global inland water dynamics from 1999 to 2018 with full {Landsat} time-series},
  journal       = {Remote Sensing of Environment},
  volume        = {243},
  pages         = {111792},
  eid           = {111792},
  year          = {2020},
  url           = {https://doi.org/10.1016/j.rse.2020.111792},
}

@misc{copdem,
  author        = {{European Space Agency} and {Airbus}},
  title         = {{Copernicus DEM} -- Global and {European} Digital Elevation Model},
  year          = {2022},
  howpublished  = {European Space Agency},
  note          = {Dataset collection; GLO-30 instance},
  url           = {https://doi.org/10.5270/ESA-c5d3d65},
}

@article{glwd2,
  author        = {Lehner, Bernhard and Anand, Mira and Fluet-Chouinard, Etienne and Tan, Florence and Aires, Filipe and Allen, George H. and Bousquet, Philippe and Canadell, Josep G. and Davidson, Nick and Ding, Meng and Finlayson, C. Max and Gumbricht, Thomas and Hilarides, Lammert and Hugelius, Gustaf and Jackson, Robert B. and Korver, Maartje C. and Liu, Liangyun and McIntyre, Peter B. and Nagy, Szabolcs and Olefeldt, David and Pavelsky, Tamlin M. and Pekel, Jean-Francois and Poulter, Benjamin and Prigent, Catherine and Wang, Jida and Worthington, Thomas A. and Yamazaki, Dai and Zhang, Xiao and Thieme, Michele},
  title         = {Mapping the world's inland surface waters: an upgrade to the {Global Lakes and Wetlands Database (GLWD v2)}},
  journal       = {Earth System Science Data},
  volume        = {17},
  pages         = {2277--2329},
  year          = {2025},
  url           = {https://doi.org/10.5194/essd-17-2277-2025},
}

@article{soilgrids,
  author        = {Poggio, Laura and de Sousa, Luis M. and Batjes, Niels H. and Heuvelink, Gerard B. M. and Kempen, Bas and Ribeiro, Eloi and Rossiter, David},
  title         = {{SoilGrids 2.0}: producing soil information for the globe with quantified spatial uncertainty},
  journal       = {SOIL},
  volume        = {7},
  pages         = {217--240},
  year          = {2021},
  url           = {https://doi.org/10.5194/soil-7-217-2021},
}

@inproceedings{climplicit,
  author        = {Dollinger, Johannes and Robert, Damien and Plekhanova, Elena and Drees, Lukas and Wegner, Jan Dirk},
  title         = {{Climplicit}: Climatic Implicit Embeddings for Global Ecological Tasks},
  booktitle     = {ICLR Workshop on Tackling Climate Change with Machine Learning},
  year          = {2025},
}

@article{czerkawski2024global,
  author        = {Czerkawski, Mikolaj and Kluczek, Marcin and Bojanowski, J{\k{e}}drzej S.},
  title         = {Global and Dense Embeddings of {Earth}: {Major TOM} Floating in the Latent Space},
  journal       = {arXiv preprint arXiv:2412.05600},
  year          = {2024},
}

@misc{ai2_olmoearth_platform_2026,
  author        = {{Ai2}},
  title         = {The {OlmoEarth} Platform: Geospatial Inference at Planetary Scale},
  year          = {2026},
  month         = jul,
  day           = {28},
  note          = {Accessed: 2026-09-10},
  url           = {https://allenai.org/blog/olmoearth-infrastructure},
  urldate       = {2026-09-10},
}

@misc{naturalearth,
  author        = {{Natural Earth}},
  title         = {Admin 0 -- Countries (1:110m Cultural Vectors)},
  year          = {2024},
  howpublished  = {\url{https://www.naturalearthdata.com/downloads/110m-cultural-vectors/110m-admin-0-countries/}},
  note          = {Accessed: 2026-09-22},
}

@article{peel2007updated,
  author        = {Peel, Murray C and Finlayson, Brian L and McMahon, Thomas A},
  title         = {Updated World Map of the {K{\"o}ppen-Geiger} Climate Classification},
  journal       = {Hydrology and Earth System Sciences},
  volume        = {11},
  number        = {5},
  pages         = {1633--1644},
  year          = {2007},
}

@misc{source_cooperative,
  author        = {{Radiant Earth}},
  title         = {{Source Cooperative}},
  year          = {2026},
  howpublished  = {\url{https://source.coop}},
  note          = {Accessed: 2026-09-13},
}

@inproceedings{miam,
  author        = {Robin Zbinden and Wesley Monteith-Finas and Gencer Sumbul and Nina van Tiel and Chiara Vanalli and Devis Tuia},
  title         = {{MIAM}: Modality Imbalance-Aware Masking for Multimodal Ecological Applications},
  booktitle     = {International Conference on Learning Representations (ICLR)},
  year          = {2026},
  url           = {https://openreview.net/forum?id=oljjAkgZN4},
}

@article{lane2026sled,
  author        = {Lane, Kevin and Wang, Zhongying and Rolf, Esther and Karimzadeh, Morteza},
  title         = {{SLED}: Scalable Location Encoding via Distillation},
  journal       = {arXiv preprint arXiv:2608.06612},
  year          = {2026},
}

@inproceedings{strainer,
  author        = {Vyas, Kushal and Humayun, Ahmed Imtiaz and Dashpute, Aniket and Baraniuk, Richard G. and Veeraraghavan, Ashok and Balakrishnan, Guha},
  title         = {Learning Transferable Features for Implicit Neural Representations},
  booktitle     = {Advances in Neural Information Processing Systems (NeurIPS)},
  volume        = {37},
  pages         = {42268--42291},
  year          = {2024},
  url           = {https://doi.org/10.52202/079017-1337},
}

@article{liu2024cd,
  author        = {Liu, Zhenhuan and Liu, Shuai and Ning, Zhiwei and Yang, Jie and Zuo, Yifan and Fang, Yuming and Liu, Wei},
  title         = {{CD-NGP}: A Fast Scalable Continual Representation for Dynamic Scenes},
  journal       = {arXiv preprint arXiv:2409.05166},
  year          = {2024},
}

@inproceedings{hiner,
  author        = {Shi, Junqi and Jiang, Mingyi and Lu, Ming and Chen, Tong and Cao, Xun and Ma, Zhan},
  title         = {{HINER}: Neural Representation for Hyperspectral Image},
  booktitle     = {ACM International Conference on Multimedia (ACM MM)},
  pages         = {9837--9846},
  year          = {2024},
}

@inproceedings{huang2023compressing,
  author        = {Langwen Huang and Torsten Hoefler},
  title         = {Compressing multidimensional weather and climate data into neural networks},
  booktitle     = {International Conference on Learning Representations (ICLR)},
  year          = {2023},
  url           = {https://openreview.net/forum?id=Y5SEe3dfniJ},
}

@article{simumba2025geo,
  author        = {Naomi Simumba and Nils Lehmann and Paolo Fraccaro and Hamed Alemohammad and Geeth De Mel and Salman Khan and Manil Maskey and Nicolas Long{\'e}p{\'e} and Xiao Xiang Zhu and Hannah Kerner and Juan Bernabe Moreno and Alexandre Lacoste},
  title         = {{GEO-Bench-2}: From Performance to Capability, Rethinking Evaluation in Geospatial {AI}},
  journal       = {Transactions on Machine Learning Research},
  year          = {2026},
  url           = {https://openreview.net/forum?id=NPf175jnP1},
}

@inproceedings{gordon2026mmearthbench,
  author        = {Lucia Gordon and Serge Belongie and Christian Igel and Nico Lang},
  title         = {{MMEarth-Bench}: Global Model Adaptation via Multimodal Test-Time Training},
  booktitle     = {European Conference on Computer Vision (ECCV)},
  year          = {2026},
}

@inproceedings{raomm,
  author        = {Rao, Arjun and Rolf, Esther},
  title         = {{Using} Multiple Input Modalities can Improve Data-Efficiency and {O.O.D.} Generalization for {ML} with Satellite Imagery},
  booktitle     = {TerraBytes ICML Workshop: Towards Global Datasets and Models for Earth Observation},
  volume        = {292},
  pages         = {166--188},
  series        = {Proceedings of Machine Learning Research},
  year          = {2025},
  url           = {https://proceedings.mlr.press/v292/rao25a.html},
}

@article{van2026better,
  author        = {van der Plas, Thijs L and Bakermans, Jacob JW and Nedungadi, Vishal and Tij{\=u}naityt{\.e}, Gabriel{\.e} and Ru{\ss}wurm, Marc and Athanasiadis, Ioannis N},
  title         = {Better Together: Evaluating the Complementarity of {Earth} Embedding Models},
  journal       = {arXiv preprint arXiv:2605.18667},
  year          = {2026},
}

@article{mind,
  author        = {Isaac Corley and Arjun Rao and Esther Rolf and Konstantin Klemmer and Evan Shelhamer and Nils Lehmann and Marc Rußwurm and Gengchen Mai and Nathan Jacobs and Hannah Kerner},
  title         = {{MIND} the Gap: A Geographic Implicit Neural Representation with Adjustable Spatial Scale},
  journal       = {arXiv preprint arXiv:2609.25454},
  year          = {2026},
}

@inproceedings{jyhne2026superf,
  author        = {Jyhne, Sander R and Igel, Christian and Goodwin, Morten and Andersen, Per-Arne and Belongie, Serge and Lang, Nico},
  title         = {{SuperF}: Neural implicit fields for multi-image super-resolution},
  booktitle     = {International Conference on Learning Representations (ICLR)},
  volume        = {2026},
  pages         = {99513--99536},
  year          = {2026},
  url           = {https://openreview.net/forum?id=FiiItlSqqL},
}
\bibliographystyle{iclr2026_conference}
\clearpage
\appendix

\appendix

\newcommand{\midsize}{\fontsize{8pt}{9.2pt}\selectfont}
\begin{table}[h]
\caption{\textbf{EO products used to train PFFs.}}
\vspace{2pt}
\label{tab:modalities}
\setlength{\tabcolsep}{3.4pt}
\renewcommand{\arraystretch}{1.18}
\centering\midsize
\begin{tabular}{@{}c l l r r c l r@{}}
& Product & Source & $d$ & GSD & Years & Units & Precision \\
\midrule
\multirow{5}{*}{\rotatebox[origin=c]{90}{\tiny\shortstack{Raw\\observations}}}
 & Sentinel-2 \texttt{s2pc} & Sentinel-2 L2A & 10 & 10\,m & 2017--2025 & surface reflectance & float16 \\
 & Sentinel-1 \texttt{s1rtc} & Sentinel-1 RTC, asc+desc & 4 & 10\,m & 2017--2025 & $\gamma^{0}$ dB & float16 \\
 & Landsat \texttt{landsat} & Landsat C2 L2, L8/L9 & 6 & 30\,m & 2017--2025 & surface reflectance & float16 \\
 & Temperature \texttt{lst} & Landsat C2 L2 thermal & 1 & 30\,m & 2017--2025 & $\circ$C & float16 \\
 & PALSAR \texttt{palsar} & ALOS PALSAR-2 mosaic & 2 & 20\,m & 2017--2021 & $\gamma^{0}$ dB & float16 \\
 \arrayrulecolor{black!35}\cmidrule(lr){2-8}\arrayrulecolor{black}
\multirow{2}{*}{\rotatebox[origin=c]{90}{\tiny\shortstack{Precomputed\\embeddings}}}
 & AlphaEarth \texttt{AE} & AlphaEarth Foundations & 64 & 10\,m & 2017--2025 & embedding & int8 \\
 & TESSERA \texttt{TE} & TESSERA & 128 & 10\,m & 2017--2025 & embedding & int8 \\
\arrayrulecolor{black!35}\cmidrule(lr){2-8}\arrayrulecolor{black}
\multirow{3}{*}{\rotatebox[origin=c]{90}{\tiny\shortstack{Map\\products}}}
 & Pasture \texttt{gpw} & Global Pasture Watch v2 & 3 & 30\,m & 2017--2024 & \% cover & float16 \\
 & Biomass \texttt{agb} & ESA CCI Biomass v7.0 & 2 & 90\,m & 2017--2024 & Mg/ha (AGB, s.d.) & float16 \\
 & Surface water \texttt{water} & GLAD surface water & 1 & 30\,m & 2017--2021 & \% & float16 \\
\arrayrulecolor{black!35}\cmidrule(lr){2-8}\arrayrulecolor{black}
\multirow{4}{*}{\rotatebox[origin=c]{90}{\tiny\shortstack{Static\\maps}}}
 & Elevation \texttt{dem} & Copernicus DEM GLO-30 & 2 & 30\,m & static & m, degrees & float16 \\
 & Canopy \texttt{canopy} & ETH Canopy Height 2020 & 2 & 10\,m & static & m (height, s.d.) & float16 \\
 & Wetland \texttt{glwd} & GLWD v2 wetland & 1 & 464\,m & static & \% & float16 \\
 & Soil \texttt{soilgrids} & SoilGrids v2.0, 0--5\,cm & 6 & 250\,m & static & 6 soil properties & float16 \\
\end{tabular}
\end{table}
\providecolor{pffRowBg}{HTML}{FCF4E9}  
\providecommand{\pffz}[1]{\makebox[0pt][c]{#1}}
\begin{table}[h]
\caption{\textbf{PFFs and most hybrid neural fields saturate performance on
patch-level classification tasks.} We report linear-probe transfer results using stats pooling. All values are percentages of the original
features (gray). We also evaluate traditional geographic location encoders (fully implicit, global) SatCLIP \citep{satclip}, GeoCLIP \citep{geoclip}, and Tessellating the Earth (TTE) \citep{cher2026tte} on classification tasks and report their retention as a percentage of AlphaEarth's original features.}
\label{tab:transfer}
\vspace{2pt}
\small
\renewcommand{\arraystretch}{1.15}
\setlength{\tabcolsep}{2pt}
\begin{tabularx}{\linewidth}{@{}l *{8}{>{\centering\arraybackslash}X}@{}}
\textbf{(a) BEN-v2} & \multicolumn{2}{c}{AlphaEarth} & \multicolumn{2}{c}{TESSERA} & \multicolumn{2}{c}{Sentinel-2} & \multicolumn{2}{c@{}}{Landsat} \\
\cmidrule(lr){2-3}\cmidrule(lr){4-5}\cmidrule(lr){6-7}\cmidrule(l){8-9}
 & \pffz{\mbox{Micro-F1}} & mAP & \pffz{\mbox{Micro-F1}} & mAP & \pffz{\mbox{Micro-F1}} & mAP & \pffz{\mbox{Micro-F1}} & mAP \\
\midrule
\textcolor{black!55}{Original features} & \textcolor{black!55}{74.9} & \textcolor{black!55}{68.8} & \textcolor{black!55}{74.2} & \textcolor{black!55}{68.3} & \textcolor{black!55}{61.5} & \textcolor{black!55}{51.8} & \textcolor{black!55}{58.6} & \textcolor{black!55}{47.3} \\
\addlinespace[2.5pt]
TensoRF  & \underline{100\%} & \underline{101\%} & \textbf{101\%} & \underline{100\%} & \underline{102\%} & \underline{100\%} & \underline{101\%} & \underline{102\%} \\
K-Planes & 100\% & 100\% & 101\% & 98\% & 99\% & 97\% & 98\% & 97\% \\
iNGP     & \textbf{101\%} & \textbf{101\%} & 100\% & 95\% & 99\% & 97\% & 101\% & 101\% \\
\rowcolor{pffRowBg}
\textbf{PFF-VM} & \underline{100\%} & \underline{101\%} & \underline{101\%} & \textbf{101\%} & \textbf{103\%} & \textbf{103\%} & \textbf{103\%} & \textbf{103\%} \\
\addlinespace[2.5pt]
\textcolor{black!55}{\textit{Reference (Location Encoders)}} & \multicolumn{8}{c@{}}{\textcolor{black!55}{SatCLIP\ 74\%\,/\,66\% \qquad GeoCLIP\ 71\%\,/\,63\% \qquad TTE\ 72\%\,/\,63\%}} \\
\end{tabularx}

\vspace{6pt}
\begin{tabularx}{\linewidth}{@{}l *{8}{>{\centering\arraybackslash}X}@{}}
\textbf{(b) So2Sat} & \multicolumn{2}{c}{AlphaEarth} & \multicolumn{2}{c}{TESSERA} & \multicolumn{2}{c}{Sentinel-2} & \multicolumn{2}{c@{}}{Landsat} \\
\cmidrule(lr){2-3}\cmidrule(lr){4-5}\cmidrule(lr){6-7}\cmidrule(l){8-9}
 & OA & $\kappa$ & OA & $\kappa$ & OA & $\kappa$ & OA & $\kappa$ \\
\midrule
\textcolor{black!55}{Original features} & \textcolor{black!55}{63.2} & \textcolor{black!55}{59.6} & \textcolor{black!55}{65.1} & \textcolor{black!55}{61.4} & \textcolor{black!55}{62.4} & \textcolor{black!55}{58.6} & \textcolor{black!55}{55.6} & \textcolor{black!55}{51.1} \\
\addlinespace[2.5pt]
TensoRF  & \underline{98\%} & \underline{98\%} & \underline{97\%} & \underline{97\%} & 88\% & 85\% & \underline{96\%} & \underline{95\%} \\
K-Planes & 97\% & 97\% & 95\% & 95\% & 83\% & 79\% & 92\% & 90\% \\
iNGP     & 90\% & 88\% & 90\% & 89\% & \underline{88\%} & \underline{86\%} & 93\% & 92\% \\
\rowcolor{pffRowBg}
\textbf{PFF-VM} & \textbf{99\%} & \textbf{99\%} & \textbf{99\%} & \textbf{100\%} & \textbf{90\%} & \textbf{88\%} & \textbf{98\%} & \textbf{97\%} \\
\addlinespace[2.5pt]
\textcolor{black!55}{\textit{Reference (Location Encoders)}} & \multicolumn{8}{c@{}}{\textcolor{black!55}{SatCLIP\ 20\%\,/\,4\% \qquad GeoCLIP\ 30\%\,/\,18\% \qquad TTE\ 20\%\,/\,4\%}} \\
\end{tabularx}
\label{tab:classification_tables}
\end{table}

\begingroup
\providecolor{pffMergeLearned}{HTML}{6B5AA8}
\providecolor{pffMergeSignal}{HTML}{35708E}
\providecolor{pffMergeMap}{HTML}{4F8A5C}
\providecolor{pffMergeV}{HTML}{8574C9}
\providecolor{pffMergeT}{HTML}{5FA3A8}
\providecolor{pffMergeM}{HTML}{B06FB2}
\providecolor{pffMergeS}{HTML}{35708E}
\providecolor{pffMergeG}{HTML}{6F9A4D}
\providecolor{pffMergeL}{HTML}{DF6E3D}
\providecolor{pffMergeD}{HTML}{B48C60}
\providecolor{pffMergeE}{HTML}{7E6DBE}

\newcommand{\pffMergeIcon}[1]{%
  \begingroup
  \ifcase#1
    \def\pffMergeCells{pffMergeV!50,pffMergeT!50,pffMergeM!50,pffMergeS!50,pffMergeG!50,pffMergeL!50,pffMergeD!50,pffMergeE!50}%
  \or
    \def\pffMergeCells{pffMergeLearned!48,pffMergeLearned!60,pffMergeLearned!52,pffMergeLearned!68,pffMergeLearned!55,pffMergeLearned!62,pffMergeLearned!48,pffMergeLearned!66}%
  \or
    \def\pffMergeCells{pffMergeSignal!48,pffMergeSignal!60,pffMergeSignal!52,pffMergeSignal!68,pffMergeSignal!55,pffMergeSignal!62,pffMergeSignal!48,pffMergeSignal!66}%
  \or
    \def\pffMergeCells{pffMergeMap!48,pffMergeMap!60,pffMergeMap!52,pffMergeMap!68,pffMergeMap!55,pffMergeMap!62,pffMergeMap!48,pffMergeMap!66}%
  \or
    \def\pffMergeCells{pffMergeLearned!48,pffMergeLearned!60,pffMergeLearned!68,pffMergeSignal!48,pffMergeSignal!60,pffMergeSignal!68,pffMergeMap!48,pffMergeMap!66}%
  \fi
  \begin{tikzpicture}[x=0.72pt,y=0.9pt,baseline=1.0pt,
      line join=round,line width=0.20pt,draw=black!38]
    \foreach \c [count=\i from 0] in \pffMergeCells {%
      \colorlet{pffMergeCell}{\c}%
      \path[fill=pffMergeCell!65!white,draw]
        ({3*\i},6) -- ({3*\i+2.6},8)
        -- ({3*\i+5.6},8) -- ({3*\i+3},6) -- cycle;
      \path[fill=pffMergeCell,draw] ({3*\i},0) rectangle ({3*\i+3},6);
      \ifnum\i=7
        \path[fill=pffMergeCell!78!black,draw]
          (24,0) -- (26.6,2) -- (26.6,8) -- (24,6) -- cycle;
      \fi
    }%
  \end{tikzpicture}%
  \endgroup
}
\newcommand{\pffMergeName}[2]{\pffMergeIcon{#1}\hspace{4pt}#2}
\newcommand{\pffMergeDim}[1]{\textcolor{black!55}{#1}}

\begin{table*}[t]
\centering
\begin{minipage}[c]{0.55\linewidth}
\small
\renewcommand{\arraystretch}{1.18}
\setlength{\tabcolsep}{0pt}
\begin{tabularx}{\linewidth}{@{}>{\raggedright\arraybackslash}X
  @{\hspace{8pt}}>{\centering\arraybackslash}p{20pt}
  @{\hspace{3pt}}>{\centering\arraybackslash}p{20pt}
  @{\hspace{10pt}}>{\centering\arraybackslash}p{28pt}
  @{\hspace{4pt}}>{\centering\arraybackslash}p{28pt}@{}}
\textbf{Representation} & \multicolumn{2}{c@{\hspace{10pt}}}{Dim} & \multicolumn{2}{c@{}}{PASTIS mIoU $\uparrow$} \\
\cmidrule(r{10pt}){2-3}\cmidrule{4-5}
 & VM & K-Pl & VM & K-Pl \\
\midrule
\pffMergeName{0}{Shared feature} & \pffMergeDim{64} & \pffMergeDim{128} & 17.72 & 21.62 \\
\addlinespace[2.5pt]
\pffMergeName{1}{Embedding Stack} & \pffMergeDim{192} & \pffMergeDim{192} & \textbf{55.68} & \underline{55.52} \\
\pffMergeName{2}{Obs Stack} & \pffMergeDim{23} & \pffMergeDim{23} & 30.01 & 26.92 \\
\pffMergeName{3}{Map Stack} & \pffMergeDim{17} & \pffMergeDim{17} & 13.00 & 12.94 \\
\addlinespace[2.5pt]
\pffMergeName{4}{All Stack} & \pffMergeDim{232} & \pffMergeDim{232} & \underline{55.54} & \textbf{55.72} \\
\addlinespace[5pt]
\multicolumn{5}{@{}l}{\textcolor{black!55}{\textit{Single-product controls}}} \\
\addlinespace[1pt]
\textcolor{black!55}{Raw AE} & \textcolor{black!55}{64} & \textcolor{black!55}{64} & \multicolumn{2}{c@{}}{\textcolor{black!55}{53.56}} \\
\textcolor{black!55}{Raw TE} & \textcolor{black!55}{128} & \textcolor{black!55}{128} & \multicolumn{2}{c@{}}{\textcolor{black!55}{61.32}} \\
\textcolor{black!55}{PFF-reconstructed AE} & \textcolor{black!55}{64} & \textcolor{black!55}{64} & \textcolor{black!55}{48.20} & \textcolor{black!55}{47.66} \\
\textcolor{black!55}{PFF-reconstructed TE} & \textcolor{black!55}{128} & \textcolor{black!55}{128} & \textcolor{black!55}{54.58} & \textcolor{black!55}{53.33} \\
\end{tabularx}
\end{minipage}\hfill%
\begin{minipage}[c]{0.41\linewidth}
\setlength{\abovecaptionskip}{0pt}
\setlength{\belowcaptionskip}{0pt}
\caption{\textbf{Probing shared and stacked PFF features.}
PASTIS test mIoU (\%) from a pooled linear probe across 17 regional fields,
using identical evaluation pixels and train-only standardization.
Stack combine observations, embeddings, maps, or all 14 products.
\textbf{Bold} / \underline{underline}: best / second-best merging result per architecture.
Gray rows show single-product controls; raw scores span both architectures.}
\label{tab:pff-feature-merging}
\end{minipage}
\end{table*}
\endgroup

\providecolor{pffBudgetTint}{HTML}{FCF4E9}
\providecolor{pffBudgetOrange}{HTML}{EF8A2C}
\providecolor{pffBudgetMeanGray}{HTML}{EEEEEE}
\begin{table}[t]
\centering
\caption{\textbf{Reconstruction scores ($\uparrow$) on \textsc{PFFBench}.} Shared PFFs are compared with rate-matched single-product neural fields across 100 globally distributed tiles (\Cref{fig:pffbench-selection}). Single-product fields use either uniform field budgets ($1/14$ per product) or bit-proportional budgets. Scores are means $\times100$, with one standard error across tiles beneath each score. Mean (14) averages products within each tile before aggregation. Bold and underline denote the best and second-best scores in each column, including ties.}
\label{tab:bitweighted}
\vspace{3pt}
\begingroup
\color{black}
\fontsize{7.2}{9}\selectfont
\setlength{\tabcolsep}{0.35pt}
\renewcommand{\arraystretch}{1.25}
\setlength{\aboverulesep}{2pt}
\setlength{\belowrulesep}{2pt}
\newcommand{\budgetHeader}[1]{\rotatebox{75}{\textbf{#1}}}
\newcommand{\budgetCell}[2]{\begingroup\setlength{\unitlength}{1pt}\raisebox{-4.5pt}{\begin{picture}(18,15)(-9,-5)\put(0,3){\makebox[0pt][c]{#1}}\put(0,-3){\makebox[0pt][c]{\fontsize{4.8}{5.3}\selectfont\textcolor{black!65}{\textpm\,#2}}}\end{picture}}\endgroup}
\newsavebox{\pffBudgetTableBox}
\sbox{\pffBudgetTableBox}{%
\begin{tabular*}{\linewidth}{@{\extracolsep{\fill}}l l<{\hspace{1pt}} *{15}{c}@{}}
\textbf{Model} & \textbf{Allocation} & \budgetHeader{AlphaEarth} & \budgetHeader{TESSERA} & \budgetHeader{Sentinel-2} & \budgetHeader{Landsat} & \budgetHeader{LST} & \budgetHeader{S1 RTC} & \budgetHeader{PALSAR} & \budgetHeader{Water} & \budgetHeader{Pasture} & \budgetHeader{Biomass} & \budgetHeader{DEM} & \budgetHeader{Canopy} & \budgetHeader{Wetlands} & \budgetHeader{SoilGrids} & \budgetHeader{Mean (14)} \\
\midrule
Solo VM & {\fontsize{6.2}{7}\selectfont Uniform} & \budgetCell{88.8}{0.41} & \budgetCell{56.4}{0.90} & \budgetCell{84.3}{0.62} & \budgetCell{91.9}{0.37} & \budgetCell{\textbf{98.5}}{0.09} & \budgetCell{78.5}{0.95} & \budgetCell{75.5}{1.28} & \budgetCell{\underline{95.1}}{0.46} & \budgetCell{92.0}{0.37} & \budgetCell{\textbf{92.8}}{0.67} & \budgetCell{95.0}{0.38} & \budgetCell{94.3}{0.33} & \budgetCell{\textbf{100.0}}{\textless{}0.01} & \budgetCell{\textbf{99.4}}{0.03} & \budgetCell{88.8}{0.35} \\
 & {\fontsize{6.2}{7}\selectfont Bit-weighted} & \budgetCell{\underline{91.2}}{0.31} & \budgetCell{65.1}{0.81} & \budgetCell{83.0}{0.65} & \budgetCell{88.9}{0.49} & \budgetCell{94.1}{0.31} & \budgetCell{71.0}{1.29} & \budgetCell{67.6}{1.52} & \budgetCell{87.5}{1.13} & \budgetCell{85.6}{0.68} & \budgetCell{83.5}{1.38} & \budgetCell{78.1}{1.10} & \budgetCell{71.2}{1.08} & \budgetCell{94.9}{0.25} & \budgetCell{95.8}{0.21} & \budgetCell{82.7}{0.54} \\
\addlinespace[2pt]
Solo K-Pl. & {\fontsize{6.2}{7}\selectfont Uniform} & \budgetCell{86.9}{0.42} & \budgetCell{61.1}{0.90} & \budgetCell{84.4}{0.61} & \budgetCell{92.0}{0.36} & \budgetCell{\underline{98.4}}{0.09} & \budgetCell{78.1}{0.97} & \budgetCell{75.7}{1.27} & \budgetCell{\textbf{95.5}}{0.42} & \budgetCell{92.1}{0.37} & \budgetCell{\textbf{92.8}}{0.67} & \budgetCell{94.5}{0.41} & \budgetCell{93.9}{0.36} & \budgetCell{\textbf{100.0}}{\textless{}0.01} & \budgetCell{\underline{99.3}}{0.03} & \budgetCell{88.9}{0.36} \\
 & {\fontsize{6.2}{7}\selectfont Bit-weighted} & \budgetCell{\textbf{91.7}}{0.32} & \budgetCell{\textbf{69.1}}{0.79} & \budgetCell{83.0}{0.65} & \budgetCell{88.5}{0.50} & \budgetCell{93.4}{0.35} & \budgetCell{69.4}{1.37} & \budgetCell{65.5}{1.59} & \budgetCell{85.1}{1.23} & \budgetCell{84.3}{0.75} & \budgetCell{82.0}{1.44} & \budgetCell{73.0}{1.18} & \budgetCell{59.1}{1.39} & \budgetCell{87.6}{0.62} & \budgetCell{94.5}{0.27} & \budgetCell{80.4}{0.59} \\
\midrule
\textcolor{pffBudgetOrange}{\textbf{PFF}}-VM & {\fontsize{6.2}{7}\selectfont Shared} & \budgetCell{89.2}{0.50} & \budgetCell{63.9}{0.95} & \budgetCell{\underline{88.4}}{0.65} & \budgetCell{\textbf{93.6}}{0.45} & \budgetCell{94.8}{0.25} & \budgetCell{\underline{81.6}}{0.73} & \budgetCell{\underline{82.9}}{1.14} & \budgetCell{95.0}{0.46} & \budgetCell{\underline{93.0}}{0.26} & \budgetCell{92.6}{0.48} & \budgetCell{\underline{96.3}}{0.22} & \budgetCell{\textbf{97.1}}{0.25} & \budgetCell{\underline{99.9}}{0.01} & \budgetCell{98.9}{0.04} & \budgetCell{\underline{90.5}}{0.36} \\
\textcolor{pffBudgetOrange}{\textbf{PFF}}-K-Pl. & {\fontsize{6.2}{7}\selectfont Shared} & \budgetCell{89.5}{0.42} & \budgetCell{\underline{67.5}}{0.89} & \budgetCell{\textbf{89.0}}{0.63} & \budgetCell{\underline{92.4}}{0.33} & \budgetCell{96.6}{0.16} & \budgetCell{\textbf{83.5}}{0.67} & \budgetCell{\textbf{83.2}}{1.17} & \budgetCell{\textbf{95.5}}{0.41} & \budgetCell{\textbf{93.9}}{0.23} & \budgetCell{\underline{92.7}}{0.52} & \budgetCell{\textbf{96.9}}{0.18} & \budgetCell{\underline{96.1}}{0.22} & \budgetCell{\underline{99.9}}{0.01} & \budgetCell{98.9}{0.04} & \budgetCell{\textbf{91.1}}{0.33} \\
\end{tabular*}}
\newsavebox{\pffBudgetRowsBox}
\sbox{\pffBudgetRowsBox}{\begin{tabular}{ll*{15}{c}}\textcolor{pffBudgetOrange}{\textbf{PFF}}-VM & {\fontsize{6.2}{7}\selectfont Shared} & \budgetCell{89.2}{0.50} & \budgetCell{63.9}{0.95} & \budgetCell{\underline{88.4}}{0.65} & \budgetCell{\textbf{93.6}}{0.45} & \budgetCell{94.8}{0.25} & \budgetCell{\underline{81.6}}{0.73} & \budgetCell{\underline{82.9}}{1.14} & \budgetCell{95.0}{0.46} & \budgetCell{\underline{93.0}}{0.26} & \budgetCell{92.6}{0.48} & \budgetCell{\underline{96.3}}{0.22} & \budgetCell{\textbf{97.1}}{0.25} & \budgetCell{\underline{99.9}}{0.01} & \budgetCell{98.9}{0.04} & \budgetCell{\underline{90.5}}{0.36} \\
\textcolor{pffBudgetOrange}{\textbf{PFF}}-K-Pl. & {\fontsize{6.2}{7}\selectfont Shared} & \budgetCell{89.5}{0.42} & \budgetCell{\underline{67.5}}{0.89} & \budgetCell{\textbf{89.0}}{0.63} & \budgetCell{\underline{92.4}}{0.33} & \budgetCell{96.6}{0.16} & \budgetCell{\textbf{83.5}}{0.67} & \budgetCell{\textbf{83.2}}{1.17} & \budgetCell{\textbf{95.5}}{0.41} & \budgetCell{\textbf{93.9}}{0.23} & \budgetCell{\underline{92.7}}{0.52} & \budgetCell{\textbf{96.9}}{0.18} & \budgetCell{\underline{96.1}}{0.22} & \budgetCell{\underline{99.9}}{0.01} & \budgetCell{98.9}{0.04} & \budgetCell{\textbf{91.1}}{0.33} \\\end{tabular}}
\newsavebox{\pffBudgetSoloRowsBox}
\sbox{\pffBudgetSoloRowsBox}{\begin{tabular}{ll*{15}{c}}Solo VM & {\fontsize{6.2}{7}\selectfont Uniform} & \budgetCell{88.8}{0.41} & \budgetCell{56.4}{0.90} & \budgetCell{84.3}{0.62} & \budgetCell{91.9}{0.37} & \budgetCell{\textbf{98.5}}{0.09} & \budgetCell{78.5}{0.95} & \budgetCell{75.5}{1.28} & \budgetCell{\underline{95.1}}{0.46} & \budgetCell{92.0}{0.37} & \budgetCell{\textbf{92.8}}{0.67} & \budgetCell{95.0}{0.38} & \budgetCell{94.3}{0.33} & \budgetCell{\textbf{100.0}}{\textless{}0.01} & \budgetCell{\textbf{99.4}}{0.03} & \budgetCell{88.8}{0.35} \\
 & {\fontsize{6.2}{7}\selectfont Bit-weighted} & \budgetCell{\underline{91.2}}{0.31} & \budgetCell{65.1}{0.81} & \budgetCell{83.0}{0.65} & \budgetCell{88.9}{0.49} & \budgetCell{94.1}{0.31} & \budgetCell{71.0}{1.29} & \budgetCell{67.6}{1.52} & \budgetCell{87.5}{1.13} & \budgetCell{85.6}{0.68} & \budgetCell{83.5}{1.38} & \budgetCell{78.1}{1.10} & \budgetCell{71.2}{1.08} & \budgetCell{94.9}{0.25} & \budgetCell{95.8}{0.21} & \budgetCell{82.7}{0.54} \\
\addlinespace[2pt]
Solo K-Pl. & {\fontsize{6.2}{7}\selectfont Uniform} & \budgetCell{86.9}{0.42} & \budgetCell{61.1}{0.90} & \budgetCell{84.4}{0.61} & \budgetCell{92.0}{0.36} & \budgetCell{\underline{98.4}}{0.09} & \budgetCell{78.1}{0.97} & \budgetCell{75.7}{1.27} & \budgetCell{\textbf{95.5}}{0.42} & \budgetCell{92.1}{0.37} & \budgetCell{\textbf{92.8}}{0.67} & \budgetCell{94.5}{0.41} & \budgetCell{93.9}{0.36} & \budgetCell{\textbf{100.0}}{\textless{}0.01} & \budgetCell{\underline{99.3}}{0.03} & \budgetCell{88.9}{0.36} \\
 & {\fontsize{6.2}{7}\selectfont Bit-weighted} & \budgetCell{\textbf{91.7}}{0.32} & \budgetCell{\textbf{69.1}}{0.79} & \budgetCell{83.0}{0.65} & \budgetCell{88.5}{0.50} & \budgetCell{93.4}{0.35} & \budgetCell{69.4}{1.37} & \budgetCell{65.5}{1.59} & \budgetCell{85.1}{1.23} & \budgetCell{84.3}{0.75} & \budgetCell{82.0}{1.44} & \budgetCell{73.0}{1.18} & \budgetCell{59.1}{1.39} & \budgetCell{87.6}{0.62} & \budgetCell{94.5}{0.27} & \budgetCell{80.4}{0.59} \\\end{tabular}}
\newsavebox{\pffBudgetMeanCellBox}
\sbox{\pffBudgetMeanCellBox}{\budgetCell{88.9}{0.36}}
\begin{tikzpicture}
\path[use as bounding box] (0,0) rectangle (\linewidth,\ht\pffBudgetTableBox+\dp\pffBudgetTableBox);
\pgfmathsetlengthmacro{\pffBudgetGrayBottom}{\ht\pffBudgetRowsBox+\dp\pffBudgetRowsBox+\aboverulesep+\belowrulesep+\lightrulewidth}
\fill[pffBudgetMeanGray] (\linewidth-\wd\pffBudgetMeanCellBox-2pt,\pffBudgetGrayBottom-\aboverulesep) rectangle (\linewidth,\pffBudgetGrayBottom+\ht\pffBudgetSoloRowsBox+\dp\pffBudgetSoloRowsBox+\belowrulesep);
\fill[pffBudgetTint] (0,0) rectangle (\linewidth,\ht\pffBudgetRowsBox+\dp\pffBudgetRowsBox);
\node[anchor=south west,inner sep=0pt,outer sep=0pt] at (0,0) {\usebox{\pffBudgetTableBox}};
\end{tikzpicture}
\endgroup
\vspace{-5pt}
\end{table}

\providecolor{pffRowBg}{HTML}{FCF4E9}
\providecommand{\pffz}[1]{\makebox[0pt][c]{#1}}
\providecommand{\pc}{{\scriptsize\%}}
\begin{table}[t]
\caption{\textbf{EuroMineNet footprint mapping.}
Gray rows show original-feature scores (\(\times100\)); other rows show retention relative to those scores.
B-F1 measures mining-class F1 within three pixels of a ground-truth boundary; OF1 measures mining-class F1 over reconstructed site scenes.}
\label{tab:emn-footprint-retention}
\vspace{2pt}
\small
\renewcommand{\arraystretch}{1.08}
\setlength{\tabcolsep}{2pt}
\begin{tabularx}{\linewidth}{@{}l *{10}{>{\centering\arraybackslash}X}@{}}
 & \multicolumn{2}{c}{AlphaEarth} & \multicolumn{2}{c}{TESSERA} & \multicolumn{2}{c}{Sentinel-2} & \multicolumn{2}{c}{Landsat} & \multicolumn{2}{c@{}}{Sentinel-1} \\
\cmidrule(lr){2-3}\cmidrule(lr){4-5}\cmidrule(lr){6-7}\cmidrule(lr){8-9}\cmidrule(l){10-11}
 & B-F1 & OF1 & B-F1 & OF1 & B-F1 & OF1 & B-F1 & OF1 & B-F1 & OF1 \\
\midrule
\textcolor{black!55}{Original features} & \textcolor{black!55}{66.5} & \textcolor{black!55}{78.0} & \textcolor{black!55}{67.4} & \textcolor{black!55}{77.3} & \textcolor{black!55}{63.9} & \textcolor{black!55}{70.7} & \textcolor{black!55}{56.4} & \textcolor{black!55}{64.8} & \textcolor{black!55}{49.6} & \textcolor{black!55}{45.7} \\
\addlinespace[2.5pt]
TensoRF(VM) & \underline{97\pc} & \pffz{\textbf{100\pc}} & 94\pc & \pffz{\textbf{103\pc}} & 85\pc & \pffz{\underline{104\pc}} & 89\pc & \pffz{\underline{103\pc}} & 91\pc & \pffz{104\pc} \\
K-Planes & \underline{97\pc} & \pffz{\textbf{100\pc}} & 94\pc & \pffz{\textbf{103\pc}} & 85\pc & \pffz{\textbf{105\pc}} & 90\pc & \pffz{\underline{103\pc}} & 92\pc & \pffz{\underline{105\pc}} \\
iNGP & 88\pc & \pffz{98\pc} & 83\pc & \pffz{100\pc} & 82\pc & \pffz{102\pc} & 86\pc & \pffz{101\pc} & 85\pc & \pffz{100\pc} \\
\rowcolor{pffRowBg}
\textbf{PFF-VM} & \textbf{99\pc} & \pffz{\textbf{100\pc}} & \underline{98\pc} & \pffz{\underline{102\pc}} & \underline{91\pc} & \pffz{\underline{104\pc}} & \underline{97\pc} & \pffz{\textbf{104\pc}} & \underline{98\pc} & \pffz{\textbf{106\pc}} \\
\rowcolor{pffRowBg}
\textbf{PFF-K-Planes} & \textbf{99\pc} & \pffz{\underline{99\pc}} & \textbf{99\pc} & \pffz{\underline{102\pc}} & \textbf{93\pc} & \pffz{\underline{104\pc}} & \textbf{100\pc} & \pffz{\textbf{104\pc}} & \textbf{100\pc} & \pffz{\underline{105\pc}} \\
\end{tabularx}
\label{tab:eurominenet_footprint_mapping}
\end{table}

\section{Additional Latency Experiments}
\label{sec:materialization_cost}

PFF replaces stored EO feature products with local neural fields, so a query that previously read pixels is now an inference operation. 
We measure the scalability of this effort in practice on retrieval of AlphaEarth and TESSERA embeddings \citep{alphaearth,tessera}. 
PFF-materialized features can be used for large-scale model training where data loading is a bottleneck, or for large-scale mapping efforts. 
We measure the time to materialize one million embeddings relative to public retrieval methods (\Cref{fig:latency}), the effect of memory capacity on serving multiple PFFs (\Cref{tab:e2-fleet}), and resident-model throughput across hardware (\Cref{fig:e3-throughput}).

\paragraph{Experimental setup.} \input{figures/appendix/latency_e0_map}
We evaluate feature materialization speed using AlphaEarth  \citep{alphaearth} and TESSERA \citep{tessera} embeddings. 
For each product, we construct two frozen traces of \(10^6\) coordinate--year queries within valid source coverage and use identical queries across methods. 
AlphaEarth and TESSERA queries span 100 regions and the years 2017--2025.
We compare AlphaEarth retrieval through the Google Earth Engine API \citep{eeapi}, HTTP range reads of source Cloud-Optimized GeoTIFFs (COGs), and complete COG downloads. 
The API uses batched \texttt{reduceRegions} requests, range reads retrieve and decode the required raster blocks, and complete downloads gather the requested pixels from local files.
For TESSERA, we compare complete native-tile downloads with batched queries through GeoTessera's Zarr  reader, both accessing Source Cooperative \citep{source_cooperative}. 
Retrieval clients use concurrent requests and bounded working sets to limit memory consumption. 
PFFs use compiled ONNX graphs with fp16 weights and fp32 computation, with a separate inference session for each region.
All retrieval measurements run on the same Google Cloud virtual machine in Oregon (\texttt{us-west1-a}), with eight vCPUs, 32\,GB of host memory, and an NVIDIA L4. 
Each run starts with empty local data and model caches and ends when all requested embeddings are available as ordered float32 vectors in memory (GPU or CPU). 
Timings include initialization, data or checkpoint acquisition, decoding or model loading, inference, and required memory transfers; training and ONNX export are excluded. 
GPU PFF inference writes directly into the output tensor, while CPU methods include the final transfer to GPU. 
We report mean latency over the two traces in \Cref{fig:latency}, retaining overlapping download and decoding work as a combined measured stage. Remote caches are not controlled, and API response time includes service computation and transfer. 
We separately evaluate repeated queries across the 100-region AlphaEarth PFF collection with checkpoints already on local disk (\Cref{tab:e2-fleet}), and resident-model throughput across hardware with checkpoint loading excluded (\Cref{fig:e3-throughput}). 

\providecolor{pffRowBg}{HTML}{FCF4E9}
\begin{table}[t]
\caption{\textbf{Serving PFFs on hardware-constrained environments.}  We benchmark serving $10^6$ AlphaEarth embeddings with PFFs spread over approximately 100 fully valid regions globally across a variety of hardware environments. PFFs are faster to materialize compared to the Earth Engine API, even when run on a 2-core, 8 GB RAM CPU machine, averaging 6k embeddings/second compared to max API throughput of $\approx$1800 embeddings/s. }
\vspace{-10pt}
\label{tab:e2-fleet}
\vspace{2pt}
\small
\renewcommand{\arraystretch}{1.15}
\setlength{\tabcolsep}{2pt}
\begin{tabular*}{\linewidth}{@{\extracolsep{\fill}}lrrrr@{}}
 & & \multicolumn{2}{c}{\shortstack[c]{Seconds to answer one query\\[-1.5pt]
     {\scriptsize\textcolor{black!55}{A query $=$ one request for $10^{6}$ embeddings}}}} & \\
\cmidrule(lr){3-4}
\multicolumn{1}{@{}l}{Hardware}
 & \multicolumn{1}{r}{\shortstack[r]{PFFs held\\in memory}}
 & \multicolumn{1}{r}{\shortstack[r]{First query\\
     {\scriptsize\textcolor{black!55}{Loads all 100 PFFs}}}}
 & \multicolumn{1}{r}{\shortstack[r]{Each later query\\
     {\scriptsize\textcolor{black!55}{Reloads what did not fit}}}}
 & \multicolumn{1}{r@{}}{\shortstack[r]{Throughput\\(emb\,/\,s)}} \\
\midrule
Cascade Lake 2.8\,GHz $\cdot$ 2 cores $\cdot$ 8\,GB
 & 13\,/\,100 & 178.1 & 167.5 & 6.0\,k \\
Sapphire Rapids $\cdot$ 8 cores $\cdot$ 31\,GB
 & 60\,/\,100 & 57.4 & 32.7 & 30.6\,k \\
EPYC 7763 $\cdot$ 32 threads $\cdot$ 503\,GB
 & 100\,/\,100 & 26.5 & 14.5 & 69.1\,k \\
\rowcolor{pffRowBg}
NVIDIA L4 GPU $\cdot$ 24\,GB
 & 100\,/\,100 & 79.5 & \textbf{0.91} & \textbf{1.10\,M} \\
\end{tabular*}

\end{table}

\paragraph{Serving PFFs globally in compute-scarce environments.} \providecolor{aPFF}{HTML}{FF9100}
\providecolor{aPFFlight}{HTML}{FFB74D}
\providecolor{mAPI}{HTML}{1976D2}
\providecolor{ink}{HTML}{1a1a1a}
\providecolor{gridc}{HTML}{bdbdbd}
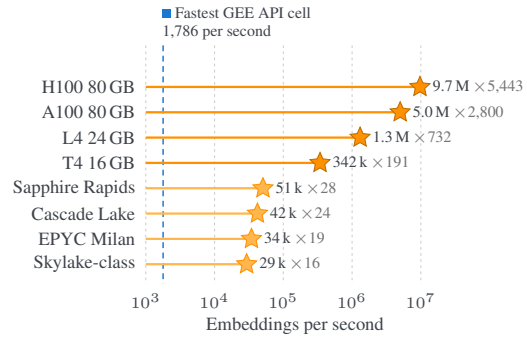
\begin{wrapfigure}{r}{0.52\linewidth}
  \vspace{-10pt}
  \centering
\begin{tikzpicture}
\draw[gridc, dash pattern=on 1pt off 1.3pt, line width=0.3pt, opacity=0.8] (1.800,0.55) -- (1.800,-2.605);
\node[font=\tiny, text=ink, anchor=north] at (1.800,-2.605) {$10^{3}$};
\draw[gridc, dash pattern=on 1pt off 1.3pt, line width=0.3pt, opacity=0.8] (2.710,0.55) -- (2.710,-2.605);
\node[font=\tiny, text=ink, anchor=north] at (2.710,-2.605) {$10^{4}$};
\draw[gridc, dash pattern=on 1pt off 1.3pt, line width=0.3pt, opacity=0.8] (3.619,0.55) -- (3.619,-2.605);
\node[font=\tiny, text=ink, anchor=north] at (3.619,-2.605) {$10^{5}$};
\draw[gridc, dash pattern=on 1pt off 1.3pt, line width=0.3pt, opacity=0.8] (4.529,0.55) -- (4.529,-2.605);
\node[font=\tiny, text=ink, anchor=north] at (4.529,-2.605) {$10^{6}$};
\draw[gridc, dash pattern=on 1pt off 1.3pt, line width=0.3pt, opacity=0.8] (5.439,0.55) -- (5.439,-2.605);
\node[font=\tiny, text=ink, anchor=north] at (5.439,-2.605) {$10^{7}$};
\node[font=\scriptsize, text=ink, anchor=north] at (3.775,-2.925) {Embeddings per second};
\draw[mAPI, dash pattern=on 2.2pt off 1.7pt, line width=0.55pt] (2.029,0.55) -- (2.029,-2.605);
\node[font=\tiny, text=ink, anchor=south west, inner sep=1pt, align=left] at (1.979,0.58) {\tikz\fill[mAPI] (0,0) rectangle (0.55em,0.55em);\ Fastest GEE API cell\\1{,}786 per second};
\node[font=\scriptsize, text=ink, anchor=east, inner sep=1.5pt] at (1.720,0.000) {H100 80\,GB};
\draw[aPFF, line width=0.85pt] (1.800,0.000) -- (5.427,0.000);
\node[star, star points=5, star point ratio=2.1, inner sep=0pt, minimum size=1.75ex, fill=aPFF, draw=aPFF!75!black] at (5.427,0.000) {};
\node[font=\tiny, anchor=west, inner sep=1.5pt, text=ink] at (5.537,0.000) {9.7\,M\,{\color{black!55}$\times$5,443}};
\node[font=\scriptsize, text=ink, anchor=east, inner sep=1.5pt] at (1.720,-0.335) {A100 80\,GB};
\draw[aPFF, line width=0.85pt] (1.800,-0.335) -- (5.165,-0.335);
\node[star, star points=5, star point ratio=2.1, inner sep=0pt, minimum size=1.75ex, fill=aPFF, draw=aPFF!75!black] at (5.165,-0.335) {};
\node[font=\tiny, anchor=west, inner sep=1.5pt, text=ink] at (5.275,-0.335) {5.0\,M\,{\color{black!55}$\times$2,800}};
\node[font=\scriptsize, text=ink, anchor=east, inner sep=1.5pt] at (1.720,-0.670) {L4 24\,GB};
\draw[aPFF, line width=0.85pt] (1.800,-0.670) -- (4.635,-0.670);
\node[star, star points=5, star point ratio=2.1, inner sep=0pt, minimum size=1.75ex, fill=aPFF, draw=aPFF!75!black] at (4.635,-0.670) {};
\node[font=\tiny, anchor=west, inner sep=1.5pt, text=ink] at (4.745,-0.670) {1.3\,M\,{\color{black!55}$\times$732}};
\node[font=\scriptsize, text=ink, anchor=east, inner sep=1.5pt] at (1.720,-1.005) {T4 16\,GB};
\draw[aPFF, line width=0.85pt] (1.800,-1.005) -- (4.105,-1.005);
\node[star, star points=5, star point ratio=2.1, inner sep=0pt, minimum size=1.75ex, fill=aPFF, draw=aPFF!75!black] at (4.105,-1.005) {};
\node[font=\tiny, anchor=west, inner sep=1.5pt, text=ink] at (4.215,-1.005) {342\,k\,{\color{black!55}$\times$191}};
\node[font=\scriptsize, text=ink, anchor=east, inner sep=1.5pt] at (1.720,-1.340) {Sapphire Rapids};
\draw[aPFFlight, line width=0.85pt] (1.800,-1.340) -- (3.350,-1.340);
\node[star, star points=5, star point ratio=2.1, inner sep=0pt, minimum size=1.75ex, fill=aPFFlight, draw=aPFF] at (3.350,-1.340) {};
\node[font=\tiny, anchor=west, inner sep=1.5pt, text=ink] at (3.460,-1.340) {51\,k\,{\color{black!55}$\times$28}};
\node[font=\scriptsize, text=ink, anchor=east, inner sep=1.5pt] at (1.720,-1.675) {Cascade Lake};
\draw[aPFFlight, line width=0.85pt] (1.800,-1.675) -- (3.277,-1.675);
\node[star, star points=5, star point ratio=2.1, inner sep=0pt, minimum size=1.75ex, fill=aPFFlight, draw=aPFF] at (3.277,-1.675) {};
\node[font=\tiny, anchor=west, inner sep=1.5pt, text=ink] at (3.387,-1.675) {42\,k\,{\color{black!55}$\times$24}};
\node[font=\scriptsize, text=ink, anchor=east, inner sep=1.5pt] at (1.720,-2.010) {EPYC Milan};
\draw[aPFFlight, line width=0.85pt] (1.800,-2.010) -- (3.198,-2.010);
\node[star, star points=5, star point ratio=2.1, inner sep=0pt, minimum size=1.75ex, fill=aPFFlight, draw=aPFF] at (3.198,-2.010) {};
\node[font=\tiny, anchor=west, inner sep=1.5pt, text=ink] at (3.308,-2.010) {34\,k\,{\color{black!55}$\times$19}};
\node[font=\scriptsize, text=ink, anchor=east, inner sep=1.5pt] at (1.720,-2.345) {Skylake-class};
\draw[aPFFlight, line width=0.85pt] (1.800,-2.345) -- (3.134,-2.345);
\node[star, star points=5, star point ratio=2.1, inner sep=0pt, minimum size=1.75ex, fill=aPFFlight, draw=aPFF] at (3.134,-2.345) {};
\node[font=\tiny, anchor=west, inner sep=1.5pt, text=ink] at (3.244,-2.345) {29\,k\,{\color{black!55}$\times$16}};
\end{tikzpicture}
  \vspace{-2pt}
  \caption{\textbf{PFF max throughput by hardware.} CPUs use 8 threads. Multiplier comparisons are made against the fastest Google Earth Engine API throughput recorded on a virtual machine in \texttt{us-west1-a}.}
  \label{fig:e3-throughput}
\end{wrapfigure}
\Cref{tab:e2-fleet} evaluates repeated requests for \(10^6\) AlphaEarth embeddings across 100 regions on four hardware configurations: a Google Cloud virtual machine with two Intel Xeon vCPUs and 8\,GB RAM, a \texttt{c3-standard-8} VM with eight Sapphire Rapids vCPUs and 31\,GB RAM, an HPC node with AMD EPYC 7763 processors and 503\,GB RAM using 32 inference threads, and an NVIDIA L4 with 24\,GB GPU memory. 
Each region has its own PFF, so answering a query requires evaluating 100 models. 
Checkpoints are already on local disk, but models must be loaded into memory before they can run. 
Keeping them in memory avoids this loading cost on subsequent queries. 
The 2-core, 8\,GB machine retains 13 PFFs and reloads the other 87 for each query, taking \(167.5\)\,s to produce one million embeddings. 
The 31\,GB machine retains 60 PFFs and completes subsequent queries in \(32.7\)\,s. 
Both the EPYC machine and the L4 GPU machine retain all 100 models, eliminating repeated model loading; they answer subsequent queries in \(14.5\)\,s and \(0.91\)\,s, respectively. 
The L4's first query takes \(79.5\)\,s because it also loads the models, but subsequent queries sustain \(1.10\) million embeddings per second. 
Thus, even a small CPU machine can serve the full collection by loading models as needed, while sufficient memory and faster computation substantially reduce the cost of repeated access.

\paragraph{Peak throughput by hardware.}

\Cref{fig:e3-throughput} measures how quickly a PFF can produce AlphaEarth embeddings once the model is loaded. 
We evaluate the same regional PFF on four CPU platforms, each using eight inference threads, and on NVIDIA T4 (16\,GB), L4 (24\,GB), A100 (80\,GB), and H100 (80\,GB) GPUs. 
The CPU platforms include Intel Xeon processors at 2.30 and 2.80\,GHz, an Intel Xeon Platinum 8481C, and an AMD EPYC 7B13. 
We repeatedly evaluate a precomputed set of 8.4 million spatial coordinates for 2020, test several batch sizes, and measure sustained throughput at the fastest tested batch size on each machine. 
Model loading and coordinate preparation are excluded. All reported runs return embeddings to CPU memory, so GPU timings include transferring the results back to the host. CPU throughput ranges from \(29\)k to \(51\)k embeddings per second. 
The T4 produces \(342\)k embeddings per second, while the L4, A100, and H100 reach \(1.3\), \(5\), and \(9.70\) million, respectively. 
The L4 explicitly enables TF32, and the A100 and H100 use the default CUDA settings documented as enabling TF32; the T4 and CPU runs use fp32 computation.
These measurements significantly outperform current inference pipelines in EO. At max throughput on an NVIDIA H100, PFFs can materialize AlphaEarth features at 1.03 milliseconds per squared kilometer, assuming parallel compute in a setting similar to recent efforts to generate large maps \citep{ai2_olmoearth_platform_2026}.

\input{figures/appendix/pffbench_sampling}
\section{PFFBench}
\label{sec:pffbench}

\textsc{PFFBench} contains 1,000 global tiles. Figures 10 and 11 visualize this full benchmark. Our main reconstruction evaluation in  \Cref{fig:rd-fairshare-ladder} uses a 100-tile subset sampled from \textsc{PFFBench}.
Selection balances multi-year product availability, geographic diversity, and variation in feature values.
Of 33,385 AlphaEarth footprints \citep{alphaearth}, 21,370 have all nine years from 2017--2025, at least $50\%$ land-cell coverage, at least $60\%$ of the nominal tile extent estimated from footprint bounds, and an assigned climate class. We additionally require at least $50\%$ valid pixels in the 2019 overview used for scoring. 
TESSERA availability is checked against the v1.1 dClimate registry snapshot of 17 September 2026 \citep{tessera}.\footnote{\url{https://registry.opendata.aws/tessera/}} 
For each year from 2018--2025, \begin{wrapfigure}{r}{2.45in}
\centering
\definecolor{heroInk}{HTML}{26313A}
\definecolor{heroMuted}{HTML}{737B81}
\definecolor{heroRule}{HTML}{BCC3C8}
\definecolor{heroSelected}{HTML}{267E8D}
\definecolor{heroRandom}{HTML}{3F7FBF}
\providecolor{pffRowBg}{HTML}{F7DFC2}
\pgfplotsset{compat=1.18}
\begin{tikzpicture}[x=1in,y=1in,text=heroInk,font=\fontsize{8}{9}\selectfont]
\path[use as bounding box] (0,0) rectangle (2.45,1.44);
\begin{axis}[
  at={(.32in,.34in)},anchor=south west,width=2.05in,height=1.02in,scale only axis,
  xmin=.9,xmax=3.6,ymin=0,ymax=4.0,
  axis lines*=left,axis line style={heroRule,line width=.4pt},
  tick style={heroRule,line width=.4pt},tick align=outside,
  xtick={1.0,1.5,2.0,2.5,3.0,3.5},ytick={0,2,4},
  tick label style={font=\fontsize{7.5}{9}\selectfont,text=heroInk},
  xlabel={Spectral entropy (nats)},ylabel={Density},
  label style={font=\fontsize{8}{9}\selectfont},
  xlabel style={at={(axis description cs:.5,-.16)},anchor=north},
  ylabel style={at={(axis description cs:-.06,.5)},anchor=south},
  legend style={at={(.025,.98)},anchor=north west,draw=none,fill=none,
    font=\fontsize{7.5}{9}\selectfont,cells={anchor=west},inner sep=0pt,row sep=1pt},
  clip=true,
]
  \addplot[const plot,draw=heroInk!70,line width=.8pt,fill=pffRowBg,no marks,area legend]
    table[x=H,y=picked] {figures/appendix/pffbench_sampling_artifacts/hero_hist.dat} \closedcycle;
  \addlegendentry{\textsc{PFFBench}}
  \addplot[const plot,heroRandom,line width=.8pt,densely dashed,no marks]
    table[x=H,y=random] {figures/appendix/pffbench_sampling_artifacts/hero_hist.dat};
  \addlegendentry{Stratified random}
\end{axis}
\end{tikzpicture}
\caption{Entropy ranking shifts the distribution toward higher scores. Mean $H$ is 3.253 nats for \textsc{PFFBench} and 2.804 for constrained random selection (Cohen's $d=1.15$).}
\label{fig:pffbench-entropy}
\vspace{-10pt}
\end{wrapfigure} at least $95\%$ of registered blocks whose bounding boxes intersect the footprint within its UTM zone must be marked as embedded. 
Coverage in 2017 is optional. These filters retain 20,964 tiles. 

We group candidates by continent and K\"oppen--Geiger climate group \citep{naturalearth,peel2007updated}. 
Each group receives a quota proportional to the square root of its candidate count, balancing coverage of smaller groups against the availability of tiles in larger groups. 
Quotas are rounded by largest remainder and capped at the available count; the two singleton groups receive no quota. 
Selected tile centers must be at least $100\,\mathrm{km}$ apart. These constraints spread the benchmark across the Earth without letting the largest groups dominate (\Cref{fig:pffbench-selection}).

Within each group, we prioritize tiles with greater variation in their AlphaEarth values. 
We score the  valid pixels of each tile's 2019 $80\,\mathrm{m}$ overview using mean per-band histogram entropy, $H_i=-\frac{1}{64}\sum_{b=1}^{64}\sum_{k=1}^{100}p_{i,b,k}\log p_{i,b,k}$, where $p_{i,b,k}$ is the fraction of valid pixels in bin $k$ of band $b$. 
We use 100 equal-width bins over the stored int8 range $[-128,128)$. This score provides a low-cost measure of feature-value diversity without fitting a neural field. 
We visit groups in round-robin order, selecting the highest-scoring candidate that satisfies the distance constraint until each quota is met or no eligible candidates remain. 
We fill the remaining places with the highest-scoring eligible tiles across the pool. 

We compare the selected set with 20 random selections using the same quotas and distance constraint. 
Mean entropy increases from $2.8\pm0.005$ across these random selections to $3.2$ for \textsc{PFFBench} (\Cref{fig:pffbench-entropy}). 
Across the candidate pool, entropy also correlates with lossless zlib bytes per valid pixel (Spearman $\rho=0.90$), which is a rough proxy for neural compression reconstruction difficulty. 
\textsc{PFFBench} therefore emphasizes higher-entropy content rather than estimating average reconstruction performance over global land. 

\begin{figure}[t!]
\centering

\usetikzlibrary{arrows.meta,calc,patterns}
\definecolor{bsInk}{HTML}{3E434A}
\definecolor{bsFrame}{HTML}{BFC3CA}
\definecolor{bsPurple}{HTML}{8574C9}
\definecolor{bsPurpleD}{HTML}{46397E}
\definecolor{bsBlue}{HTML}{35708E}
\definecolor{bsGreen}{HTML}{659A4C}
\definecolor{bsReplace}{HTML}{4CAF50} 
\definecolor{bsRemove}{HTML}{F44336} 
\definecolor{bsOrange}{HTML}{EF8A2C}
\definecolor{bsOrangeL}{HTML}{FFAB40}
\definecolor{bsDecoder}{HTML}{4A7FA5}
\definecolor{bsDecoderFill}{HTML}{DEEBF3}
\definecolor{bsLoss}{HTML}{A569BD}
\tikzset{
  bs/flow/.style={draw=bsInk,line width=.65pt,-{Stealth[length=2.7pt,width=2.5pt]},line cap=round,line join=round},
  bs/guide/.style={draw=bsFrame,line width=.35pt,dash pattern=on 1.4pt off 1.4pt},
  bs/math/.style={font=\fontsize{7.3}{8.5}\selectfont,inner sep=1pt,text=bsInk},
  bs/small/.style={font=\fontsize{6.4}{7.5}\selectfont,inner sep=1pt,text=bsInk},
  bs/title/.style={font=\fontsize{7.5}{9}\selectfont\rmfamily\bfseries,inner sep=1pt,text=black},
  bs/subtitle/.style={font=\fontsize{8}{9.5}\selectfont\rmfamily\bfseries,inner sep=1pt,text=black},
  bs/note/.style={font=\fontsize{6.2}{7.4}\selectfont\rmfamily,inner sep=1pt,text=bsInk,align=center},
  bs/missing/.style={fill=white,draw=bsFrame,line width=.45pt,dash pattern=on 1.3pt off 1.0pt},
  bs/gradient/.style={draw=bsOrange,line width=.75pt,dash pattern=on 2.5pt off 1.6pt,line cap=round,line join=round,-{Stealth[length=3pt,width=2.8pt]}},
  bs/edge/.style={line width=.45pt,line join=round},
}

\newcommand{\bsMap}[4]{
  \path[fill=#1!12,draw=#1!60,bs/edge] (0,0) rectangle (1,1);
  \begin{scope}
    \clip (0,0) rectangle (1,1);
    \ifnum#2=1
      \path[fill=#1!38] (-.1,.04) .. controls (.28,.31) and (.21,.55) .. (.54,.62)
        .. controls (.73,.66) and (.86,.91) .. (1.1,1.03) -- (1.1,.3)
        .. controls (.64,.22) and (.75,.06) .. (.48,-.1) -- cycle;
      \path[fill=#1!64] (.03,.62) .. controls (.22,.89) and (.15,1.1) .. (.48,1.06)
        .. controls (.7,.98) and (.63,.73) .. (.46,.65)
        .. controls (.32,.42) and (.09,.4) .. cycle;
      \path[fill=bsPurpleD!58] (.61,.18) .. controls (.62,.39) and (.97,.5) .. (1.08,.26)
        -- (1.1,-.1) -- (.66,-.1) -- cycle;
      \draw[white,opacity=.7,line width=.5pt] (-.04,.35) .. controls (.47,.26) and (.41,.62) .. (.75,.76)
        .. controls (.92,.84) and (.85,.95) .. (.96,1.05);
    \fi
    \ifnum#2=2
      \path[fill=#1!24] (-.1,.1) .. controls (.41,.52) and (.5,.35) .. (1.1,.9)
         -- (1.1,1.1) -- (-.1,1.1) -- cycle;
      \foreach \v in {-.15,.04,.23,.42,.61,.80, .99}{
        \draw[#1!65,line width=.65pt] (-.1,\v) .. controls (.30,{\v+.30}) and (.55,{\v-.16}) .. (1.1,{\v+.23});}
      \foreach \xx/\yy in {.13/.16,.72/.21,.46/.80,.82/.76,.32/.42,.89/.49}{
        \fill[#1!80] (\xx,\yy) circle[radius=.014];}
    \fi
    \ifnum#2=3
      \path[fill=#1!38,draw=#1!58,line width=.3pt] (-.1,.2) -- (.32,.14) -- (.43,.46) -- (.17,.72) -- (-.1,.56) -- cycle;
      \path[fill=#1!65,draw=#1!70,line width=.3pt] (.48,.32) -- (.86,.22) -- (1.1,.48) -- (.9,.87) -- (.61,.91) -- (.39,.65) -- cycle;
      \path[fill=#1!28,draw=#1!60,line width=.3pt] (.1,.8) -- (.42,.72) -- (.6,1.1) -- (.1,1.1) -- cycle;
      \draw[white,line width=.7pt] (.1,-.1) .. controls (.69,.36) and (.23,.64) .. (.7,1.1);
    \fi
    \foreach \u in {.25,.5,.75}{
      \draw[white,opacity=.82,line width=.28pt] (\u,0)--(\u,1) (0,\u)--(1,\u);}
  \end{scope}
  \draw[#1!70,bs/edge] (0,0) rectangle (1,1);
  \draw[white,line width=1.5pt] ({#3/4},{#4/4}) rectangle ({(#3+1)/4},{(#4+1)/4});
  \draw[bsInk,line width=.75pt] ({#3/4},{#4/4}) rectangle ({(#3+1)/4},{(#4+1)/4});
}

\newcommand{\bsBundle}[7]{%
  \begin{scope}[shift={(#1,#2)},scale=#6,opacity=#7]
    \foreach \bsC/\bsKind/\bsH in {bsGreen/3/0,bsBlue/2/.145,bsPurple/1/.29}{
      \begin{scope}[shift={(0,\bsH)}]
        \path[fill=\bsC!28,draw=\bsC!55,line width=.35pt]
          (0,0)--(.82,-.15)--(.82,-.195)--(0,-.045)--cycle;
        \path[fill=\bsC!42,draw=\bsC!55,line width=.35pt]
          (.82,-.15)--(1.18,.21)--(1.18,.165)--(.82,-.195)--cycle;
        \begin{scope}[cm={.82,-.15,.36,.36,(0,0)}]
          \begin{scope}
            \clip (0,0) rectangle (1,1);
            \begin{scope}[scale=4,shift={({-#4/4},{-#5/4})}]
              \bsMap{\bsC}{\bsKind}{#4}{#5}
            \end{scope}
          \end{scope}
          \path[draw=\bsC!75,line width=.4pt,fill=\bsC!8,fill opacity=.22]
            (0,0) rectangle (1,1);
          \foreach \bsU in {.25,.5,.75}{
            \draw[white,opacity=.70,line width=.25pt] (\bsU,0)--(\bsU,1) (0,\bsU)--(1,\bsU);}
          \ifnum#3=2
            \ifnum\bsKind=2
              \path[bs/missing] (0,0) rectangle (1,1);
            \fi
          \fi
        \end{scope}
      \end{scope}}
  \end{scope}}

\newcommand{\bsPin}[7]{%
  \pgfmathsetmacro{\bsPX}{#1+#5*(.82*#3+.36*#4)}
  \pgfmathsetmacro{\bsPY}{#2+#5*(-.15*#3+.36*#4)}
  \draw[white,line width=1.7pt] (\bsPX,\bsPY)--(\bsPX,{\bsPY+#5*.43});
  \draw[bsInk,line width=.55pt] (\bsPX,\bsPY)--(\bsPX,{\bsPY+#5*.43});
  \foreach \bsC/\bsH in {bsGreen/0,bsBlue/.145,bsPurple/.29}{
    \filldraw[fill=\bsC,draw=white,line width=.4pt] (\bsPX,{\bsPY+#5*\bsH}) circle[radius=.033];
    \ifnum#6=2
      \ifdim\bsH pt=.145pt
        \filldraw[fill=white,draw=bsFrame,line width=.4pt] (\bsPX,{\bsPY+#5*\bsH}) circle[radius=.033];
      \fi
    \fi}
  \coordinate (#7) at (\bsPX,{\bsPY+#5*.43});
  \filldraw[fill=bsInk,draw=white,line width=.45pt] (#7) circle[radius=.040];
}

\resizebox{\linewidth}{!}{%
\begin{tikzpicture}[x=1cm,y=1cm]
\path[use as bounding box] (0,.58) rectangle (13.97,4.20);
\node[bs/subtitle] at (2.20,4.00) {(a) Available Observations};
\node[bs/subtitle] at (8.05,4.00) {(b) Stream Blocks};
\node[bs/subtitle] at (12.35,4.00) {(c) Sample Pixels};
\draw[bs/flow,line width=.45pt] (.44,1.22)--(.44,3.46) node[bs/math,above] {$t$};
\draw[bs/flow,line width=.45pt] (.44,1.22)--(.93,1.10)
  node[bs/math,below right,inner sep=.4pt] {$x$};
\draw[bs/flow,line width=.45pt] (.44,1.22)--(.67,1.49)
  node[bs/math,above left,inner sep=.4pt] {$y$};
\foreach \zz/\tm in {1.36/1,2.14/2,2.92/3}{
  \draw[bsFrame,line width=.4pt] (.38,\zz)--(.50,\zz);
  \node[bs/small,anchor=east] at (.35,\zz) {$t_{\tm}$};
}
\foreach \xx/\cc/\kind/\mm in {.84/bsPurple/1/1,2.02/bsBlue/2/2,3.20/bsGreen/3/3}{
  \draw[bs/guide] (\xx,1.36)--(\xx,2.92);
  \draw[bs/guide] ({\xx+1.08},1.56)--({\xx+1.08},3.12);
\foreach \zz/\col/\row/\tt in {1.36/1/1/1,2.14/2/2/2,2.92/1/2/3}{
    \begin{scope}[shift={(\xx,\zz)}]
      \path[fill=\cc!25,draw=\cc!50,line width=.3pt]
        (0,0)--(.76,-.18)--(.76,-.225)--(0,-.045)--cycle;
      \path[fill=\cc!38,draw=\cc!50,line width=.3pt]
        (.76,-.18)--(1.08,.20)--(1.08,.155)--(.76,-.225)--cycle;
    \end{scope}
    \begin{scope}[shift={(\xx,\zz)},cm={.76,-.18,.32,.38,(0,0)}]
      \bsMap{\cc}{\kind}{\col}{\row}
      \ifnum\tt=1
        \path[bs/missing] (.75,.75) rectangle (1,1);
      \fi
      \ifnum\kind=1
        \ifnum\tt=3
          \path[bs/missing] (0,0)--(.25,0)--(.25,.25)--(.5,.25)--(.5,.5)--(0,.5)--cycle;
        \fi
        \ifnum\tt=2
          \path[bs/missing] (.75,0) rectangle (1,.5);
        \fi
      \fi
      \ifnum\kind=2
        \ifnum\tt=3
          \path[bs/missing] (0,.75) rectangle (.5,1);
        \fi
        \ifnum\tt=2
          \path[bs/missing] (0,0) rectangle (1,1);
        \fi
      \fi
      \ifnum\kind=3
        \ifnum\tt=1
          \path[bs/missing] (0,0) rectangle (.25,.5);
        \fi
        \ifnum\tt=3
          \path[bs/missing] (.5,0)--(1,0)--(1,.5)--(.75,.5)--(.75,.25)--(.5,.25)--cycle;
        \fi
      \fi
    \end{scope}}
  \node[bs/math] at ({\xx+.54},3.55) {$p_{\mm}$};
}

\path[bs/missing] (1.56,.73) rectangle ++(.19,.12);
\node[bs/note,anchor=west] at (1.81,.79) {Unobserved};

\draw[bs/flow,line width=.55pt] (4.61,2.08)--(4.69,2.08)--(4.89,2.30)--(5.06,2.30);
\draw[white,line width=2.2pt] (4.67,2.30)--(4.90,2.08);
\draw[bs/flow,line width=.55pt] (4.61,2.30)--(4.69,2.30)--(4.89,2.08)--(5.06,2.08);

\node[bs/title] at (5.94,3.35) {CPU Prefetch};
\bsBundle{5.20}{2.43}{2}{2}{2}{.91}{.65}
\bsBundle{5.42}{1.85}{3}{1}{2}{1.02}{1}
\node[bs/note,anchor=north] at (6.00,1.59) {New Block};

\node[bs/title] at (9.05,3.35) {GPU-Resident Blocks};
\begin{scope}[shift={(7.34,1.44)},scale=1.08]
\begin{scope}[shift={(-6.62,-1.44)}]
\begin{scope}[shift={(6.62,1.44)},cm={1,-.18,.44,.43,(0,0)}]
  \path[fill=bsFrame!9] (.07,-.12)--(2.27,-.12)--(2.27,2.08)--(.07,2.08)--cycle;
\end{scope}

\path[fill=bsFrame!25,draw=bsFrame,line width=.45pt]
  (6.62,1.44)--(8.82,1.044)--(8.82,.934)--(6.62,1.33)--cycle;
\path[fill=bsFrame!17,draw=bsFrame,line width=.45pt]
  (8.82,1.044)--(9.788,1.990)--(9.788,1.880)--(8.82,.934)--cycle;
\begin{scope}[shift={(6.62,1.44)},cm={1,-.18,.44,.43,(0,0)}]
  \path[fill=bsFrame!4,draw=bsFrame,line width=.5pt] (0,0) rectangle (2.2,2.2);
  \foreach \bsX/\bsY in {.10/.10,1.20/.10,.10/1.20,1.20/1.20}{
    \path[fill=white,draw=bsFrame!80,line width=.35pt]
      (\bsX,\bsY) rectangle ++(.95,.95);}

  \path[fill=bsReplace!8,draw=bsReplace,line width=.65pt]
    (.10,.10) rectangle ++(.95,.95);
\end{scope}

\bsBundle{7.248}{1.938}{1}{1}{1}{1.02}{1}
\bsBundle{8.348}{1.740}{2}{2}{2}{1.02}{1}
\bsBundle{7.864}{1.267}{3}{0}{2}{1.02}{1}
\node[bs/small,fill=white,inner sep=.6pt] at (7.89,2.85) {$t_1$};
\node[bs/small,fill=white,inner sep=.6pt] at (9.22,2.50) {$t_2$};
\draw[bsFrame,line width=.35pt] (8.30,.98)--(8.30,.90);
\node[bs/small] at (8.30,.80) {$t_3$};

\bsPin{7.248}{1.938}{.29}{.67}{1.02}{1}{pickA}
\bsPin{8.348}{1.740}{.72}{.58}{1.02}{2}{pickB}
\bsPin{7.864}{1.267}{.65}{.37}{1.02}{3}{pickC}
\end{scope}
\end{scope}

\draw[bs/flow,draw=bsReplace,line width=.85pt]
  (6.68,2.06) to[out=-8,in=155] (7.83,1.68);
\node[bs/note,text=bsReplace,anchor=south] at (7.09,2.21) {Replace};

\draw[bs/flow,draw=bsInk!55,line width=.60pt]
  (7.90,1.53) to[out=-115,in=25] (7.32,1.40);
\bsBundle{6.60}{1.14}{1}{2}{0}{.60}{.42}
\node[bs/note,text=bsRemove,text opacity=.65,anchor=north] at (6.95,.99) {Removed};

\draw[bsInk!65,line width=.50pt,preaction={draw=white,line width=1.6pt}]
  (pickA) to[out=30,in=175] (11.68,2.33);
\draw[bsInk!65,line width=.50pt,preaction={draw=white,line width=1.6pt}]
  (pickB) to[out=5,in=180] (11.68,2.33);
\draw[bsInk!65,line width=.50pt,preaction={draw=white,line width=1.6pt}]
  (pickC) to[out=-5,in=190] (11.68,2.33);
\fill[bsInk] (11.68,2.33) circle[radius=.028];
\draw[bs/flow] (11.68,2.33)--(12.28,2.33);
\node[bs/math] at (11.72,2.66) {$(x,y,t)$};
\node[bs/note,anchor=north] at (11.64,1.99) {Uniform Valid\\Pixels};
\path[rounded corners=4pt,fill=bsOrangeL!7,draw=bsOrange,line width=.65pt]
  (12.35,1.83) rectangle (13.84,2.83);
\node[bs/title,align=center] at (13.095,2.33) {Shared\\[-1pt]Field};
\end{tikzpicture}%

}
\caption{
\textbf{Sampling across space, time, and products.}
\textbf{(a)} We randomly shuffle spatial block--year pairs from
Cloud-Optimized GeoTIFFs (COGs), sampling across locations and years without
replacement within each pass.
\textbf{(b)} CPU workers read all available products for each selected block
and year. Each stack has one layer per product; pixels with no observations
are excluded. Prefetched blocks fill a fixed GPU buffer, where new blocks
replace the oldest.
\textbf{(c)} We sample uniformly, with replacement, from all valid pixels in
this buffer, using the same coordinates for all active products. 
}
\label{fig:blockstream-sampling}
\vspace{-10pt}
\end{figure}

\section{Implementation Details and Hyperparameters}
\subsection{PFF Training Details}
\label{sec:appendix_pff_training}

\textbf{Training batches.}
We train every PFF jointly on all 14 products. Each 8192$\times$8192 region is divided into $1024\times1024$ spatial blocks across the available years. We maintain a fixed-capacity GPU buffer containing several block--year pairs while CPU workers prefetch the next blocks. Each iteration samples $1{,}048{,}576$ valid coordinates uniformly from the resident buffer. The same coordinates are used for all products, and missing targets are masked independently. The block--year list is traversed without replacement before reshuffling, while pixels may be sampled repeatedly while their block remains resident.

\textbf{Decoders and losses.}
Each product uses a two-hidden-layer ReLU MLP decoder. We use width 1024 for AlphaEarth \citep{alphaearth} and TESSERA \citep{tessera} and 256 for the remaining products. AlphaEarth predictions are $\ell_2$-normalized and optimized with cosine distortion, $1-\cos(\hat{\mathbf{y}},\mathbf{y})$. All other products are standardized per channel and optimized with mean-squared error in standardized units. Each product loss is averaged over its observed targets, and the product losses are summed with equal weight.

\textbf{Optimization.}
We optimize the field and all decoders with Adam using $\beta_1=0.9$, $\beta_2=0.99$, and $\epsilon=10^{-8}$. For VM fields, we use a learning rate of $2\times10^{-2}$ for the factors and $10^{-3}$ for the basis projection and decoders. For K-Planes, we use $5\times10^{-3}$ for the planes and $10^{-3}$ for the decoders. Learning rates are warmed up linearly for 500 iterations and then cosine-decayed to $0.1$ of their initial value. K-Planes additionally uses spatial total variation, temporal smoothness, and temporal $L_1$ regularization with weights $2\times10^{-4}$, $10^{-3}$, and $10^{-4}$, respectively. Training uses mixed precision (\texttt{bf16}).

\textbf{Training length and checkpoint selection.}
For the experiments in this paper, every PFF is trained for $20{,}000$ iterations. Every 1000 iterations, we evaluate a fixed set of uniformly sampled coordinates and retain the checkpoint with the lowest mean normalized distortion across products. We use $1-R^2$ for Euclidean-valued products and $1-\mathrm{skill}$ for AlphaEarth. 

\subsection{Downstream Evaluation}
\label{sec:appendix_downstream}

We evaluate whether reconstructed features preserve the information used by downstream models. 
For each benchmark, we extract one feature vector per labeled pixel from either the original product or its reconstruction at the same location and year. 
Features remain frozen, and each source is evaluated with the same labels, split, sampling procedure, and probe configuration. 
We report \emph{retention} as
\[
    100 \times
    \frac{\text{score using reconstructed features}}
         {\text{score using original features}}.
\]
A retention of $100\%$ matches the source features, while values above $100\%$ indicate better downstream performance after reconstruction.

For PASTIS \citep{pastis}, SwissCrop25 \citep{swisscrop25}, and EuroMineNet \citep{eurominenet}, we transform each label-pixel center into the coordinate system of its regional field and query the corresponding location and calendar year. 
The original product is sampled at the same source cell. 
FTW is evaluated on its native chip grid, where the field is queried continuously at each chip-pixel center. 
In all cases, the probe receives only the feature vector at that pixel; we do not provide neighboring features or spatial context.

\textbf{PASTIS.}
PASTIS provides dense crop labels over France with five official folds \citep{pastis}. We query all products at 2019, use folds 1--3 for training, fold 4 for validation, and fold 5 for testing. Background and void labels are removed, leaving 18 crop classes. We standardize the frozen features and fit a multinomial logistic-regression probe on up to $500{,}000$ training pixels. The regularization parameter is selected from $C\in\{0.1,1,10\}$ using validation mIoU. We score the full test fold and report macro IoU over the 18 crop classes.

\textbf{SwissCrop25.}
SwissCrop25 contains dense crop-type labels from 2019--2025 over Switzerland \citep{swisscrop25}. We follow its leave-one-year-out evaluation for test years 2021--2025: year $T$ is held out for testing, year $T-1$ is used for validation, and the remaining years are used for training. This evaluates transfer across years, but is not a causal forecasting split because training may include observations after the test year. The label space contains 65 crop and 5 non-crop classes. Each fold contains approximately $1.5$--$1.6$ billion labeled training pixels, making iterative optimization over individual pixels unnecessarily expensive. Because the probe is linear, \textbf{we aggregate the training features into sufficient statistics---class counts, feature sums, and second moments---and solve the ridge-regression objective in closed form.} This exactly recovers the linear solution while using all labeled training pixels, without storing or repeatedly iterating over the full matrix. Regularization and class weighting are selected using validation crop mIoU, and we report overall accuracy and mean IoU over the 65 crop classes, averaged across the five test years. 

\textbf{EuroMineNet.}
EuroMineNet provides annual mine-footprint masks with an official site-level train, validation, and test split \citep{eurominenet}. We evaluate the years 2017--2024. For each year, we standardize the frozen per-pixel features and fit a balanced logistic-regression probe to predict mine footprint.

We report two footprint metrics. \emph{Overall footprint F1 (OF1)} measures pixel-level agreement over the complete reconstructed mine masks. \emph{Boundary F1 (B-F1)} computes the same F1 score only within a three-pixel band around the ground-truth footprint boundary, emphasizing the spatial detail needed to localize mine edges. For both metrics, predictions are reassembled into full site scenes, accumulated across years, and averaged across test sites with weights proportional to the number of scored pixels.

For change detection, the probe receives consecutive-year features and their difference,
\[
    [\mathbf{f}_t,\mathbf{f}_{t-1},\mathbf{f}_t-\mathbf{f}_{t-1}],
\]
and predicts the XOR of the corresponding footprint masks. Only $0.52\%$ of evaluated pixel pairs change. We therefore report area under the precision--recall curve (AUPRC) and $\mathrm{F1}@r$. AUPRC measures how well changed pixels are ranked across all thresholds. For $\mathrm{F1}@r$, if $k$ pixel pairs change in the ground truth, the $k$ highest-scoring predictions are labeled as change before computing F1. This evaluates whether the model places the correct amount of change in the correct locations.

\textbf{Fields of the World.}
Fields of the World (FTW) provides field-boundary and field-extent labels across multiple countries \citep{ftw}. We evaluate AlphaEarth and Sentinel-2 features in western Bahia, Brazil, and South Africa using the official train, validation, and test splits and the acquisition year associated with each region. Frozen features are standardized and evaluated with a balanced logistic-regression probe; $C\in\{0.1,1,10\}$ is selected on the validation set.

The two regions require different metrics because their labels provide different supervision. The Brazil subset contains field annotations without labeled background, so false-positive fields cannot be interpreted reliably. We therefore report \emph{object recall}: predicted field pixels are grouped into connected components, and a ground-truth field is recovered when a predicted component overlaps it at IoU $>0.5$. South Africa contains labeled field and background pixels, so we report pixel IoU for field extent. Both evaluations use the native FTW label grid.

\input{figures/appendix/full_temporal_append_result}
\input{figures/appendix/adapt_recon}

\input{figures/appendix/sample_recons}

\end{document}